\documentclass[twoside,11pt]{article}

\usepackage{blindtext}

\usepackage{jmlr2e}

\usepackage{amsmath}
\usepackage{amssymb}
\usepackage{subfigure}
\usepackage[utf8]{inputenc} % allow utf-8 input
\usepackage[T1]{fontenc}    % use 8-bit T1 fonts
\usepackage{hyperref}       % hyperlinks
\usepackage{url}  % simple URL typesetting
\usepackage{booktabs}       % professional-quality tables
\usepackage{amsfonts}       % blackboard math symbols
\usepackage{nicefrac}       % compact symbols for 1/2, etc.
\usepackage{microtype}      % microtypography
\usepackage{xcolor}         % colors
\usepackage{graphicx}
\usepackage{wasysym}
\usepackage{wrapfig}
\usepackage{graphicx}
\usepackage{bbm}
\usepackage{bm}
\usepackage{booktabs}
\usepackage{colortbl}
\usepackage{multirow}
\usepackage{tcolorbox}
\usepackage{float}
\usepackage{color}
\usepackage[misc]{ifsym}
\usepackage{algorithmic}
\usepackage{algorithm}

\definecolor{mygray}{gray}{0.9}
\definecolor{mygray1}{gray}{0.95}
\newcommand{\ie}{\textit{i}.\textit{e}.}
\newcommand{\eg}{\textit{e}.\textit{g}.}
\newtheorem{assumption}[theorem]{Assumption}
\ShortHeadings{Keypoint-Guided Optimal Transport: Models, Algorithms, and Applications}{Gu, Yang, Zeng, Sun, and Xu}

\firstpageno{1}
\begin{document}

\title{Keypoint-Guided Optimal Transport: Models, \\Algorithms, and Applications}
% with Applications in Heterogeneous Domain  Adaptation and \\ Image-to-Image Translation}

% \author{\name Author One \email one@stat.washington.edu \\
%        \addr Department of Statistics\\
%        University of Washington\\
%        Seattle, WA 98195-4322, USA
%        \AND
%        \name Author Two \email two@cs.berkeley.edu \\
%        \addr Division of Computer Science\\
%        University of California\\
%        Berkeley, CA 94720-1776, USA}

% \author{\name Xiang Gu\textsuperscript{1,2} \email xianggu@stu.xjtu.edu.cn \\
%        \name Yucheng Yang\textsuperscript{1} \email ycyang@stu.xjtu.edu.cn \\
%        \name Wei Zeng\textsuperscript{1} \email wz@xjtu.edu.cn \\
%        \name Jian Sun\textsuperscript{1}\thanks{Jian Sun is the corresponding author.} \email jiansun@xjtu.edu.cn \\
%        \name Zongben Xu\textsuperscript{1} \email zbxu@xjtu.edu.cn \\
%        \addr \textsuperscript{1}School of Mathematics and Statistics, Xi'an Jiaotong University, Xi'an, China\\
%        \textsuperscript{2}Peng Cheng Laboratory, Shenzhen, China}
\author{\name Xiang Gu \email xianggu@xjtu.edu.cn \\
       \addr School of Mathematics and Statistics\\
       Xi'an Jiaotong University, Xi'an, China\\
       Peng Cheng Laboratory, Shenzhen, China
       \AND
       \name Yucheng Yang \email ycyang@stu.xjtu.edu.cn \\
       \addr School of Mathematics and Statistics\\
       Xi'an Jiaotong University, Xi'an, China
       \AND
       \name Wei Zeng \email wz@xjtu.edu.cn \\
       \addr School of Mathematics and Statistics\\
       Xi'an Jiaotong University, Xi'an, China
       \AND
       \name Jian Sun\thanks{Jian Sun is the corresponding author.} \email jiansun@xjtu.edu.cn \\
       \addr School of Mathematics and Statistics\\
       Xi'an Jiaotong University, Xi'an, China
       \AND
       \name Zongben Xu \email zbxu@xjtu.edu.cn \\
       \addr School of Mathematics and Statistics\\
       Xi'an Jiaotong University, Xi'an, China}
\editor{}

\maketitle

\begin{abstract}%   <- trailing '%' for backward compatibility of .sty file
Existing Optimal Transport (OT) methods mainly derive the optimal transport plan/matching under the criterion of transport cost/distance minimization, which may cause incorrect matching in some cases. In real applications, annotating a few matched keypoints across domains is reasonable or even effortless in annotation burden. It is valuable to investigate how to leverage the annotated keypoints to guide the correct matching in OT. In this paper, we propose a novel KeyPoint-Guided model by ReLation preservation (KPG-RL) that searches for the optimal matching (\ie, transport plan) guided by the keypoints in OT. KPG-RL exploits a mask-based constraint of the transport plan to preserve the matching of keypoint pairs in transport, and guides the matching by relation of each data point to the keypoints.
% To impose the keypoints in OT, first, we propose a mask-based constraint of the transport plan that preserves the matching of keypoint pairs in transport. Second, we propose to preserve the relation of each data point to the keypoints to guide the matching. 
The KPG-RL is developed in both balanced and unbalanced/partial transport settings in Kantorovich and Gromov-Wasserstein formulations.
% The proposed KPG-RL model can be solved by Sinkhorn's algorithm and is applicable even when distributions are supported in different spaces. We further utilize the relation preservation constraint in the Kantorovich Problem and Gromov-Wasserstein model to impose the guidance of keypoints in them. Meanwhile, the proposed KPG-RL model is extended to the partial OT and unbalanced settings. 
% Moreover, we propose a neural transport approach for KPG-RL, which learns the transport plan from the dual formulation of $\chi^2$-regularized KPG-RL model based on deep learning techniques, scaling better to larger numbers of data. With the transport plan, the manifold barycentric projection and manifold sampling strategies 
Moreover, we deduce the dual formulation of $\chi^2$-regularized KPG-RL model, from which we learn the transport based on deep learning techniques, scaling better to larger numbers of data. 
With the learned transport plan, two novel neural transport strategies, named manifold barycentric projection and manifold sampling, are developed to transport source data to the target domain data manifold. 
As applications, we apply the proposed approach to the heterogeneous domain adaptation, multi-omic single-cell alignment, and image-to-image translation. Experiments verified the effectiveness of our approach. 
% \blindtext
\end{abstract}

\begin{keywords}
Keypoint-guided optimal transport, relation preservation, masked plan, neural transport, heterogeneous domain adaptation, image-to-image translation, multi-omic single-cell alignment
\end{keywords}

\begin{figure}[t]
	\centering
	\subfigure[Samples]{\label{fig:toy_data}\includegraphics[width=0.4\columnwidth]{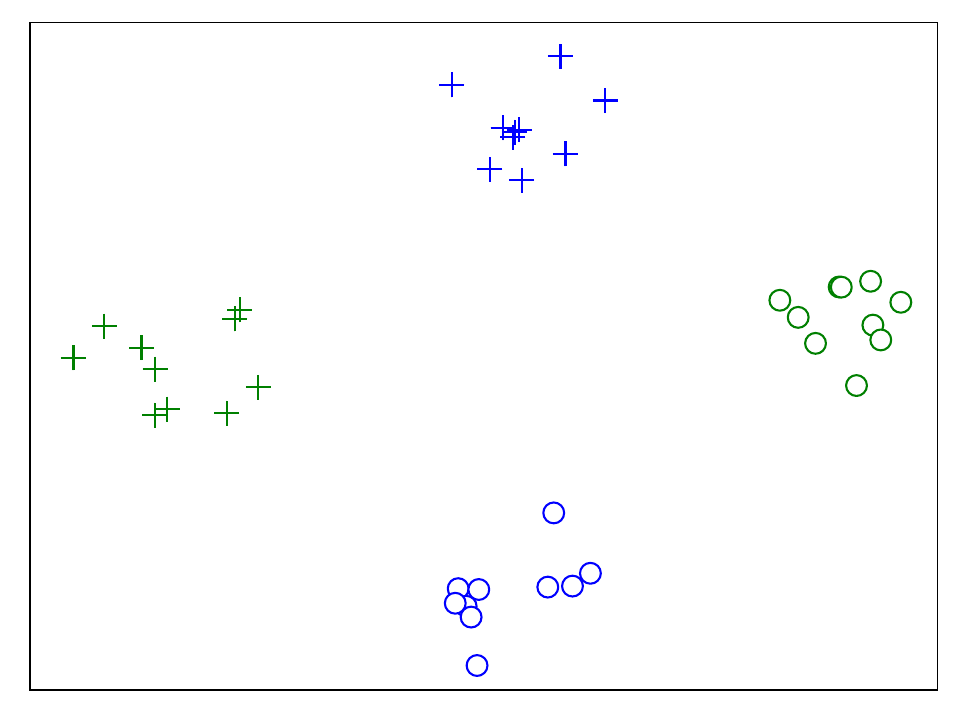} 
		}
	\subfigure[KP]{ \label{fig:toy_ot}\includegraphics[width=0.4\columnwidth]{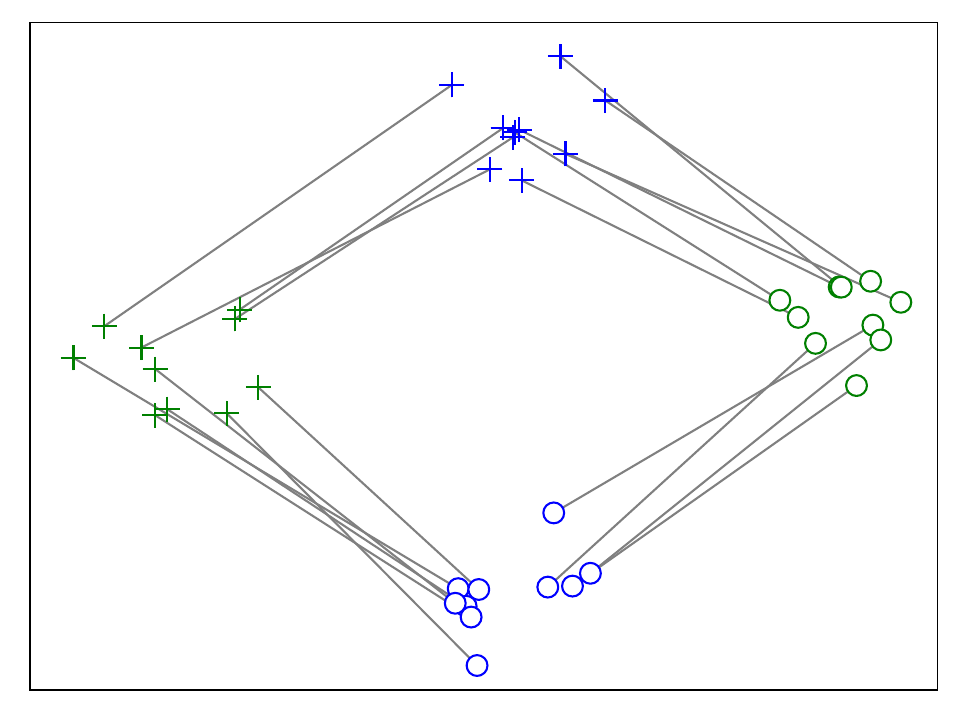} 
		}
	\subfigure[GW] { \includegraphics[width=0.4\columnwidth]{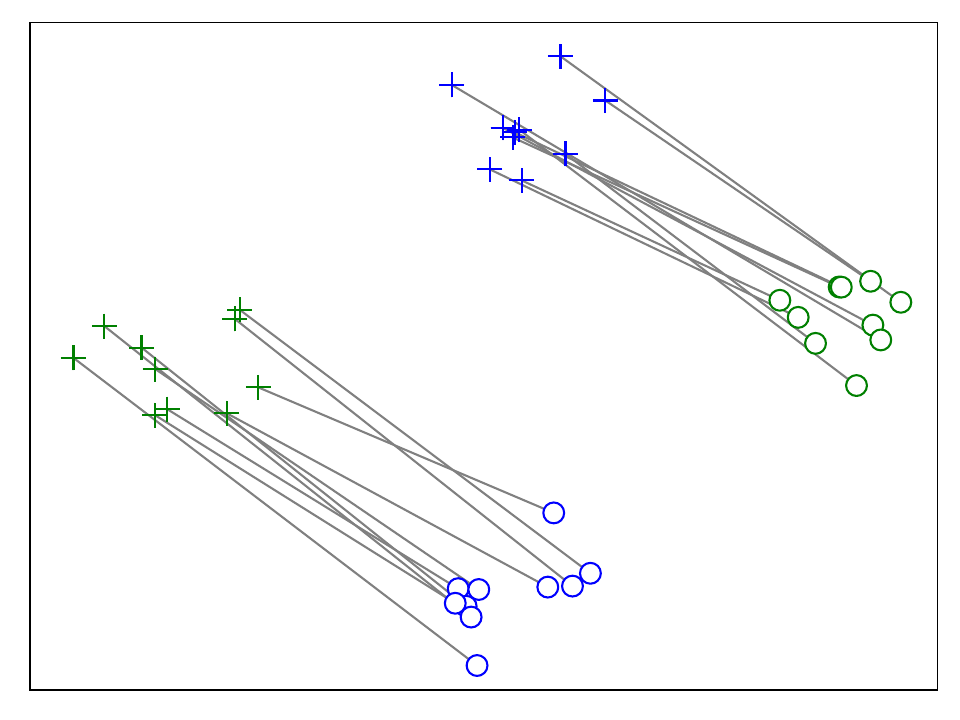}
		\label{fig:toy_gw} }
	\subfigure[KPG-RL (ours)]{ \includegraphics[width=0.4\columnwidth]{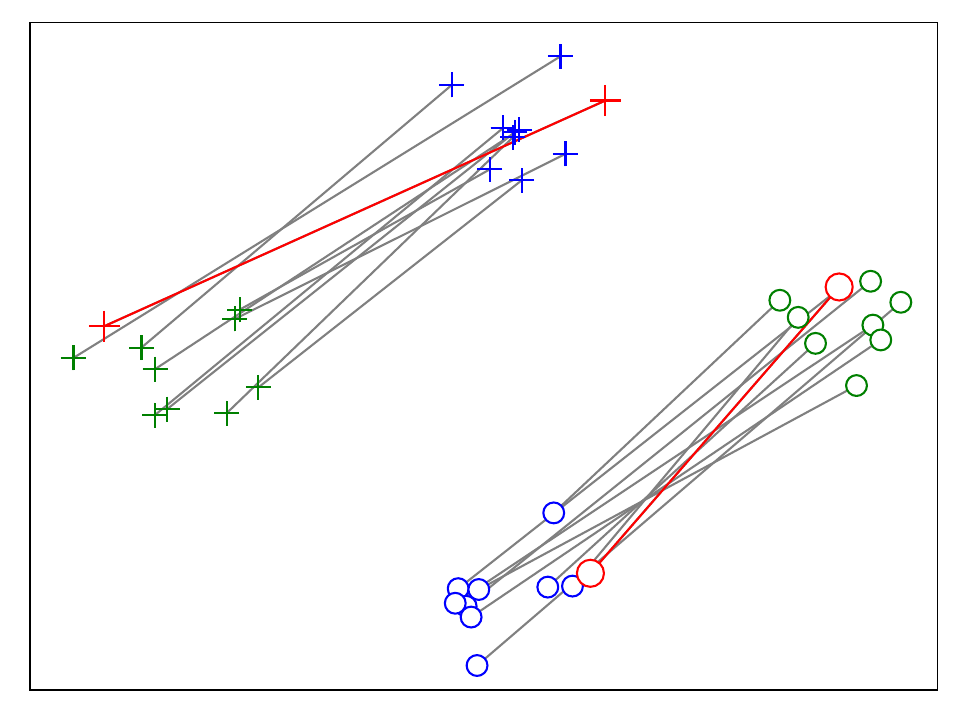} 
		\label{fig:toy_kpg}}
	\caption{(a) Positive (cross) and negative (circle) samples of source (in blue) and target (in green) distributions. (b) KP distorts the data structure, leading to mismatch of some positive and negative samples. (c) GW model better preserves the data structure, but completely mismatches the samples. (d) Our proposed KPG-RL model utilizes some keypoints (red pairs) to guide the transport and produce correct matching.}
	\label{fig:toy_example}
% 	\vspace{-0.3cm}
\end{figure}

\section{Introduction}\label{sec:introduction}
Optimal Transport (OT)~\citep{villani2009optimal} is a mathematical tool for distribution alignment, mass transport, \textit{etc.} OT has gained increasing attention in the machine learning community. OT aims to derive a transport map or plan between a source and a target distribution, such that the transport cost is minimized.  Due to its capacity to exploit the geometric property of data, OT has been employed in many applications, \eg, computer
vision~\citep{bonneel2011displacement,solomon2015convolutional}, natural language processing~\citep{kusner2015word,huang2016supervised,alvarez2018gromov,yurochkin2019hierarchical}, generative adversarial network~\citep{arjovsky2017wasserstein}, domain adaptation~\citep{7586038}, clustering~\citep{ho2017multilevel,huynh2021efficient}, anomaly detection~\citep{tong2022fixing}, dataset comparison~\citep{alvarez2020geometric}, \textit{etc}.
OT has two typical formulations, \ie, Monge's formulation~\citep{monge1781memoire} and Kantorovich's formulation~\citep{kantorovich1942translocation}. Kantorovich's formulation~\citep{kantorovich1942translocation} relaxes Monge's formulation and attracts broader studies in applications. 
We mainly deal with Kantorovich's formulation in this paper.
% Unless otherwise stated, by OT, we refer to Kantorovich's formulation in this paper. 
The original OT model, \ie, the Kantorovich Problem~\citep{kantorovich1942translocation} (KP), is a linear program that is computationally expensive. The entropy-regularized OT~\citep{cuturi2013sinkhorn,dessein2018regularized} introduces the entropy as regularization to the OT model, which is solved by the computationally cheaper Sinkhorn-Knopp algorithm~\citep{sinkhorn1967concerning,lin2022efficiency}, allowing the use of automatic differentiation~\citep{genevay2018learning}. 
KP transports all the mass of source distribution to exactly match the mass of target distribution. But in some cases, only partial mass of source and target distributions should be matched, \eg, the source or target samples contain outliers. To overcome this limitation, the partial OT~\citep{guittet2002extended,figalli2010optimal,caffarelli2010free,chapel2020partial}, unbalanced OT~\citep{chizat2018interpolating,liero2018optimal}, and robust OT~\citep{balaji2020robust,mukherjee2021outlier,le2021robust} models are presented, which allow to only transport partial mass of distribution. Another extension of KP is the Gromov-Wasserstein (GW) model~\citep{sturm2006geometry,memoli2011gromov} that computes the distance between metrics defined within each domain rather than between samples across domains as in KP. 

In most of the above OT models, the main criterion for the optimal transport plan/matching is the minimization of total transport distance or distortion over all samples. 
Though having achieved promising results in many applications, without any additional guidance, the OT models may lead to incorrect matching or undesired alignment of samples. Figures~\ref{fig:toy_ot} and ~\ref{fig:toy_gw} illustrate examples of incorrect matching produced by KP and GW models.
In real applications, it is reasonable to annotate some paired keypoints across domains for guiding the matching in OT. For instance, in non-rigid point set or image registration~\citep{myronenko2010point,crum2004non}, a few keypoint pairs that should be matched are annotated in two point sets/images.  
In semi-supervised or heterogeneous domain adaptation~\citep{saito2019semi,tsai2016learning}, along with a large amount of labeled source domain data, there are a few labeled target domain data available, that could be directly taken as keypoints.
Therefore, it is valuable and important to investigate how to take advantage of those keypoints to guide the correct matching in OT. Figure~\ref{fig:toy_kpg} shows an example that if the guidance of a few keypoints is properly imposed, the correctness of matching can be improved. 

In this paper, we propose a novel KeyPoint-Guided model by ReLation preservation (KPG-RL) for leveraging the annotated keypoints to guide the matching in OT. 
% To impose the guidance of keypoints, first, we
The KPG-RL model preserves the matching of keypoint pairs in OT using a mask-based constraint of the transport plan, and enforce the matching of the data points near to paired keypoints across domains by aligning the relation of each data point to the keypoints.
% In KPG-RL, we first preserve the matching of keypoint pairs in OT using a mask-based constraint of the transport plan. We then propose to preserve the relation of each data point to the keypoints in transport, enforcing the matching of the data points near to paired keypoints across domains. 
KPG-RL is applicable even when distributions lie in different spaces. The above mask and relation-based techniques are employed in KP and GW to impose the keypoint guidance, developing keypoint-guided KP and GW. We show that the keypoint-guided KP and GW respectively provide a metric and divergence between distributions. 
% We further enforce the relation preservation constraint in KP and GW to impose the guidance of keypoints in them, for which the theoretical properties are developed.
To handle the applications where only partial mass should be transported, we extend the KPG-RL model to both partial OT and unbalanced OT settings in KP and GW formulations, and develop corresponding solution algorithms. 

Meanwhile, we present the dual formulation of $\chi^2$-regularized KPG-RL model, enabling us to learn the transport plan using mini-batch-based deep learning techniques, scaling better to larger numbers of data points.  
% Specifically, we deduce the dual formulation of the $\chi^2$-regularized KPG-RL model, in which we parameterize the optimization variables, \textit{say potentials}, with deep neural networks. The optimal transport plan can be calculated using the trained potentials, according to the strong duality. 
With the learned transport plan, we propose two novel neural transport strategies, named manifold barycentric projection (dubbed KPG-RL-MBP) and manifold sampling (dubbed KPG-RL-MSP), to transport source data to the target domain data manifold.  KPG-RL-MBP transports the source samples to the barycenter of the conditional transport plan constrained into the target data manifold, and KPG-RL-MSP aims to sample data on the target data manifold from the conditional transport plan implicitly given the source samples.

{As applications, we apply the proposed KPG-RL to the heterogeneous domain adaptation (HDA)~\citep{tsai2016learning}, multi-omic single-cell alignment~\citep{demetci2022scot}, and image-to-image (I2I) translation~\citep{isola2017image} tasks. 1) \textit{HDA} is a transfer learning task that aims to transfer the knowledge of large amounts of labeled source domain data to the target domain where a few labeled and larger amounts of unlabeled data are available for training. The ``heterogeneity'' implies that the source and target domain data are in heterogeneous feature spaces,
\eg, generated by different deep networks. This heterogeneity poses a major obstacle in adapting the source-trained model to the target domain. We take the labeled target domain data and source class centers as keypoints and transport source domain data to the target domain by our KPG-RL model (or partial KPG-RL model). Upon the transported source domain data and the labeled target domain data, a classification model is trained that is transferable to the target domain. 
% Experiments show that the proposed KPG-RL model is effective for HDA.
2) For \textit{Multi-Omic Single-Cell Alignment} task, the goal is to match multiple modalities of cells, where we only observe the multiple modalities simultaneously for a bunch of cells, which is of great practical importance in biological data analysis. In this task, we take the setting that a few matched cross-modality cells and a larger number of unmatched cells of each modality are provided. We take the matched cells as keypoints. Considering that the cell types in different modalities are often unbalanced, we apply our partial and unbalanced KPG-RL models to search for the matching of different cell modalities. 
3) The \textit{I2I Translation} task is for evaluating our neural transport strategies, \ie, KPG-RL-MBP and KPG-RL-MSP. In I2I translation task, we are given the images of two distinct domains with different styles, objects, \textit{etc}. We consider the task setting that a large number of unpaired and a few paired images across the source and target domains are given in training. The goal is to transport the source images to the target domain using the paired images as guidance. We take the paired images as keypoints and apply the proposed KPG-RL-MBP and KPG-RL-MSP to transport the source images to the target domain. Extensive experimental results in the above three tasks demonstrated the effectiveness of our approach to real data.}
% Experiments on digits and natural animal images confirm that KPG-RL-MBP and KPG-RL-MSP can realize the guidance of keypoints and produce clear transported images. 

Our contributions are summarized as follows:
\begin{itemize}
    \item We propose the novel KPG-RL model to leverage the given paired keypoints to guide the correct matching in OT. KPG-RL employs a mask-based constraint on the transport plan to enforce the matching of keypoint pairs, and impose the guidance of keypoints ueing the relation of each data point to the keypoints.
    \item We develop the KPG-LR in both balanced and unbalanced/partial OT settings for both Kantorovich and Gromov-Wasserstein formulations, for which the solution algorithms are devised and the metric properties are analyzed.
    \item We deduce the dual formulation of the KPG-RL model, based on which two neural transport strategies, \ie, manifold barycentric projection and manifold sampling, are proposed to transport the source data to the target domain data manifold.
    \item We apply the proposed method to heterogeneous domain adaptation, multi-omic single-cell alignment, and image-to-image translation tasks. Extensive experiments verified the effectiveness of our approach.
\end{itemize}

{This paper is extended from our conference version~\citep{gu2022keypointguided}. In the conference version~\citep{gu2022keypointguided}, we developed the keypoint-guided OT models in balanced setting for Kantorovich and Gromov-Wasserstein formulations, and in partial transport setting for Kantorovich formulation, which are evaluated in HDA application. The major extensions of this journal paper over conference version are the neural transport strategies based on deduced dual formulation of KPG-RL, the unbalanced keypoint-guided Kantorovich and Gromov-Wasserstein models, and applications to multi-omic single-cell alignment and I2I translation. We summarize additional contributions as follows:
(1) We develop the keypoint-guided unbalanced OT in both Kantorovich and Gromov-Wasserstein formulations, solved by generalized Sinkhorn iteration. Meanwhile, the keypoint-guided partial OT is extended to the Gromov-Wasserstein model.
(2) We present the dual formulation for the $\chi^2$-regularized KPG-RL model, which is an unconstrained optimization problem. To solve the dual problem, we parameterize the potentials using neural networks that are trained using mini-batch-based optimization algorithms. The transport plan is given by the strong duality using the learned potentials.
(3) With the learned transport plan, we propose two neural transport strategies, namely manifold barycentric projection and manifold sampling, to transport source data to the target data manifold. 
(4) We additionally conduct experiments of image-to-image translation task on MNIST and real natural animal data, and multi-omic single-cell alignment task on real cell data. Extensive experimental results demonstrate the effectiveness of the additionally proposed techniques.}

% We extend the KPG-RL model to the large-scale OT situation, as in Sect.~\ref{sec:large_scale}. Specifically, we deduce the dual formulation of the KPG-RL model which can be solved via mini-batch-based training. We further propose a GAN-based approach to transport the source domain data to the target domain. We apply the above large-scale KPG-RL to the I2I translation task, as in Sect.~\ref{sec:i2i_translation}. Experiments on digits and real images show the effectiveness of the large-scale KPG-RL.

% \vspace{0.4\baselineskip}\textit{}
In the following sections, we discuss the related works in Sect.~\ref{sec:related_work}, and introduce the background of OT in Sect.~\ref{sec:background}. In Sect.~\ref{sec:kpg_ot}, we detail our KPG-RL model with extensions to Kantorovich Problem and Gromov-Wasserstein model in partial and unbalanced OT settings. In Sect.~\ref{sec:large_scale}, we elaborate on the dual KPG-RL model with the neural transport strategies of manifold barycentric projection and manifold sampling. 
In Sect.~\ref{sec:app_hda}, we apply our approach to HDA,  multi-omic single-cell alignment and I2I translation. Section~\ref{sec:conclusion} concludes this paper.

{
\vspace{0.4\baselineskip}\textit{Notations.}  $\Sigma^{m}=\{\bm{p}\in\mathbb{R}_+^{m+1}|\sum_{i}p_i=1\}$ is the probability simplex. $\langle \cdot,\cdot\rangle_F$ is the Frobenius dot product of two matrices. $\odot$ stands for the Hadamard product. $\mathbbm{1}_m$ denotes $m$-dimensional all-one vector. $\mathbbm{1}_{m\times n}$ denotes the $m\times n$ matrix with all-one elements.  $\pi_{i,:}$ and $\pi_{:,j}$ are respectively the $i$-th row and $j$-th column of matrix $\pi$. We use bolded lowercase letters to denote vectors, and corresponding unbolded letters to denote their elements, \eg, for $\bm{p}\in\mathbb{R}^m$, $\bm{p}=(p_1,p_2,\cdots,p_m)$. In the absence of ambiguity, we also denote empirical distributions using bolded lowercase letters. We finally denote $[n]=\{1,2,\cdots,n\}$.
}
% Acknowledgements and Disclosure of Funding should go at the end, before appendices and references
\section{Related Works}\label{sec:related_work}

We review below the most related OT models, HDA methods,  multi-omic single-cell alignment methods, and I2I translation approaches to our work.

{\vspace{0.4\baselineskip}\textit{OT models.} The GW model~\citep{memoli2011gromov} seeks a ``distance-preserving'' transport plan such that the distance between transported points in the target domain is the same as the distance between the original points in source domain. Our KPG-RL model aims to use keypoint pairs to guide the matching (\ie, transport plan) in OT by preserving the relation of each point to the keypoints. Our ``relation-preserving'' scheme preserves the relation of data \textit{w.r.t.} the given keypoints, different from the pairwise distance-preserving constraint in GW. We experimentally verified the effectiveness of the relation-preserving scheme for introducing the guidance of keypoints in OT. 
% Meanwhile, our KPG-RL model can be combined with GW to impose the guidance of keypoints in GW. 
From the computational point of view, GW is a non-convex quadratic program, while KPG-RL is a linear program.  
% Sato~\etal~\citep{sato2020fast} propose the Anchor Wasserstein distance that is an approximation of GW and can be efficiently computed.
\citet{lin2021making} use the anchors to encourage clustering of data and to impose rank constraints on the transport plan to improve its robustness to outliers. The ``anchors'' in~\citep{lin2021making} are intermediate points in computation for improving robustness, different from the ``keypoints'' in this paper, which are the annotated paired data for guiding the matching in OT. 
% Graph OT~\citep{chen2020graph} and fused GW~\citep{titouan2019optimal} fuse the GW and KP models for structure and node matching between graphs. 
% TLB~\citep{memoli2011gromov,sato2020fast} is a lower bound of GW that can be computed faster. TLB takes the ordered distance of each point to all the points in the same domain as features, and then performs the Kantorovich formulation of OT using such features.
% Differently, our method uses a carefully designed relation of each point to the keypoints to impose the guidance of keypoints to the other points.
{FlowAlign~\citep{le2021flow} and AEAW~\citep{sato2020fast} aim to reduce the computation of comparison of probability measures in heterogeneous spaces by aligning/preserving the distance between data points to anchor points in each space, sharing similar spirits with our method in methodology. FlowAlign~\citep{le2021flow} considers the tree metric space and aligns flows (\ie, tree distances) from a root/anchor to other nodes to find the transport plan.  AEAW~\citep{sato2020fast} represents each data point as its distance to all the other points (anchors), based on which the energy and Wasserstein distances are defined. Different from FlowAlign~\citep{le2021flow} and AEAW~\citep{sato2020fast}, our method aims to utilize a few paired keypoints annotated by humans to guide correct transport, which is realized by preserving the relation scores (defined in Sect.~\ref{sec:relation}) of each data point to keypoints. 
% In experiments in Sect.~\ref{sec:hda}, we show that our relation-preservation scheme is more effective than the distance-preservation scheme for imposing the guidance of paired keypoints.
}
% As will be shown in experiments, our method is more effective than TLB when some keypoints are given. 

Hierarchical OT~\citep{yurochkin2019hierarchical,lee2019hierarchical,xu2020learning} transports points by dividing them into some subgroups and then derives the transport of these subgroups using OT. Different in goal and methodology from Hierarchical OT, we impose the guidance of keypoints for pursuing correct matching in OT by preserving the relation to the keypoints. We do not explicitly divide the points into subgroups, and there is no hierarchy in our method. 
\citet{7586038} constrain the cost function to encourage the matching of labeled data across source and target domains that share the same class labels for domain adaptation. They use the Laplacian regularization to preserve the data structure. Differently, we explicitly model the guidance of keypoints matching to the other data points in our OT formulation. The matching of paired keypoints is enforced by our mask-based constraint on the transport plan. \citet{zhang2022fine} propose Masked OT model as a regularization term to preserve the local feature invariances between fine-tuned and pretrained graph neural network (GNN) for the fine-tuning of GNNs. Though the Masked OT~\citep{zhang2022fine} shares similar spirits to the mask-based modeling in our method, our main contribution is the relation preservation for imposing the guidance of keypoints, different from~\citep{zhang2022fine}. For our mask-based modeling, it is utilized to impose the matching of keypoints with theoretical guarantee. While the mask in~\citep{zhang2022fine} aims to preserve the local information of pretrained models. The motivation and design of our mask are different from those in~\citep{zhang2022fine}. 
}
% Our KPG-RL is orthogonal to the graph, fused, and hierarchical OT, and can be potentially combined with them to impose the guidance of keypoints as constraint in these OT models.

% \textbf{Large-scale OT methods.}

\vspace{0.4\baselineskip}\textit{Heterogeneous domain adaptation.} HDA methods could be roughly categorized into cross-domain mapping and common subspace learning methods. The cross-domain mapping approaches~\citep{tsai2016learning,yan2018semi,shen2018unsupervised,mozafari2016svm,zhou2019multi} learn a transform to map the source features or model parameters to target domain to achieve adaptation. The common subspace learning approaches~\citep{wang2011heterogeneous,wu2021heterogeneous,yan2017learning,zhou2019deep,wang2022cross,fang2022semi} learn domain-specific projections to map source and target domain data into a common subspace such that their distributions are aligned. Our method for HDA belongs to the first category, and could be mostly related to~\citep{yan2018semi}. The method in~\citep{yan2018semi} transports source samples to target domain using GW model regularized by the distance between the center of transported source samples and the center of labeled target samples having the same class labels. Different from~\citep{yan2018semi}, we take each labeled target data and its corresponding source class center as a keypoint pair and preserve the relation of each data point to the set of keypoints when conducting OT. 

{\vspace{0.4\baselineskip}\textit{Multi-omic single-cell alignment.} Biological data translation/alignment has achieved much attention recently~\citep{tong2024improving,pmlr-v238-tong24a,pariset2024unbalanced,monge-map,demetci2022scot}. The multi-omic single-cell alignment aims to match cells of different modalities. Previous multi-omic single-cell alignment methods~\citep{demetci2022scot,10.1093/bioinformatics/btab594,10.1093/bioadv/vbad171} often align cells based on partial/unbalanced Gromov-Wasserstein models~\citep{cao2020unsupervised,10.1093/bioinformatics/btab594,demetci2022scot} or GAN~\citep{10.1093/bioadv/vbad171}, in an unsupervised manner. This paper handles a semi-supervised setting that a few matched cross-modality cells are provided along with the unmatched cells. We take the matched cells as keypoints, and apply our proposed keypoint-guided partial and unbalanced Gromov-Wasserstein models to find the matching of cells. We show in experiments that the matching performance can be improved by our approach, given a few keypoint pairs.}
% The setting is reasonable because of the present of a few recent co-assay techniques. 

\vspace{0.4\baselineskip}\textit{Image-to-image translation.}  In the earlier supervised I2I translation works~\citep{isola2017image,wang2018discriminative,zhu2017toward,zhang2020cross,bansal2017pixelnn}, researchers use many aligned cross-domain image pairs to obtain the translation model that translates the source images to the desired target images. However, training supervised translation is not very practical because of the difficulty and high cost of acquiring these large, paired training data in real-world tasks. Unsupervised I2I translation methods~\citep{zhu2017unpaired,kim2017learning,yi2017dualgan,choi2021ilvr,meng2021sdedit,Albergo2023StochasticIA,bortoli2021diffusion,chen2022likelihood,shi2023diffusion,pmlr-v202-pooladian23a,pmlr-v238-tong24a,tong2024improving} use two large but unpaired sets of training images to convert images between representations. Though promising, the unsupervised methods require additional knowledge, \eg, domain-common knowledge~\citep{choi2021ilvr}, to guide the desired translation. Semi-supervised I2I translation approach~\citep{mustafa2020transformation} leverages source images alongside a few source-target aligned image pairs for training, reducing the cost of human labeling or expert guidance in supervised I2I. In this paper, we tackle the semi-paired I2I translation task in which a large number of unpaired source and target images, and a few source-target aligned image pairs are available for training, because obtaining unlabeled target images could be reasonable in many applications.  We take the given a few source-target aligned image pairs as keypoints and utilize our proposed KPG-RL-MBP and KPG-RL-MSP to translate the source images to the target domain. In methodology, the diffusion or flow approaches~\citep{shi2023diffusion,pmlr-v202-pooladian23a,tong2024improving,pmlr-v238-tong24a} based on {S}chrödinger bridges or conventional OT are mostly related to our proposed KPG-RL-MBP and KPG-RL-MSP. Differently, KPG-RL-MBP and KPG-RL-MSP are implemented using adversarial training based on our keypoint-guided OT plan. We also use our keypoint-guided OT model to replace the conventional OT in~\citep{pmlr-v202-pooladian23a,tong2024improving,pmlr-v238-tong24a} to guide flow matching, achieving improved performance.  

\section{Background on Optimal Transport}\label{sec:background}
As background to our approach, we reintroduce the Kantorovich problem, partial and unbalanced OT models, and Gromov-Wassrstein model in this section.

\vspace{0.4\baselineskip}\textit{Kantorovich Problem (KP).} Optimal transport considers two sets of data points, \ie, the source data $\bm{X} = \{{x}_{i} \}_{i=1}^{m}$ and the target data $\bm{Y} = \{{y}_{j} \}_{j=1}^{n}$, of which the empirical distributions are $\bm{p}=\sum_{i=1}^{m}p_{i}\delta_{x_{i}}$ and $\bm{q}=\sum_{j=1}^{n}q_{j}\delta_{y_{j}}$. With a slight abuse of notations, we also denote $\bm{p}=(p_1,p_2,\cdots,p_m)^{\top} \in\Sigma^{m-1}$ and $\bm{q}=(q_1,q_2,\cdots,q_n)^{\top}\in\Sigma^{n-1}$ as the mass supported on $\bm{X}$ and $\bm{Y}$, respectively. We define the cost matrix between $\bm{X}$ and $\bm{Y}$ as $C=(C_{i,j}) \in \mathbb{R}^{m\times n}$ with $C_{i,j} = c(x_{i}, y_{j})$, where $c$ is a cost function, which is set to the squared $L_2$-distance between $x_{i}$ and $y_j$ in our experiments.
OT aims to optimally transport $\bm{p}$ towards $\bm{q}$ at the smallest cost, formulated as the following Kantorovich Problem (KP):
\begin{equation}\label{eq:ot}
\min_{\pi \in \Pi(\bm{p},\bm{q})}L_{kp}(\pi)\triangleq\langle \pi , C \rangle_F,
% = \sum_{i=1}^{m}\sum_{j=1}^{n} \pi_{i,j} C_{i,j}, 
\mbox{ s.t. } \Pi(\bm{p},\bm{q}) = \{ \pi \in \mathbb{R}^{m\times n}_+ \vert \pi \mathbbm{1}_{n} = \bm{p}, \pi^{\top} \mathbbm{1}_{m} = \bm{q}\}.
\end{equation} 
When $c$ is taken as a distance metric (\textit{aka.} ground metric), the minimum value of the objective function in Eq.~\eqref{eq:ot} is a distance between $\bm{p}$ and $\bm{q}$, named Wasserstein distance. 

\vspace{0.4\baselineskip}\textit{Partial OT model.} The KP in Eq.~\eqref{eq:ot} takes the mass preserving assumption that all the mass of $\bm{p}$ should be transported to exactly match the mass of $\bm{q}$. 
In many applications, only partial mass should be transported. 
The partial OT model~\citep{figalli2010optimal,caffarelli2010free}  seeks the minimal cost of transporting only $s$ unit mass from $\bm{p}$ to $\bm{q}$, where $0 \leqslant s \leqslant \min(\Vert \bm{p}\Vert_1, \Vert \bm{q}\Vert_1)$, formulated as 
\begin{equation}\label{eq:pot}
\min_{\pi \in \Pi^{s}(\bm{p},\bm{q})}L_{kp}(\pi), \mbox{ s.t. } \Pi^{s}(\bm{p},\bm{q}) = \{ \pi \in \mathbb{R}^{m\times n}_+ \vert \pi \mathbbm{1}_{n} \leqslant \bm{p}, \pi^{\top} \mathbbm{1}_{m} \leqslant \bm{q}, \mathbbm{1}_{m}^{\top}\pi \mathbbm{1}_{n} = s \}.
\end{equation}
Note that in partial OT, the total mass $\Vert \bm{p}\Vert_{1}$ and $\Vert \bm{q}\Vert_{1}$ are not necessarily equal. 
% In the definition of $\Pi(\bm{p},\bm{q})$ and $\Pi^{s}(\bm{p},\bm{q})$, $m$ and $n$ are respectively the length of $\bm{p}$ and $\bm{q}$.

{\vspace{0.4\baselineskip}\textit{Unbalanced OT model.} 
% Similar to partial OT, the unbalanced OT considers that the data points may contain outliers that should not be transported.  
Unbalanced OT~\citep{liero2018optimal} allows the variation of total mass during transport.
The unbalanced OT model relaxes the constraints in KP (Eq.~\eqref{eq:pot}) as a regularization term added to the objective function, formulated as
\begin{equation}\label{eq:uot}
\min_{\pi \in \mathbb{R}^{m\times n}_+} L_{kp}(\pi) + \mu \mathcal{D}(\pi\mathbbm{1}_{n}|\bm{p}) + \mu \mathcal{D}(\pi^{\top} \mathbbm{1}_{m} | \bm{q}),
\end{equation}
where $\mathcal{D}$ is a distribution divergence or distance. Without loss of generality, we choose the KL-divergence in this paper, \ie, $\mathcal{D}(\bm{p}|\bm{q})=\sum_ip_i\log\left(\frac{p_i}{q_i}\right)$. $\mu$ is the regularization coefficient.}

\vspace{0.4\baselineskip}\textit{Gromov-Wasserstein (GW) model.} The GW~\citep{memoli2011gromov} model minimizes the distortion when transporting the whole set of points from one space to another. The GW model relies on the intra-domain distance of source domain as $C^s=(C^s_{i,k})\in \mathbb{R}^{m\times m}$ and target domain as $C^t=(C^t_{j,l})\in \mathbb{R}^{n\times n}$, where $C^s_{i,k}$ is the distance between source domain data $x_i$ and $x_k$, and $C^t_{j,l}$ is the distance between target domain data $y_j$ and $y_l$. The GW model is given by 
\begin{equation}
    \min_{\pi \in \Pi(\bm{p},\bm{q})} L_{gw}(\pi) \triangleq \sum_{i,k=1}^m\sum_{j,l=1}^n \pi_{i,j}\pi_{k,l}\vert C^s_{i,k}-C^t_{j,l}\vert^2.
\end{equation}
By replacing $L_{kp}$ with $L_{gw}$ in Eqs.~\eqref{eq:pot} and \eqref{eq:uot}, we obtain the partial and unbalanced Gromov-Wasserstein models, respectively.
\begin{figure}[t]
	\centering
	\includegraphics[width=0.98\columnwidth]{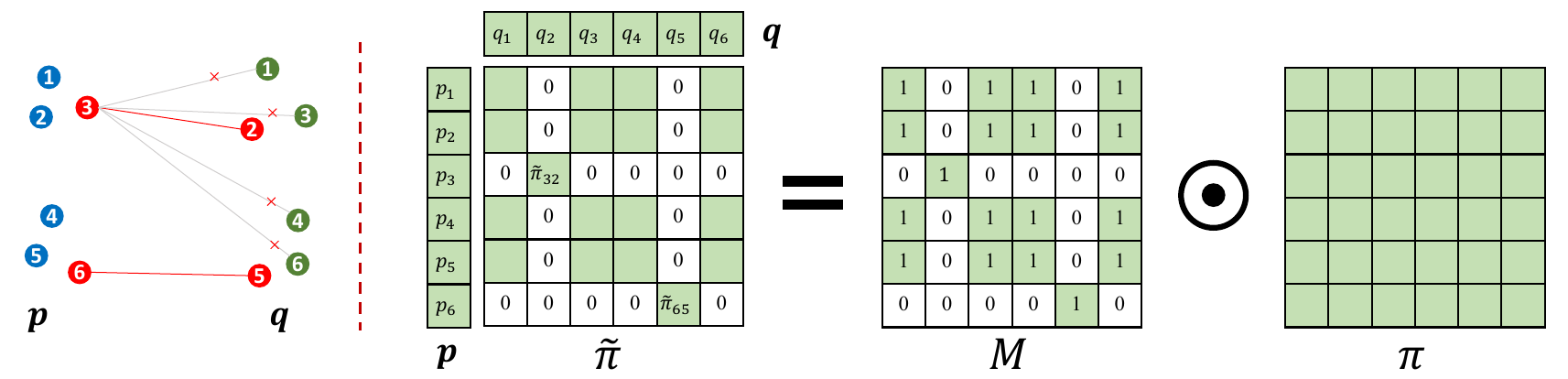} 
	\caption{Example of modeling the matching of keypoints (red) using mask. The left part illustrates the matching of data points with keypoint pairs $\mathcal{K}=\{(3,2),(6,5)\}$. To preserve the matching of keypoint index pair $(i,j)$, \eg, $(3,2)$, the $i$-th row and $j$-th column of transport plan $\Tilde{\pi}$ must be zeros except $\Tilde{\pi}_{i,j}$. Therefore, we can model $\Tilde{\pi}$ by $\Tilde{\pi} = M \odot \pi$, where $M$ is the mask determined by the keypoints, and $\pi$ is to be optimized.}
	\label{fig:mask_modeling_main}
\end{figure}

\vspace{0.5\baselineskip}
We next build our keypoint-guided OT in both balanced (with mass preserving assumption) and partial/unbalanced transport settings in the formulations of KP and GW.

\section{Keypoint-Guided Optimal Transport by Relation Preservation}\label{sec:kpg_ot}
This section details our proposed keypoint-guided model that leverages the keypoints to guide the matching in OT. The guidance is imposed by preserving the matching of keypoint pairs and the relation of each data point to the keypoints. We next discuss how to preserve the matching of keypoints, introduce the modeling of relation, present the keypoint-guided OT models, and discuss extension to KP and GW models in both balanced and unbalanced/partial transport settings. All the proofs of theorems or propositions are given in Appendix.

\subsection{Preservation of Matching of Keypoints in Transport}\label{sec:mask_modeling}
We denote the set of keypoint index pairs as $\mathcal{K}=\{(i_u,j_u)\}_{u=1}^U$ with $U$ denoting the number of paired keypoints.  We respectively denote $\mathcal{I} = \{i_u\}_{u=1}^U$ and $\mathcal{J} = \{j_u\}_{u=1}^U$ as the sets of source and target keypoint indexes. For the example illustrated in Fig.~\ref{fig:mask_modeling_main}, $\mathcal{K} = \{(3,2),(6,5)\}, \mathcal{I} = \{3,6\}$, and $\mathcal{J} = \{2,5\}$.
To impose the guidance of these keypoints with indexes in $\mathcal{K}$ in deriving the transport plan in OT, we first guarantee the exact matching of the keypoint pairs. As illustrated in Fig.~\ref{fig:mask_modeling_main}, we preserve the matching of keypoints in transport using a mask-based constraint of the transport plan, which is motivated by the following observation. If the paired keypoints  $(i,j)\in\mathcal{K}$ are matched, the optimal transport plan $\Tilde{\pi}$ satisfies that the $i$-th row and $j$-th column of 
$\Tilde{\pi}$ must be zeros except $\Tilde{\pi}_{i,j}$, which means that the all mass of source keypoint $x_i$ must be transported to target keypoint $y_j$ and $y_j$ can only receive the mass from $x_i$. For the example in Fig.~\ref{fig:mask_modeling_main}, the 3-th row and 2-th column are zeros except that $\Tilde{\pi}_{3,2}>0$. This sparsity of $\Tilde{\pi}$ motivates us to model it as the Hadamard product of a mask matrix $M=(M_{i,j})\in\mathbb{R}^{m\times n}$ and a matrix $\pi \in \mathbb{R}_+^{m\times n}$ with positive entries, \ie,
\begin{equation}\label{eq:mask_decom}
  \Tilde{\pi} = M \odot \pi, \mbox{ with } \Tilde{\pi}_{i,j} = M_{i,j}\pi_{i,j}.  
\end{equation}
{ Since ${\pi}_{i,j}$ is unconstrained when $M_{i,j}=0$, we enforce ${\pi}_{i,j}=0$ in these cases to prevent ill-posedness.
We define the admissible solution set for our keypoint-guided OT model as  
\begin{equation}\label{eq:mask_pi}
    \Pi(\bm{p},\bm{q};M) = \{ \pi \in \mathbb{R}^{m\times n}_+ \vert (M\odot\pi) \mathbbm{1}_{n} = \bm{p}, (M\odot\pi)^{\top} \mathbbm{1}_{m} = \bm{q}, (\mathbbm{1}_{m\times n}-M)\odot \pi = 0 \}.
\end{equation}
}The entry $M_{i,j}$ of the mask matrix $M$ is set to 0 if $\Tilde{\pi}_{i,j}$ needs to be 0, otherwise  $M_{i,j}$ is set to 1. Figure~\ref{fig:mask_modeling_main} illustrates the mask-based modeling of Eq.~\eqref{eq:mask_decom}. $M$ is constructed as in Proposition~\ref{thm:thm_mask_main}. 

\begin{proposition}\label{thm:thm_mask_main}
Suppose that the mask matrix $M$ satisfies that
\begin{equation}\label{eq:mask_def}
    M_{i,j}=\begin{cases} 1, & \mbox{ if } (i,j)\in\mathcal{K},\\
    0, & \mbox{ if }  i \in \mathcal{I}$ and $(i,j)\notin \mathcal{K},\\
    0, & \mbox{ if }  j \in \mathcal{J}$ and $(i,j)\notin \mathcal{K},\\
    1, &  \mbox{ otherwise (\ie, $i \notin \mathcal{I}$ and $j \notin \mathcal{J}$)}.
    \end{cases}
\end{equation} 
and $p_i = q_j$ for $(i,j)\in\mathcal{K}$. Then, the transport plan $\Tilde{\pi}=M\odot\pi$ with $\pi\in\Pi(\bm{p},\bm{q};M)$ preserves the matching of paired keypoints with index pairs in $\mathcal{K}$.
\end{proposition}
% More explanations of the construction of $M$ and the proof of Proposition~\ref{thm:thm_mask_main} are provided in Appendix~\ref{app:prop1}.
Proposition~\ref{thm:thm_mask_main} indicates that if $p_i = q_j$ for $(i,j)\in\mathcal{K}$, the matching of keypoint pairs is preserved by the mask-based constraint. 
{For the convenience of understanding to readers, we consider the case that $p_i = q_j$, $\forall (i,j)\in\mathcal{K}$. 
Note that the mask-based modeling in Eq.~\eqref{eq:mask_decom} is also applicable even for the case that there exist some $(i,j)\in\mathcal{K}$ such that $p_i \neq q_j$. For this case, we shall use a different mask matrix. We provide the discussion on such a more general scenario in Appendix~A. 
}

\subsection{Modeling the Relation to Keypoints}\label{sec:relation}
\begin{figure}[t]
	\centering
	\includegraphics[width=0.5\columnwidth]{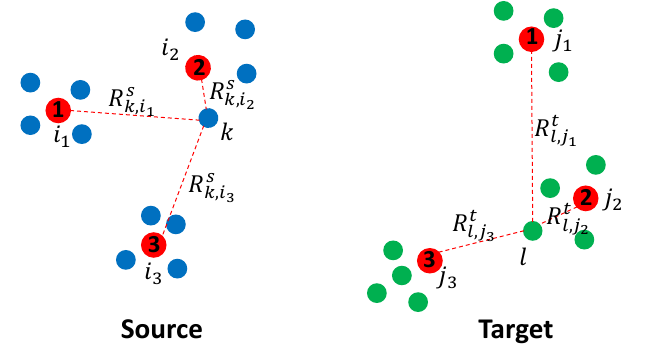} 
	\caption{ Illustration of relation of each point to the keypoints (red). In the keypoint-guided OT model, the relation should be preserved after transport.
    }
    \label{fig:relation}
    % \vspace{-1.1cm}
\end{figure}

To use the keypoints to guide the matching in transport, 
% our intuition is that the points of each domain near a paired keypoints should be matched in transport.
we propose to preserve the relation of each point to the set of keypoints in transport. Figure~\ref{fig:relation} illustrates the relation within points of each distribution. For the data point $x_k\in \bm{X}$,
its relation score
 to the keypoint $x_{i_u}$ for $i_u\in \mathcal{I}$ (illustrated by red circle in the left of  Fig.~\ref{fig:relation}) is defined as 
\begin{equation}\label{eq:r}
    R^s_{k,i_u} = \frac{e^{-C^s_{k,i_u}/\tau}}{\sum_{u'=1}^{U}e^{-C^s_{k,i_{u'}}/\tau}}, \mbox{ } \forall i_u \in \mathcal{I},
\end{equation}
% where $C^s_{k,i_u}$ is the squared $\chi^2$-distance between $x_k$ and $x_{i_u}$, and
where $\tau$ is temperature set as $\tau=\rho*\max_{i,k}\{C^s_{i,k}\}$ in experiments. Similarly, for $y_l\in\bm{Y}$, its relation score to the keypoint $y_{j_u}$ for $j_u\in\mathcal{J}$ is defined as 
\begin{equation}\label{eq:r_bar}
    R^t_{l,j_u} = \frac{e^{-C^t_{l,j_u}/\tau'}}{\sum_{u'=1}^{U}e^{-C^t_{l,j_{u'}}/\tau'}}, \mbox{ }\forall  j_u \in \mathcal{J},
\end{equation}
% where $C^t_{l,j_u}$ is the squared $\chi^2$-distance between $y_l$ and $y_{j_u}$, 
where $\tau'$ is temperature set to $\rho*\max_{j,l}\{C^t_{j,l}\}$. 
{In the definition of $\tau$ and $\tau'$, we first normalize the distances to $[0,1]$ by dividing by their maximum value, to increase the robustness of the relation score to the scale of the distances.\footnote{We normalize distances by their maximum value to reduce sensitivity to the overall distance scale, which we found effective empirically. Alternatively, one may normalize by a robust statistic such as the mean or median distance. This is an optional implementation choice and does not affect the method formulation.
} We then introduce $\rho$ to adjust the sharpness further.}
 $\rho$ is a tunable parameter set to 0.1 in our experiments, as 0.1 is a commonly used temperature in the softmax function~\citep{chen2020simple,khosla2020supervised}.
We denote $R^s_k = (R^s_{k,i_1},R^s_{k,i_2},\cdots, R^s_{k,i_U})$ and $R^t_l = (R^t_{l,j_1}, R^t_{l,j_2},\cdots, R^t_{l,j_U})$ that represent the relation of data points $x_k$ and $y_l$ to the keypoints in source and target domains, respectively. From the definition of the relation, the cross-domain points near to paired keypoints share a similar relation to the keypoints in corresponding domain. For instance, in Fig.~\ref{fig:relation}, $x_k$ and $y_l$ near to the paired keypoints with indexes $(i_2, j_2)$ have similar relation $R^s_k$ and $R^t_l$. Meanwhile, if $x_k$ is distant from all the keypoints, $R^s_k$ is close to a uniform probability vector.

% If $x_k$ is closest to  $x_{i_u}$ among the keypoints, $R^s_{k,i_u}$ is larger than the other components of $R^s_k$. If $x_k$ is distant from all the keypoints, $R^s_{k,i_u}$ is close to a uniform probability vector.

\subsection{Realizing Keypoint Guidance by Relation Preservation}\label{sec:kpg_rl}
{Based on the relation given above, we define the guiding matrix 
\begin{equation}\label{eq:guiding_matrix}
    G = (G_{k,l})\in\mathbb{R}^{m\times n} \mbox{ by } G_{k,l}=d(R^s_k,R^t_l),
\end{equation} where $d$ measures the dissimilarity of $R^s_k$ and $R^t_l$. $G_{k,l}$ is smaller if  $R^s_k$ and $R^t_l$ are similar.
$d$ is taken as the Jensen–Shannon divergence in this paper. We will study the effect of $d$ in experiments.
}
% , because $R^s_k$ and $R^t_l$ are probability vectors. 
By using the mask-based constraint of transport plan, the \textit{KeyPoint-Guided model by ReLation preservation} (\textbf{KPG-RL})
% \footnote{Given the defined guiding matrix $G$, one may argue that we can remove the keypoints and the mask in Eq.~\eqref{eq:guiding_matrix}. Then, Eq.~\eqref{eq:guiding_matrix} has the formulation of KP in Eq.~\eqref{eq:ot}, so that the solution algorithms for KP can be directly applied. Finally, the matching of keypoints are added to the optimized transport plan, obtaining the full transport plan. The mask-based modeling makes KPG-RL a principled model that automatically enforces the matching of keypoints in searching for the transport plan. With the mask-based modeling, our model provides a principled loss/metric to align or optimize distributions, as discussed in Sect.~\ref{sec:extension_kp_gw}. We will also discuss the necessity of the mask-based modeling of the transport plan in Sect.~\ref{sec:extension_kp_gw}. }
% If we remove the matching of keypoints and remove the mask in Eq.~\eqref{eq:guiding_matrix}, then Eq.~\eqref{eq:guiding_matrix} has the formulation of KP in Eq.~\eqref{eq:ot} but using cost $G$ and can be solved like KP. However, if removing the keypoints, we may face challenges. For example, in our developed keypoint-guided Gromov-Wasserstein model in Eq.~\eqref{eq:KPG_RL_GW}, the pairwise distances between all data points are required. By removing the keypoints, such pairwise distances can not be defined on all the data points. Moreover, the mask-based modeling could be a general framework by which we can enforce the other structures in the transport plan, \eg, preserving the matching between groups of points, by defining certain mask matrices. 
is defined as
\begin{equation}\label{eq:kpg}
    \min_{\pi \in \Pi(\bm{p},\bm{q};M)} L_{kpg}(\pi) \triangleq  \langle M\odot\pi , G \rangle_F.
\end{equation}
By the KPG-RL model in Eq.~\eqref{eq:kpg}, first, the matching of keypoint pairs is enforced by the mask-based constraint of the transport plan. 
{Second, the minimization of the objective function enforces that
% that the relation of each point to the keypoints is preserved after transport. 
% By the relation preservation model, i.e., KPG-RL model in Eq. (9), 
the optimal transport plan has larger entries in the locations where the entries of $G$ are smaller. Hence the cross-domain points corresponding to these locations (\eg, $k$ and $l$ shown in Fig.~\ref{fig:relation}) that are near to paired keypoints tend to be matched. Based on the softmax-based formulations in Eqs.~\eqref{eq:r} and~\eqref{eq:r_bar}, $d(R_k^s,R^t_l)$ is mainly determined by the relation score to the closest keypoint(s), since relation scores to the distant keypoints are small or close to 0. This implies that the points are mainly guided by the closest keypoints in our KPG-RL model in Eq.~\eqref{eq:kpg}. 
For points distant from all keypoints, their corresponding entries of $G$ are similar (close to zero) according to the definition of $G$. Therefore, the guidance of keypoints to these points is limited. To achieve correct matching of these points, additional information, \eg, point-wise cost or more keypoints, is needed.
}

{ Note that given the defined guiding matrix $G$, one may argue that we can remove the keypoints and the mask in Eq.~\eqref{eq:guiding_matrix}. Then, Eq.~\eqref{eq:guiding_matrix} has the formulation of KP in Eq.~\eqref{eq:ot}, so that the solution algorithms for KP can be directly applied. Finally, the matching of keypoints are added to the optimized transport plan, obtaining the full transport plan. Even though such a strategy could work for searching for the transport plan, the mask-based modeling is necessary in our model. Specifically, the mask-based modeling makes KPG-RL a principled model that automatically enforces the matching of keypoints in searching for the transport plan. With the mask-based modeling, our model provides a principled loss/metric to compare, align, or optimize distributions, as discussed in Sect.~\ref{sec:extension_kp_gw}. Meanwhile, the mask-based modeling is a general framework by which we can enforce the other structures in the transport plan, \eg, preserving the matching between groups of points, by defining certain mask matrices. For such cases, it is hard to define the corresponding transport model if removing the mask. We will further discuss the necessity of the mask-based modeling of the transport plan in Sect.~\ref{sec:extension_kp_gw}.
}

\vspace{0.4\baselineskip}\textit{Solving KPG-RL.}
Equation~\eqref{eq:kpg} is a linear program and can be solved by linear programming algorithms, \eg, the Simplex algorithm. 
We give the details for reformulating Eq.~\eqref{eq:kpg} as the standard form of linear programming in Appendix~B. 
Since Sinkhorn's algorithm offers a lightspeed computation of the entropy-regularized OT~\citep{cuturi2013sinkhorn}, a natural question is that, can Sinkhorn's algorithm be applied to the KPG-RL model with entropy regularization? We give the positive answer, and the details of the deduction are given in Appendix B. The iterative formulas are 
\begin{equation}\label{eq:sinkhorn_main}
    \bm{u}^{(l+1)} = \frac{\bm{p}}{K\bm{v}^l}, \hspace{0.2cm}\bm{v}^{(l+1)} = \frac{\bm{q}}{K^{\top}\bm{u}^{(l+1)}},
\end{equation}
where $K = M\odot e^{-G/\epsilon}$, $\epsilon$ is the coefficient of entropy regularization. 
{The division operator used above is entry-wise. If the iteration stops at $(\bm{u}, \bm{v})$, the optimal transport plan is $\mbox{diag}(\bm{u})K\mbox{diag}(\bm{v})$. We also provide the more stable log-domain version of the Sinkhorn iteration in Appendix B. 
Note that since the mask-based modeling introduces zeros in $K$, it is not clear whether the iteration converges. The following proposition shows that for the mask defined in Proposition~\ref{thm:thm_mask_main}, the iteration converges. We also empirically validate the convergence in experiments, as discussed in Appendix~B.
\begin{proposition}\label{thm:convergence_sinkhorn}
% For the mask \(M\) defined in Proposition~\ref{thm:thm_mask_main}, suppose that \(p_i=q_j\) for every \((i,j)\in\mathcal K\) and that \(m=n\). Then the Sinkhorn iteration in Eq.~\eqref{eq:sinkhorn_main} converges.
Let \(M\in\{0,1\}^{m\times n}\) be a fixed mask. Suppose that \(\Pi(\bm{p},\bm{q};M)\neq\emptyset\), and that \(G_{ij}<\infty\) for every \((i,j)\) such that \(M_{ij}=1\). Let
$K=M\odot \exp(-G/\epsilon) \mbox{ and } \epsilon>0.$
Then the transport plans generated by the Sinkhorn iteration in Eq.~\eqref{eq:sinkhorn_main} converge to the unique minimizer of the entropy-regularized KPG-RL problem.
\end{proposition}
}
% Note that \citet{sinkhorn1967concerning} show that the Sinkhorn iteration process converges if $K$ has support, \ie, there exists at least one positive diagonal in $K$. Since permuting the indexes of data points does not change the empirical distribution, and so as Sinkhorn's algorithm, we can rearrange the data points such that paired keypoints share the same indexes. According to the definition of the mask in Eqs.~\eqref{eq:mask_def}, the main diagonal of $M$ and thus $K$ are positive, which means that $K$ has support. Therefore, the Sinkhorn iteration above converges, which is verified in experiments, as discussed in Appendix~B.

Note that in this KPG-RL model in Eq.~\eqref{eq:kpg}, the points in $\bm{X}$ and $\bm{Y}$ do not necessarily lie in the same space because we only need to compute the distance within each one of $\bm{X}$ and $\bm{Y}$. Therefore, the proposed KPG-RL model in Eq.~\eqref{eq:kpg} is applicable even when $\bm{p}$ and $\bm{q}$ are supported in different spaces. As mentioned in Sect.~\ref{sec:introduction}, the GW model is applicable for transport across different spaces. However, GW is a non-convex quadratic program and is often solved by the Frank-Wolfe algorithm, in which a KP-like problem needs to be solved at each iteration by Sinkhorn's algorithm or linear programming. Surprisingly, our KPG-RL model can be directly solved using Sinkhorn's algorithm or linear programming without additional iterations, as discussed above. 

\subsection{Introducing Keypoint to Kantorovinch and Gromov-Wassestein Models}\label{sec:extension_kp_gw}
We now discuss how to impose the keypoint guidance in KP and GW models. To impose the keypoint guidance in KP, we can add $L_{kpg}(\pi)$ as a regularization term to KP, obtaining the following \textbf{KPG-RL-KP} model:
\begin{equation}\label{eq:KPG_RL_KP}
    \min_{\pi \in \Pi(\bm{p},\bm{q};M)} \left\{\alpha L_{kp}(M\odot\pi)+(1-\alpha)L_{kpg}(\pi) =  \langle M\odot\pi , \alpha C + (1-\alpha) G \rangle_F\right\}, \alpha \in (0,1).
\end{equation}
The KPG-RL-KP can be solved by Sinkhorn's algorithm or linear programming, the same as the solution of KPG-RL model. Similarly, we define the \textbf{KPG-RL-GW} model in GW  by
\begin{equation}\label{eq:KPG_RL_GW}
  \min_{\pi \in \Pi(\bm{p},\bm{q};M)} \alpha L_{gw}(M\odot\pi)+(1-\alpha)L_{kpg}(\pi), \alpha \in (0,1).
\end{equation}
The KPG-RL-GW  model is solved using the Frank-Wolfe algorithm, of which the main operations are first computing the gradient and then projecting it to the solution set. For the continuity of reading, we give the details of this algorithm in Appendix B.
% (please refer to \citep{gu2022keypointguided} for the details).
% \footnote{In the $l$-th iteration, we first calculate the gradient $g=\nabla_{\pi^l}\left(\alpha L_{gw}(M\odot\pi)+(1-\alpha)L_{kpg}(\pi)\right)$ for current solution $\pi^{l}$. We then project the gradient by $\pi'=\min_{\pi \in \Pi(\bm{p},\bm{q};M)}\langle g,\pi\rangle_F$, and finally compute $\pi^{l+1}= \omega\pi^{l}+(1-\omega)\pi'$, where $\omega$ is obtained by line search in $[0,1]$. Details are in \cite{gu2022keypointguided}.}. 
The KPG-RL-KP/KPG-RL-GW  models take the advantages of KP/GW and KPG-RL models, and could be helpful, especially when the number of paired keypoints is small. For the coefficient $\alpha$, we simply set it to 0.5 in experiments. 

\subsubsection{Theoretical Properties}\label{sec:theoretical_pp}
We next discuss the theoretical properties of KPG-RL-KP and KPG-RL-GW. We will prove that the KPG-RL-KP model provides a proper distance (named ``keypoint-guided Wasserstein distance'') for distributions supported in the same space, and the KPG-RL-GW model provides a divergence (in the sense of isomorphism) for distributions in distinct spaces, under mild conditions. Since the discrete distributions $\bm{p}=\frac{1}{m}\sum_{i}^m\delta_{x_i}$ (resp. $\bm{q}=\frac{1}{n}\sum_{j}^n\delta_{y_j}$) are invariant to the permutation of $\{x_i\}_{i=1}^m$ (resp. $\{y_j\}_{j=1}^n$), we assume that any two paired keypoints across domains share the same index. Therefore, the index set of paired keypoints is $\mathcal{K}=\{(i_u,i_u)\}_{u=1}^U$. 
% We assume $p_{i_u}=q_{i_u}, \forall i_u,$ in this section. For the convenience of description, in this section, 
We denote 
% $M^{\bm{p}\bm{q}}$ as the mask matrix for transporting $\bm{p}$ to $\bm{q}$, and 
$\mathcal{P}_{\mathcal{I}}^{\mathcal{X}}$ as the set of discrete probability distributions supported on $m$ points in ground space $\mathcal{X}$ such that all distributions in $\mathcal{P}_{\mathcal{I}}^{\mathcal{X}}$ share the keypoint index set $\mathcal{I} = \{i_u\}_{u=1}^U$.

{\vspace{0.4\baselineskip}\textit{KPG-RL-KP providing a proper metric.}
For distributions supported in the same space, we make the following assumption.
\begin{assumption}\label{ass:assum1}
    For any $\bm{p}$ and $\bm{q}$ in $\mathcal{P}_{\mathcal{I}}^{\mathcal{X}}$, if $\bm{p}=\bm{q}$, then the associated paired keypoints $(x_{i_u}, y_{i_u})$ are equal, \ie, $x_{i_u}=y_{i_u}$, $\forall i_u\in\mathcal{I}$.
\end{assumption}
Assumption~\ref{ass:assum1} indicates that if the source and target distributions are identical, the two points in a keypoint pair should be the same point.  We then have the following theorem.
% For distributions supported in the same space, the annotated keypoints are ``correct'' means that if two distributions are equal or similar, their keypoints should be the same or similar correspondingly. This is reasonable because the keypoints are annotated points by humans for guiding the matching of other points. We provide a more formal definition of the ``correctness'' of keypoints below.

% \begin{definition}[``Correctness'' of keypoints]
%     For any $i_u\in\mathcal{I}$, we say that the keypoint pair $(x_{i_u}, y_{i_u})$ associated to distributions $\bm{p}$ and $\bm{q}$ in $\mathcal{P}_{\mathcal{I}}^{\mathcal{X}}$ is ``correct'' if the following condition holds: if $\bm{p}=\bm{q}$, we have $x_{i_u}=y_{i_u}$.
% \end{definition}
%  We denote 
% \begin{equation}\label{eq:s_krk}
%     \mathcal{S}_{krk}(\bm{p},\bm{q}) = \min_{\pi \in \Pi(\bm{p},\bm{q};M)}  \sum_{i,j}{M_{i,j}\pi_{i,j}(\alpha C_{i,j}+(1-\alpha)G_{i,j})}, % \mbox{ where }\alpha\in(0,1).
% \end{equation}
% where $\alpha\in(0,1)$.
\begin{theorem}[Keypoint-guided Wasserstein distance]\label{thm:kpg_rl_kp}
Suppose $c$ is a proper distance in space $\mathcal{X}$ and $d$ is a proper distance in probability simplex $\Sigma^{U-1}$. Let 
\begin{equation}\label{eq:s_krk}
    \mathcal{S}_{krk}(\bm{p},\bm{q}) = \min_{\pi \in \Pi(\bm{p},\bm{q};M)}  \sum_{i,j}{M_{i,j}\pi_{i,j}(\alpha C_{i,j}+(1-\alpha)G_{i,j})}, % \mbox{ where }\alpha\in(0,1).
\end{equation}
where $\alpha\in(0,1)$. Then, for any  $\bm{p}$ and $\bm{q}$ in $\mathcal{P}_{\mathcal{I}}^{\mathcal{X}}$ satisfying Assumption~\ref{ass:assum1}, $\mathcal{S}_{krk}(\bm{p},\bm{q})$ is a proper distance between $\bm{p}$ and $\bm{q}$.
\end{theorem}

According to Theorem~\ref{thm:kpg_rl_kp}, $\mathcal{S}_{krk}(\bm{p},\bm{q})$ is a proper distance. To be consistent with the Wasserstein distance, we name $\mathcal{S}_{krk}(\bm{p},\bm{q})$  ``keypoint-guided Wasserstein distance''.

\vspace{0.4\baselineskip}\textit{KPG-RL-GW providing a divergence.}
For any distribution $\bm{p}\in\mathcal{P}^{\mathcal{X}}_{\mathcal{I}}$ and $\bm{q}\in\mathcal{P}^{\mathcal{Y}}_{\mathcal{I}}$, $\bm{p}$ and $\bm{q}$ are said to be isomorphic if there exists a bijection $\sigma:[m]\longmapsto[m]$ such that $c(x_i,x_k)=c'(y_{\sigma(i)},y_{\sigma(k)})$, and $p_i=q_{\sigma(i)}$, where $[m]=\{1,2,\cdots,m\}$, and $c$ and $c'$ are respectively proper distances in spaces $\mathcal{X}$ and $\mathcal{Y}$. For distributions in different spaces, we make the following assumption.
\begin{assumption}\label{ass:assum2}
    For any $\bm{p}\in\mathcal{P}_{\mathcal{I}}^{\mathcal{X}}$ and $\bm{q}\in\mathcal{P}_{\mathcal{I}}^{\mathcal{Y}}$, if $\bm{p}$ and $\bm{q}$ are isomorphic with corresponding bijection $\sigma$, then the associated keypoint pair $(x_{i_u}, y_{i_u})$ satisfy $\sigma(i_u)=i_u$, $\forall i_u\in\mathcal{I}$.
\end{assumption}
% We define the ``correctness'' of paired keypoints as follows.
% \begin{definition}[``Correctness'' of keypoints]\label{def:key_gw}
%     For any $i_u\in\mathcal{I}$, we say that the keypoint pair $(x_{i_u}, y_{i_u})$ associated to distributions $\bm{p}\in\mathcal{P}_{\mathcal{I}}^{\mathcal{X}}$ and $\bm{q}\in\mathcal{P}_{\mathcal{I}}^{\mathcal{Y}}$ is ``correct'' if the following condition holds: if $\bm{p}$ and $\bm{q}$ are isomorphic with corresponding bijection $\sigma$, we have $\sigma(i_u)=i_u$.
% \end{definition}
Assumption~\ref{ass:assum2} means that if $\bm{p}$ and $\bm{q}$ are isomorphic, $\sigma$ maps each source keypoint to its paired target keypoint, \ie, $\sigma(i_u)=i_u$. We denote
\begin{equation}
\begin{split}
    \mathcal{S}_{krg}(\bm{p},\bm{q})
    = \min_{\pi \in \Pi(\bm{p},\bm{q};M)}  \sum_{i,j} &\Big[\alpha\Big(\sum_{k,l}(M\odot\pi)_{i,j}(M\odot\pi)_{k,l}|C^s_{i,k}-C^t_{j,l}|^2\Big)\\
    &+(1-\alpha)(M\odot\pi)_{i,j}G_{i,j}\Big], %\mbox{ where }\alpha\in(0,1).
\end{split}
\end{equation}
where $\alpha\in(0,1)$. We then have the following theorem.

\begin{theorem}\label{thm:kpg_rl_gw}
Suppose $c$ and $c'$ are proper distances in spaces $\mathcal{X}$ and $\mathcal{Y}$. Suppose $d$ is a divergence in probability simplex $\Sigma^{U-1}$. Then, for any $\bm{p}$  in $\mathcal{P}_{\mathcal{I}}^{\mathcal{X}}$ and $\bm{q}$ in $\mathcal{P}_{\mathcal{I}}^{\mathcal{Y}}$ satisfying Assumption~\ref{ass:assum2}, we have $\mathcal{S}_{krg}(\bm{p},\bm{q}) = 0$ if and only if $\bm{p}$ and $\bm{q}$ are isomorphic.
\end{theorem}

The proofs of Theorems~\ref{thm:kpg_rl_kp} and~\ref{thm:kpg_rl_gw} are given in Appendix C. With the metric or divergence properties, Theorems~\ref{thm:kpg_rl_kp} and~\ref{thm:kpg_rl_gw} provide reliability for utilizing keypoint-guided KP and GW to align or compare distributions.

\vspace{0.4\baselineskip}\textit{Further discussion on the necessity of mask-based modeling.}
Building on the mask-based modeling, we have proved that our proposed KPG-RL-KP provides a proper distance between distributions, \ie, the keypoint-guided Wasserstein distance. 
The keypoint-guided Wasserstein distance $\mathcal{S}_{krk}(\bm{p},\bm{q})$ can be written as 
$\mathcal{S}_{krk}(\bm{p},\bm{q}) =  \sum_{i \in \mathcal{I}, j \in \mathcal{J}}M_{i,j}\pi_{i,j}(\alpha C_{i,j}+(1-\alpha)G_{i,j}) + \sum_{i \notin \mathcal{I}, j \notin \mathcal{J}}\pi_{i,j}(\alpha C_{i,j}+(1-\alpha)G_{i,j}).
$
If removing the mask-based modeling and the keypoints in our model, the corresponding model becomes 
$
    \mathcal{S}^{/M}_{krk}(\bm{p},\bm{q}) = \sum_{i \notin \mathcal{I}, j \notin \mathcal{J}}\pi_{i,j}(\alpha C_{i,j}+(1-\alpha)G_{i,j}),
$
it is no longer a proper metric because $\mathcal{S}^{/M}_{krk}(\bm{p},\bm{q})=0$ does not imply $\bm{p}=\bm{q}$ (the paired keypoints could be different).
 
Being a proper metric, the keypoint-guided Wasserstein distance $\mathcal{S}_{krk}(\bm{p},\bm{q})$ can be utilized to compare distributions or as a principled loss to learn distributions in generative models~\citep{arjovsky2017wasserstein,genevay2018learning}, multi-label learning~\citep{NIPS2015_a9eb8122}, \textit{etc}. 
When using $\mathcal{S}_{krk}(\bm{p},\bm{q})$ as a distribution distance metric, e.g., in learning distribution transform in a generative model, the first term over the paired keypoints will be imposed as guidance to learn the distribution transform guaranteeing the correspondence of keypoints, while preserving the relations to these keypoints. However, if removing the keypoint correspondence in $\mathcal{S}_{krk}(\bm{p},\bm{q})$ and posing the keypoint correspondence as a follow-up step, the learning of the distribution transform will be separated as two steps, i.e., first learning distribution transform over unpaired data by $\mathcal{S}^{/M}_{krk}$, followed by consistency between paired keypoints. On the contrary, $\mathcal{S}_{krk}$ learns the transform in a principled way. { Although practical implementations (e.g., Sinkhorn’s algorithm) may involve approximations, the metric structure imposes a rigorous geometric prior at the model level, defining a mathematically well-posed objective and providing a principled foundation for the learning process.}
}
\subsection{Extension to Partial and Unbalanced Transport Settings}\label{sec:ex} 
We now extend the above KPG-RL model to the more practical partial OT and unbalanced OT settings that allows to transport only partial mass. 
\subsubsection{Extension to Partial Transport Setting}\label{sec:ex_partial}
To develop the keypoint-guided partial transport model, we first apply the above mask-based constraint of transport plan to the partial OT model in Eq.~\eqref{eq:pot}.
% However, this strategy may partially transport the mass of keypoints and thus does not preserve the matching of keypoints. 
We then add more constraints to enforce that all the mass of keypoints is transported to preserve the matching of keypoints. Finally, we define the \textbf{partial KPG-RL} model as 
\begin{equation}\label{eq:partial_kpp_main}
\min_{\pi \in \Pi^{s}(\bm{p},\bm{q};M)}\left\{L_{kpg}(M\odot\pi) = \langle M\odot\pi , G \rangle_F\right\},
\end{equation}
where $\Pi^{s}(\bm{p},\bm{q};M) = \{ \pi \in \mathbb{R}^{m\times n}_+ \vert (M\odot\pi) \mathbbm{1}_{n} \leqslant \bm{p}, (M\odot\pi)^{\top} \mathbbm{1}_{m} \leqslant \bm{q}, \mathbbm{1}_{m}^{\top}(M\odot\pi) \mathbbm{1}_{n} = s; (M\odot\pi)_{i,:}\mathbbm{1}_n = p_i, \forall i \in \mathcal{I}; \mathbbm{1}^{\top}_m(M\odot\pi)_{:,j} = q_j, \forall j \in \mathcal{J}; {(\mathbbm{1}_{m\times n}-M)\odot\pi=0}\}.$ 

\begin{figure}[t]
	\centering
	\includegraphics[width=0.3\columnwidth]{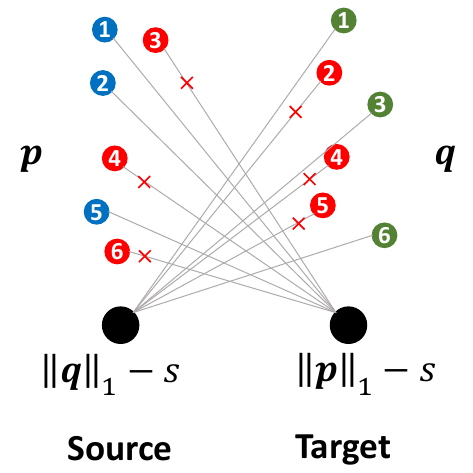} 
	\caption{ Illustration of dummy points (black circles) for source and target domains. }
    \label{fig:partial_pkg}
\end{figure}
\citet{chapel2020partial} propose a compelling method to solve the original partial OT problem in Eq.~\eqref{eq:pot} by transforming the partial OT problem into a KP-like problem.
Inspired by ~\cite{chapel2020partial}, we add a dummy point with mass $\Vert\bm{q}\Vert_1-s$ for source domain (the left black circle in Fig.~\ref{fig:partial_pkg}) and a dummy point with mass $\Vert\bm{p}\Vert_1-s$ for target domain (the right black circle in Fig.~\ref{fig:partial_pkg}). We denote $\bar{\bm{p}} = (\bm{p}^{\top},\Vert\bm{q}\Vert_1-s)^{\top}$ and $\bar{\bm{q}} = (\bm{q}^{\top},\Vert\bm{p}\Vert_1-s)^{\top}$. 
As illustrated in Fig.~\ref{fig:partial_pkg}, we aim to design the extended guiding matrix $\bar{G}$ and extended mask matrix $\bar{M}$ such that performing KPG-RL between $\bar{\bm{p}}$ and $\bar{\bm{q}}$ will transport $\Vert\bm{p}\Vert_1-s$ mass from source real data points to the target dummy point and transport $\Vert\bm{q}\Vert_1-s$ mass from source dummy point to target real data points. As a sequence, only $s$ mass of source and target real data points are matched.
Meanwhile, the keypoints should not be matched to the dummy points because they are annotated data to guide the matching.
% Inspired by \citet{chapel2020partial}, to solve our partial-KPG-RL model in Eq.~\eqref{eq:partial_kpp_main}, we add a dummy point to each of the marginal distributions to transform Eq.~\eqref{eq:partial_kpp_main} into a KPG-RL-like problem. 
To do this, we extend $G, M$ by
\begin{equation*}
% \bar{\bm{p}} = \begin{bmatrix}
% \bm{p} \\
%  \Vert\bm{q}\Vert_{1} - s
% \end{bmatrix},
% % (\bm{p}, \Vert \bm{q}\Vert_{1} - s),
% \bar{\bm{q}} = \begin{bmatrix}
% \bm{q} \\ \Vert \bm{p}\Vert_{1} - s
% \end{bmatrix},
\bar{G} =
\begin{bmatrix}
G & \xi \mathbbm{1}_{n} \\
\xi \mathbbm{1}_{m}^{\top} & 2\xi + A
\end{bmatrix},
\bar{M} = \begin{bmatrix}
M & \bm{a} \\
\bm{b}^{\top} & 1
\end{bmatrix},
\end{equation*}
where 
% $A\in\mathbb{R},\xi\in\mathbb{R}$ 
 $A>0$ and $\xi$ are two fixed scalars, and $ \bm{a}\in \mathbb{R}^m, \bm{b}\in \mathbb{R}^n$. The element $a_i$ of $\bm{a}$ is 0 if $i\in \mathcal{I}$, otherwise 1.  The element $b_j$ of $\bm{b}$ is 0 if $j\in \mathcal{J}$, and 1 otherwise. 
% For more motivations of this extension, please refer to Appendix~\ref{app:motivation_partial_kpg}.
By the following theorem, solving the optimal transport plan of problem~\eqref{eq:partial_kpp_main} boils down to
solving the problem  $\mbox{min}_{\bar{\pi} \in \Pi(\bar{\bm{p}},\bar{\bm{q}};\bar{M})} \langle \bar{M}\odot\bar{\pi} , \bar{G} \rangle_F$.

\begin{theorem}\label{thm:partial_kpp_main}
Suppose $A>0$, $\sum_{i\in \mathcal{I}} p_i <s$, and $\sum_{j\in \mathcal{J}} q_j < s$, then the optimal transport plan $M\odot\pi^*$ of problem~\eqref{eq:partial_kpp_main} is the $m$-by-$n$ block in the upper left corner of the optimal transport plan $\bar{M}\odot\bar{\pi}^*$ of problem $\mbox{min}_{\bar{\pi} \in \Pi(\bar{\bm{p}},\bar{\bm{q}};\bar{M})} \langle \bar{M}\odot\bar{\pi}, \bar{G} \rangle_F$.
\end{theorem}

According to Theorem~\ref{thm:partial_kpp_main}, problem~\eqref{eq:partial_kpp_main} is solved as follows. We fisrt solve the problem  $\mbox{min}_{\bar{\pi} \in \Pi(\bar{\bm{p}},\bar{\bm{q}};\bar{M})} \langle \bar{M}\odot\bar{\pi} , \bar{G} \rangle_F$ by linear programming or Sinkhorn's algorithm as discussed in Sect.~\ref{sec:kpg_rl}, and then take the $m$-by-$n$ block in the upper left corner of the corresponding optimal transport plan as the optimal transport plan for partial KPG-RL in Eq.~\eqref{eq:partial_kpp_main}.

{\vspace{0.4\baselineskip}\textit{Developing partial keypoint-guided GW model.}
We now develop the partial keypoint-guided GW model. 
By replacing the solution set $\Pi(\bm{p},\bm{q};M)$ in KPG-RL-GW model (Eq.~\eqref{eq:KPG_RL_GW}) with the solution set $\Pi^s(\bm{p},\bm{q};M)$, the partial KPG-RL-GW model is defined as 
\begin{equation}\label{eq:prtial_KPG_RL_GW}
  \min_{\pi \in \Pi^s(\bm{p},\bm{q};M)} \alpha L_{gw}(M\odot\pi)+(1-\alpha)L_{kpg}(\pi), \alpha \in (0,1).
\end{equation}
Inspired by \citep{chapel2020partial}, the partial KPG-RL-GW model is solved using the Frank-Wolfe algorithm, which solves a partial KPG-RL-like problem at each iteration. Details are given in Appendix B.

% Please refer to Appendix~\ref{app:proof_thm1} for the proof of Theorem~\ref{thm:partial_kpp_main}.
\subsubsection{Extension to Unbalanced Transport Setting}\label{sec:ex_unbalanced}
To develop the keypoint-guided unbalanced optimal transport, we apply the mask-based modeling of the transport plan and the relation preservation objective function to the unbalanced optimal transport model in Eq.~\eqref{eq:uot}. To further ensure the transport of mass of keypoints in unbalanced setting, we employ a spatially varying KL-divergence as regularization. 
The \textbf{unbalanced KPG-RL} model is defined as 
\begin{equation}\label{eq:unbalanced_kpg_rl}
\min_{\pi\geq 0, (\mathbbm{1}_{m\times n}-M)\odot\pi=0 }L_{kpg}(\pi) + \mu \bar{\mathcal{D}}((M\odot\pi)\mathbbm{1}_{n}|\bm{p}) + \mu \bar{\mathcal{D}}((M\odot\pi)^\top\mathbbm{1}_{m}|\bm{q}),
\end{equation}
where $\bar{\mathcal{D}}(\bm{r}|\bm{p})=\sum_ia_ir_i\log\frac{r_i}{p_i}$ is the spatially varying KL-divergence. $a_i=\infty$ if $i$ is an index of keypoints of distribution $\bm{p}$, otherwise 1. We can see that $r_i=p_i$ ($i$ is an index of keypoints) as long as $\bar{\mathcal{D}}(\bm{r}|\bm{q})<\infty$.  The spatially varying KL-divergence ensures that the keypoints are transported in the unbalanced KPG-RL model. $\mu$ is simply set to 1 in experiments.

According to~\citep{sejourne2023unbalanced}, we solve the unbalanced KPG-RL model with entropy regularization by the generalized Sinkhorn's algorithm. The iteration formulation is given by 
\begin{equation}\label{eq:generalized_sinkhorn}
    \bm{u}^{(l+1)} = \left(\frac{\bm{p}}{K\bm{v}^l}\right)^{\bm{\rho}}, \hspace{0.2cm}\bm{v}^{(l+1)} = \left(\frac{\bm{q}}{K^{\top}\bm{u}^{(l+1)}}\right)^{\bm{\rho}},
\end{equation}
where the power operator is entry-wise. The element $\rho_i$ of $\bm{\rho}$ is 1 if $i\in \mathcal{I}$, otherwise $\frac{\mu}{\mu+\epsilon}$, where $\epsilon$ is the coefficient of entropy regularization.

\vspace{0.4\baselineskip}\textit{Developing unbalanced kepoint-guided GW model.}
To develop the keypoint-guided unbalanced Gromov-Wasserstein model, \ie, unbalanced KPG-RL-GW, we apply the quadratic divergence as in unbalanced GW~\citep{sejourne2021unbalanced} to KPG-RL-GW. The unbalanced KPG-RL-GW is defined by 
\begin{equation}
    % \min_{\pi} \alpha L_{gw}(\tilde{\pi}) + (1-\alpha)L_{kpg}(\pi) + \mu \bar{\mathcal{D}}^{\otimes}(\tilde{\pi}_1|\bm{p}) + \mu \bar{\mathcal{D}}^{\otimes}(\tilde{\pi}_2|\bm{q})
    \min_{{\pi\geq 0, (\mathbbm{1}_{m\times n}-M)\odot\pi=0 }} \alpha L_{gw}(\tilde{\pi}) + (1-\alpha)L_{kpg}(\pi) + \mu \bar{\mathcal{D}}^{\otimes}(\tilde{\pi}_1\times\tilde{\pi}_1|\bm{p}\times\bm{p}) + \mu \bar{\mathcal{D}}^{\otimes}(\tilde{\pi}_2\times\tilde{\pi}_2|\bm{q}\times\bm{q})
\end{equation}
where $\tilde{\pi}=M\odot\pi$, $\tilde{\pi}_1 = \tilde{\pi}\mathbbm{1}_n$, and $\tilde{\pi}_2 = \tilde{\pi}^{\top}\mathbbm{1}_m$. $\bar{\mathcal{D}}^{\otimes}(\bm{r}\times\bm{r}'|\bm{p}\times\bm{p})=\sum_{i,j}a_{i,j}r_ir'_j\log\frac{r_ir'_j}{p_ip_j}$ is the spatially varying quadratic KL-divergence, where $a_{i,j}=\infty$ if both $i$ and $j$ are indexes of keypoints of distribution $\bm{p}$, otherwise 1. The unbalanced KPG-RL-GW model is solved similarly to the unbalanced GW model~\citep{sejourne2021unbalanced} with entropic regularization, in which an unbalanced KPG-RL-like problem is solved at each iteration. Details are given in Appendix B.
}

{
\section{Learning Neural Transport for Keypoint-Guided Optimal Transport}\label{sec:large_scale}
In Sect.~\ref{sec:kpg_ot}, we have discussed the keypoint-guided optimal transport model in primal formulation (\ie, Eq.~\eqref{eq:kpg}) that is solved by linear programming or Sinkhorn's algorithm. Though Sinkhorn's algorithm is promising, it faces challenges in terms of time and memory cost when the number of data points is larger. As mentioned in \citep{seguy2017large}, each iteration of Sinkhorn's algorithm has $\mathcal{O}(n^2)$ complexity. Imagine that if $n$ is a large number, \eg, $n>10^5$ or even larger values, there are at least $10^{10}$ variables to be optimized, and the computational time cost would be extremely high. Meanwhile, storing such high-dimensional matrices poses challenges to ordinary computers ({see Table~\ref{tab:scaling_comparison} in Appendix G}). 

Considering that deep learning has shown success in tackling large numbers of data points, in this section, we will develop a deep-learning-based approach to estimate the transport plan and map based on the dual formulation of KPG-RL. 
% Our approach is inspired by \citep{seguy2017large} that studies the KP. 
% \citet{seguy2017large} start with the dual problem of the entropy or $\chi^2$-regularized KP model, which is solved by the training of deep learning. Based on the learned transport plan, they propose the Barycentric Projection (BP) to transport source data to the target domain. 
Inspired by~\citet{seguy2017large}, we will first develop the dual formulation of the $\chi^2$-regularized KPG-RL model and solve it by training deep neural networks. Based on the learned transport plan, we further propose two novel neural transport strategies, namely Manifold Barycentric Projection (MBP) and Manifold SamPling (MSP), to transport source samples to the target domain.

% \citet{seguy2017large} propose a promising approach to estimate the transport plan and map based on deep learning techniques. They considered the dual problem of the entropy or $\chi^2$-regularized KP model, which is solved by the training of deep learning. Based on the learned transport plan, they propose the Barycentric Projection (BP) to transport source data to the target domain. Inspired by~\citet{seguy2017large}, we employ deep learning techniques to solve our keypoint-guided optimal transport model in this section.  We first develop the dual formulation of the regularized KPG-RL model and solve it through the training of deep learning. Based on the learned transport plan, we further propose a novel Manifold Barycentric Projection (MBP) to transport source samples to the target domain, for tackling the issue of the blur of transported samples by BP. 

% which take $\mathcal{O}((mn)^3)$ and $\mathcal{O}(mn\log(mn))$ time cost, respectively.  The linear programming and Sinkhorn iteration do not scale well to distributions supported on a large number of samples ($m$ and $n$ are large). We next discuss how to extend our keypoint-guided optimal transport model to large-scale transport problems in this section. Inspired by~\citep{seguy2017large,daniels2021score} that extend the KP to large-scale transport problems, we first deduce the dual formulation of the regularized KPG-RL model, based on which a GAN-based approach is proposed to transport the source samples to the target domain.

\subsection{Dual Formulation of ${\chi}^2$-regularized KPG-RL}\label{sec:dual}
The $\chi^2$-regularization has been investigated in conventional OT~\citep{seguy2017large,blondel2018smooth}, of which the duality is an unconstrained optimization problem friendly to deep learning. We apply the $\chi^2$-regularization to our KPG-RL model. 
% There are mainly two types of regularization, \ie, entropy regularization~\citep{cuturi2013sinkhorn} and $\chi^2$-regularization~\citep{seguy2017large,blondel2018smooth} in OT model. We adopt the $\chi^2$-regularization for our purpose because we find that the $\chi^2$-regularization makes the training more stable than the entropy regularization for the smaller regularization coefficient in experiments. The dual formulation of the entropy-regularized KPG-RL can be deduced analogously, left as our future work. 
The $\chi^2$-regularized KPG-RL model is given by
\begin{equation}\label{eq:l2_reg_kpg}
    \min_{\pi\in\Pi(\bm{p},\bm{q};M)} \langle M\odot\pi, G \rangle_F + \epsilon \chi^2(M\odot\pi\Vert \bm{p} \times \bm{q}),
\end{equation}
where $\chi^2(M\odot\pi\Vert \bm{p} \times \bm{q}) = \sum_{i=1}^{m}\sum_{j=1}^{n}\left(\frac{M_{i,j}\pi_{i,j}}{p_iq_j}\right)^2p_iq_j$. The following theorem provides the dual formulation of problem~\eqref{eq:l2_reg_kpg}.

\begin{theorem}\label{thm:dual}
    The $\chi^2$-regularized KPG-RL model in Eq.~\eqref{eq:l2_reg_kpg} has the dual formmulation 
    \begin{equation}\label{eq:dual}
        \max_{\phi,\psi} \sum_{i=1}^m \phi(x_i)p_i + \sum_{j=1}^n \psi(y_j)q_j - \frac{1}{4\epsilon}\sum_{i,j}M_{i,j}[(\phi(x_i)+\psi(y_j)-G_{i,j})_+]^2p_iq_j,
    \end{equation}
    where $a_+=\max\{a,0\}$.
    Let $\phi^*$, $\psi^*$ be the optimal solution of Eq.~\eqref{eq:dual}, then the optimal transport plan $M\odot\pi^*$ of Eq.~\eqref{eq:l2_reg_kpg} satisfies
    \begin{equation}\label{eq:prim_dual_solution}
        (M\odot\pi^*)_{i,j} = \frac{1}{2\epsilon}M_{i,j}(\phi^*(x_i)+\psi^*(y_j)-G_{i,j})_+p_iq_j.
    \end{equation}
\end{theorem}
Equation~\eqref{eq:dual} is an unconstrained optimization problem \textit{w.r.t.} continuous functions $\phi,\psi$. We parameterize $\phi,\psi$ using neural networks $\phi_{\theta},\psi_{\theta}$ respectively and utilize mini-batch stochastic gradient descent to optimize $\theta$. Specifically, in each iteration, we sequentially sample mini-batch samples from $\bm{p},\bm{q}$, calculate the objective function in Eq.~\eqref{eq:dual} on the mini-batch samples, compute gradient \textit{w.r.t.} $\theta$ using backpropagation, and update $\theta$ by gradient descent. Such a mini-batch-based optimization process enables the keypoint-guided optimal transport model to scale to the distributions supported on a larger number ($m$ and $n$) of samples, because we only need a mini-batch of samples to calculate the objective function in each iteration. Once the optimization process is completed, the optimal transport plan is recovered by the strong duality in Eq.~\eqref{eq:prim_dual_solution}. {It is worth noting that due to the inevitable optimization gap in non-convex network training and the finite capacity of network approximations, the recovered plan $(M \odot \pi^*)$ may not satisfy marginal constraints.  Despite this, $(M \odot \pi^*)$ reflects the coupling relationship between source and target domain samples. We next treat the recovered values in Eq. \eqref{eq:prim_dual_solution} as unnormalized coupling weights to build neural transport strategies for transporting source domain data to target domain.}

\subsection{Neural Transport Strategies}\label{sec:MBP}
Section~\ref{sec:dual} offers a neural approach to learn the transport plan for larger numbers of data points. Yet it is unclear how to transport source data to target domain. \citet{seguy2017large} propose the Barycentric Projection (BP) to transport the source data to the target domain, which can be directly applied to our approach using our learned transport plan. The BP ($T_{BP}$) is given by
\begin{equation}\label{eq:barycentric_projection}
    T_{BP} = \arg\min_{T'}\sum_{i=1}^{m}\sum_{j=1}^{n}(M\odot\pi^*)_{i,j} \bar{d}(y_j,T'(x_i)),
\end{equation}
where $\bar{d}$ is a difference measure. 
% Note that given a source sample $x$ even outside the support set of $\bm{p}$, the barycentric projection $T_{BP}$ can be applied to transport $x$ to the target domain. \citet{seguy2017large} show that when the size $m$ and $n$ of support samples is large and the regularization coefficient $\epsilon$ is small, the barycentric projection is close to the Monge map between the underlying continuous distributions.
However, for the source sample $x$, the transported sample $T_{BP}(x)$ by BP may be blurry, as shown in Fig.~\ref{fig:BP}. This is mainly because $T_{BP}(x)$ is close to the barycenter under $\bar{d}$.
% the smaller $\epsilon$ makes the learning of $\phi$ and $\psi$ difficult  due to numerical issues, see Eq.~\eqref{eq:dual}. While for larger $\epsilon$, the optimal transport plan is not sparse, leading to the blur of the image (as shown in Fig.~\ref{fig:BP}) of the barycentric projection because it is close to the weighted average of target samples, see Eq.~\eqref{eq:barycentric_projection}.

\begin{figure}[t]
	\centering
	\includegraphics[width=1.0\columnwidth]{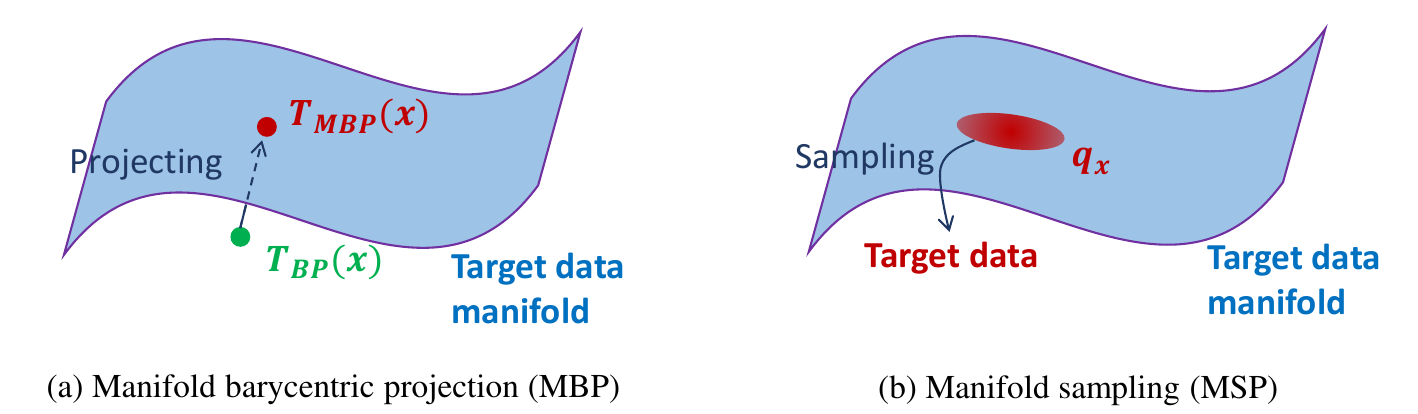}
	\caption{Illustration of our (a) manifold barycentric projection and (b) manifold sampling for transporting source data to the target domain. In (a), the image $T_{BP}(x)$ of barycentric projection is outside the target data manifold and is blurry. The transported data $T_{MBP}(x)$ by our proposed manifold barycentric projection is constrained to the target data manifold. In (b), the transport plan provides a conditional distribution $\bm{q}_{{x}}$ on target data manifold for a given source sample ${x}$. Our manifold sampling aims to sample target data from $\bm{q}_{{x}}$.}
	\label{fig:idea_gan_based_transport}
\end{figure}

We analyze that the blur caused by the BP is probably because the projected data $T_{BP}(x)$ may not be in the target data manifold.\footnote{Machine learning studies often take the ``manifold hypothesis'' that many high-dimensional real data sets, \eg, natural images, lie in low-dimensional manifolds inside the high-dimensional space~\citep{cayton2005algorithms,fefferman2016testing}.} Intuitively, if $\bar{d}$ can ideally reflect the intrinsic structure of the target data manifold, the projected data $T_{BP}(x)$ could be in the target data manifold. However, such an ideal $\bar{d}$ is difficult to define.   
To obtain clear transported data, we present two neural transport strategies, namely manifold barycentric projection and manifold sampling, to transport source data to target domain based on learned transport plan. We next discuss the details of the manifold barycentric projection and manifold sampling. 

\vspace{0.4\baselineskip}\textit{Manifold Barycentric Projection.} The goal of Manifold Barycentric Projection (MBP) is to enforce the transported data near to $T_{BP}(x)$ constrained into the target data manifold implicitly. Figure~\ref{fig:idea_gan_based_transport}(a) illustrates our idea. To achieve this goal, we add a manifold constraint to the BP model in Eq.~\eqref{eq:barycentric_projection}, forming the following model:
\begin{equation}\label{eq:gan_based}
T_{MBP}=\arg\min_{T} \lambda\sum_{i=1}^{m}\sum_{j=1}^{n}(M\odot\pi^*)_{i,j}\bar{d}(y_j,T(x_i)) + \mathcal{L}_{M}(T),
\end{equation}
where the first term enforces the transported data is near to that of the barycentric projection, and the second term encourages the transported data to lie in the target data manifold $\lambda$ is a hyper-parameter for balancing the importance of the two terms and is set to 10 in experiments. For simplicity, we use the loss of WGAN-GP~\citep{gulrajani2017improved} as $\mathcal{L}_{M}$, because WGAN-GP is often used to model the data manifold~\citep{pandey2021generalization}. $\mathcal{L}_{M}$ is given by 
\begin{equation}\label{eq:loss_gan}
    \mathcal{L}_{M}(T)=\max_D{\sum_{i=1}^m D(T(x_i))p_i -\sum_{j=1}^nD(y_j)q_j + \beta\mathbb{E}_{\hat{x}\sim \hat{\bm{p}}}(\Vert \nabla_{\hat{x}}D(\hat{x})\Vert_2-1)^2},
\end{equation}
where $\hat{\bm{p}}$ denotes the samples uniformly along lines
between pairs of points sampled from the transported source distribution $T_{\#}\bm{p}$ and target distribution $\bm{q}$, and $D$, named discriminator, is a neural network that outputs a scalar.\footnote{For a random variable $x\sim \bm{p}$ and a transform $T$ on $x$, we use $T_{\#}\bm{p}$ to denote the distribution of $T(x)$.} Following~\citet{gulrajani2017improved}, $\beta$ is set to 10, and the gradient of $\hat{x}$ \textit{w.r.t.} $T$ is stopped. Equation~\eqref{eq:gan_based} is optimized by the mini-batch stochastic gradient descent/ascent to update $T$/$D$, respectively. With the learned $T_{MBP}$, given the source domain test sample $x$ even outside the training set, we transport it to the target domain by $T_{MBP}(x)$.

\vspace{0.4\baselineskip}\textit{Manifold Sampling.}
For the Manifold SamPling (MSP), the goal is to sample data from the conditional distribution on the target domain data manifold induced by the transport plan, given the source sample. Specifically, given the learned transport plan $M\odot \pi$ and a source sample $x_i$, $M\odot \pi$ induces a empirical distribution $\bm{q}_{x_i}$ in target data manifold conditional on $x_i$, where $\bm{q}_{x_i}$ is defined by $\bm{q}_{x_i}=\frac{1}{\sum_{j'=1}^n (M\odot\pi^*)_{i,j'}}\sum_{j=1}^n(M\odot\pi^*)_{i,j}\delta_{y_j}$. The main idea of MSP is to sample data from $\bm{q}_{x_i}$, which is taken as the transported data of $x_i$, as illustrated in Fig.~\ref{fig:idea_gan_based_transport}(b). One possible strategy to achieve this goal is to choose a $y_j$ with probability $\frac{1}{\sum_{j'=1}^n (M\odot\pi^*)_{i,j'}}(M\odot\pi^*)_{i,j}$ from $\bm{Y}$. However, such a strategy can not generate new target samples outside the training set $\bm{Y}$. To address tackle this issue, inspired by GANs~\citep{Goodfellow2014GenerativeAN,gulrajani2017improved},  
we learn a map $T_{MSP}$ mapping source samples along with noise to the target data manifold, which is able to generate target samples outside $\bm{Y}$. The mathematical formulation for learning $T_{MSP}$ is defined by 
\begin{equation}\label{eq:msp}
     T_{MSP}=\arg\min_{T}\max_D{\sum_{i=1}^m p_i\left(\mathbb{E}_{z\sim \mathcal{N}} D(x_i,T(x_i,z)) -\mathbb{E}_{y\sim \bm{q}_{x_i}}D(x_i,y)\right) + \beta\mathcal{R}(D)},
\end{equation}
where $\mathcal{N}$ is the standard Gaussian distribution, and $\mathcal{R}(D)= \sum_{i=1}^m p_i\mathbb{E}_{\hat{x}\sim \hat{\bm{p}}}(\Vert \nabla_{\hat{x}}D(x_i,\hat{x})\Vert_2-1)^2$, where $\hat{\bm{p}}$ denotes the samples uniformly along lines between pairs of points sampled from $T(x_i,\cdot )_\#\mathcal{N}$ and $\bm{q}_{x_i}$. In Eq.~\eqref{eq:msp}, $D(x_i,\cdot)$ is a discriminator specific to $x_i$, distinguishing the transported samples $T(x_i,z)$ and target sample $y$ of conditional distribution $\bm{q}_{x_i}$. $T(x_i,\cdot)$ aims to fool $D(x_i,\cdot)$. In training, $T/D$ are updated using mini-batch gradient stochastic descent/ascent. As a result of the adversarial training of $T$ and $D$, $T(x_i,z)$ will approach a sample from $\bm{q}_{x_i}$.  After training, given a test source sample $x$ even outside of the training set, we randomly sample a Gaussian noise $z$, and then $T_{MSP}(x_i,z)$ are the transported data.
}

\section{Experiments}\label{sec:app_hda}
We evaluate our method on toy data, HDA, Multi-omic single-cell alignment, and I2I translation experiments. Codes will be available at \url{https://github.com/XJTU-XGU/KPG-RL}.

\subsection{Toy Experiments}
\begin{figure}[t]
	\centering
	\subfigure[KP]{ \includegraphics[width=0.31\columnwidth]{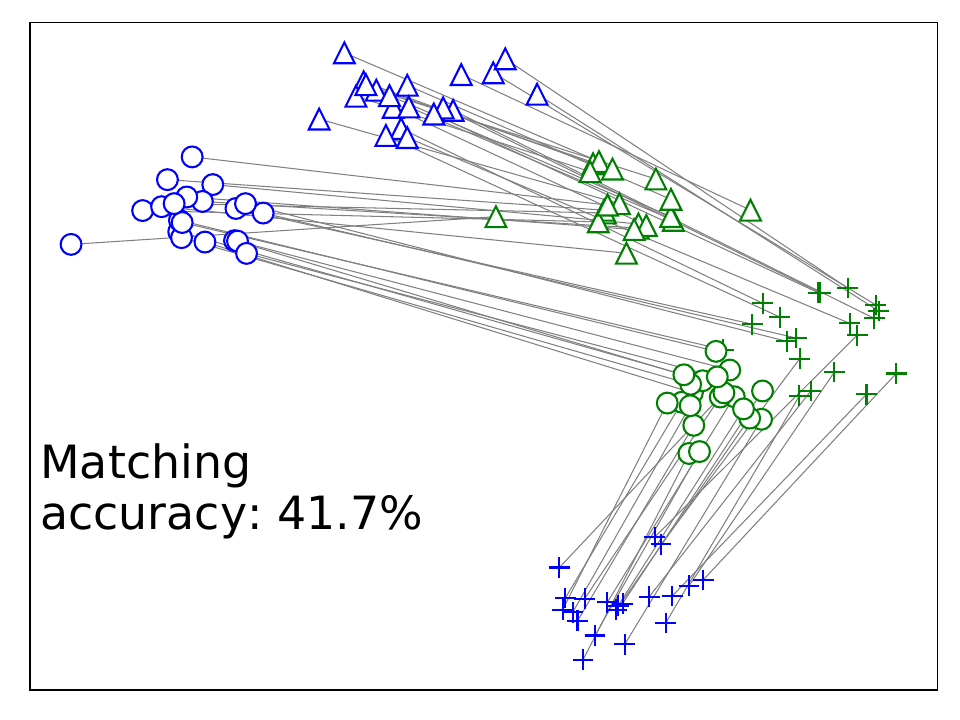} 
		\label{fig:toy_OT}}
	\subfigure[KPG-RL-KP  (w/ 2 keypoints)]{ \includegraphics[width=0.3\columnwidth]{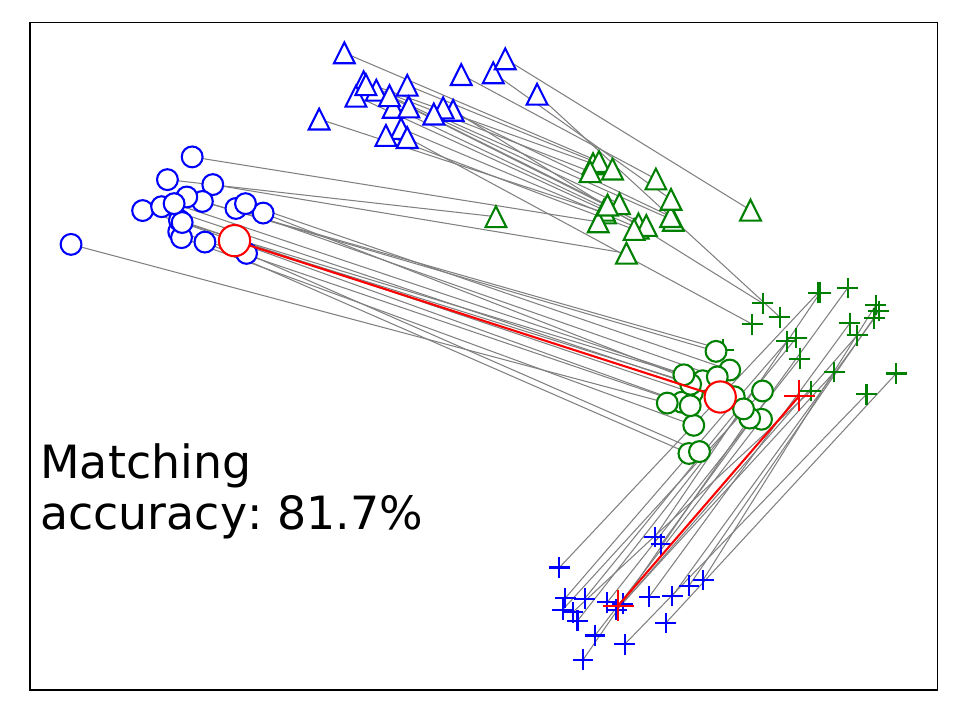} 
		\label{fig:toy_KPG_OT_21}}
	\subfigure[KPG-RL-KP  (w/ 3 keypoints)]{ \includegraphics[width=0.3\columnwidth]{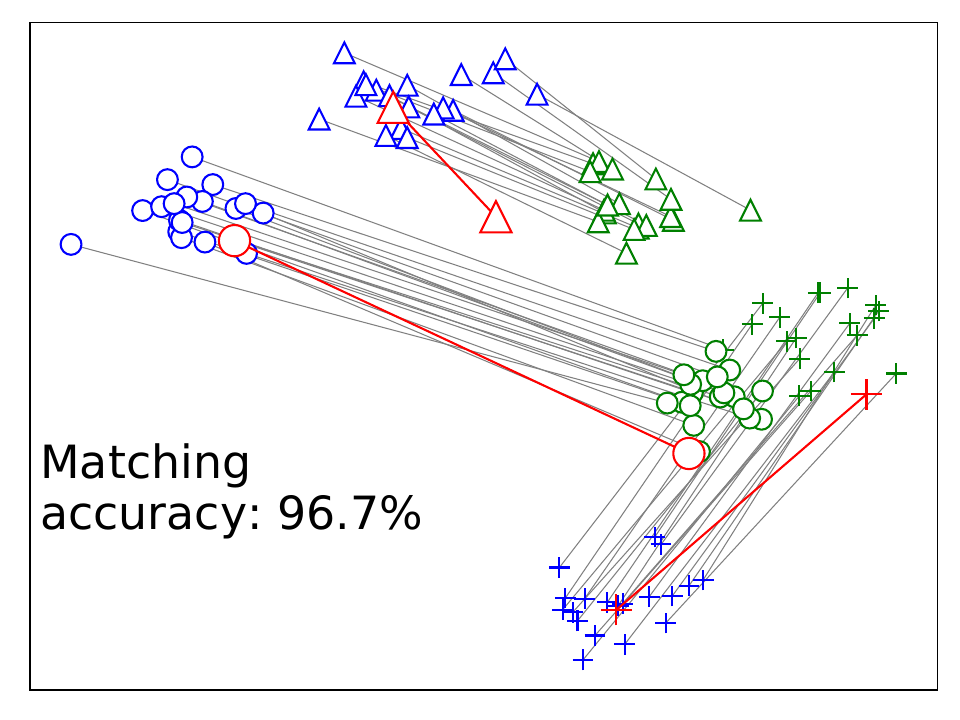} 
		\label{fig:toy_KPG_OT_3}}
	\subfigure[GW]{ \includegraphics[width=0.31\columnwidth]{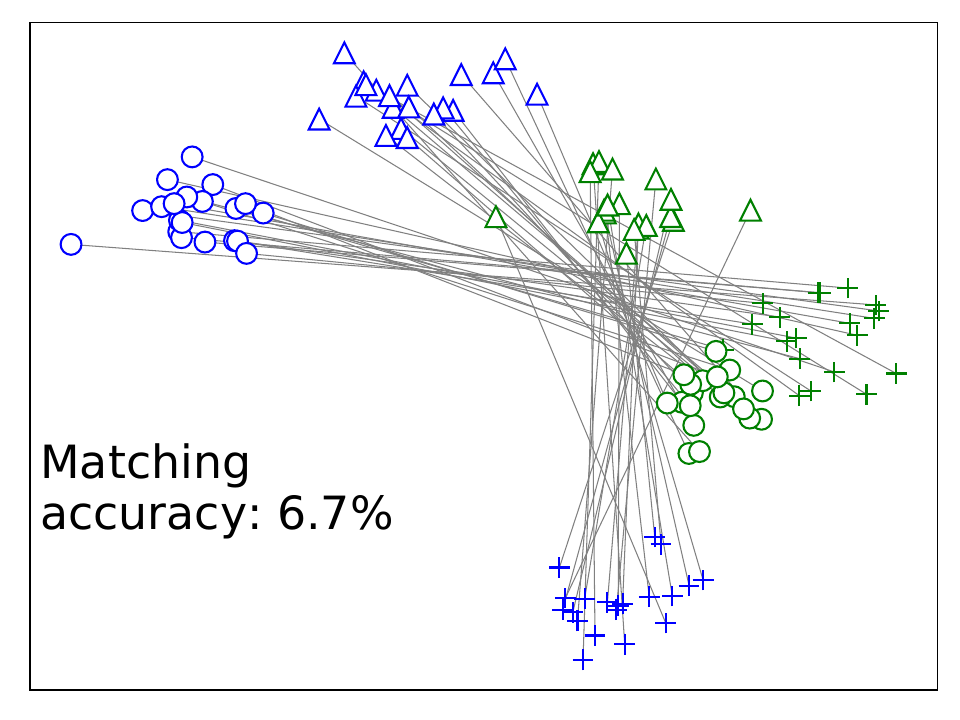} 
		\label{fig:toy_GW}}
	\subfigure[KPG-RL-GW  (w/ 2 keypoints)]{ \includegraphics[width=0.3\columnwidth]{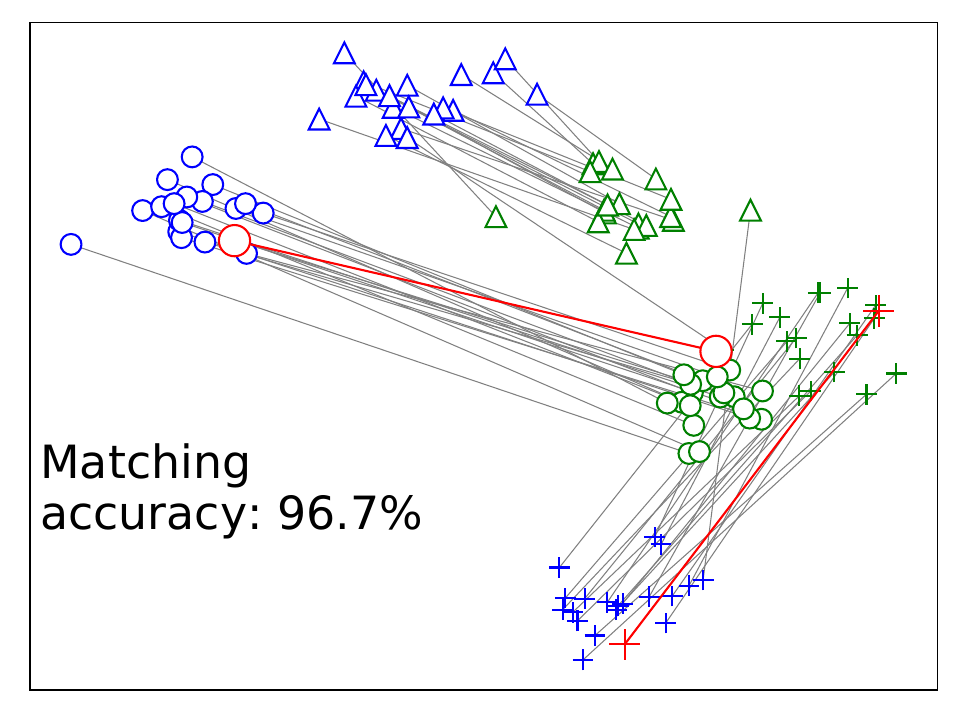} 
		\label{fig:toy_KPG_GW_21}}
	\subfigure[KPG-RL-GW  (w/ 3 keypoints)]{ \includegraphics[width=0.3\columnwidth]{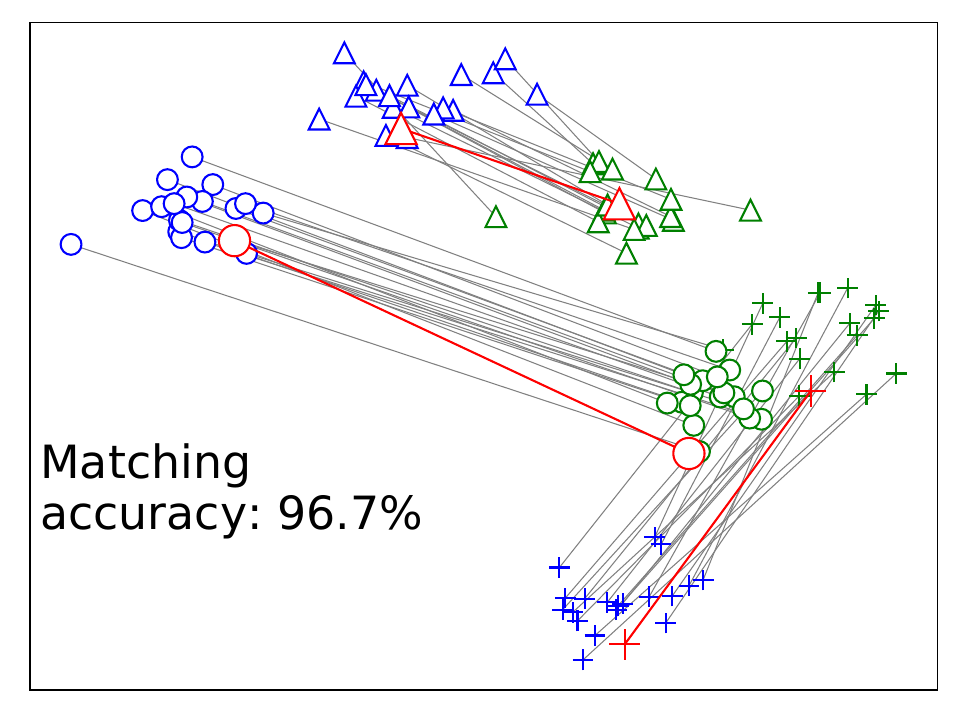} 
		\label{fig:toy_KPG_GW_3}}
	\caption{Matching produced by (a) KP, (b) KPG-RL-KP model given 2 keypoint pairs, (c) KPG-RL-KP  model given 3 keypoint pairs, (d) GW model, (e) KPG-RL-GW  model given 2 keypoint pairs, and (f) KPG-RL-GW  model given 3 keypoint pairs.}
	\label{fig:toy_experiment_main}
\end{figure}

\vspace{0.4\baselineskip}\textit{Toy experiments for evaluating KPG-RL-KP and KPG-RL-GW.} As illustrated in Fig.~\ref{fig:toy_experiment_main}, in this toy data experiment, each of the source (blue) and target (green) distributions is a Gaussian mixture composed of three distinct Gaussian components indicated by different shapes where the same shapes indicate points of the same class. In Fig.~\ref{fig:toy_experiment_main}, we have the following observations. In Fig.~\ref{fig:toy_OT}, in the KP model, the points in each component of target distribution are mismatched to points in different classes of source distributions, and only a small fraction of target points are correctly matched to source points belonging to the same class.\footnote{{In this experiment, the numbers of source and target points are equal, and the mass of all points (including keypoints) are equal}. The linear nature of KPG-RL model in Eq.~\eqref{eq:kpg} enables that for any source data point $x_i$, only one target point $y_j$ in $\bm{Y}$ takes non-zero entry in the optimal transport plan. Then $x_i$ and $y_j$ are matched and linked by the black line in Fig.~\ref{fig:toy_experiment_main}.} In Fig.~\ref{fig:toy_KPG_OT_21}, given 2 keypoint pairs (from distinct classes), the KPG-RL-KP  model apparently improves the correctness of matching.\footnote{A pair of cross-domain points being correctly matched means that they share the same class label. The matching accuracy is the ratio of correctly matched data pairs.} In Fig.~\ref{fig:toy_KPG_OT_3}, the KPG-RL-KP  model mainly matches the source points to the target points belonging to the same class, thanks to the given 3 keypoint pairs. This suggests that our keypoint-guided model helps recover the correct OT matching by leveraging a few given keypoint pairs. In Figs.~\ref{fig:toy_GW},~\ref{fig:toy_KPG_GW_21} and~\ref{fig:toy_KPG_GW_3}, the proposed KPG-RL-GW  model improves the correctness of matching of GW model by leveraging the guidance of given keypoints pairs.
% Please refer to Appendix~\ref{app:toy_partial_kpg} for the toy experiment for evaluating the partial-KPG-RL model.

\begin{figure}[t]
	\centering
	\subfigure[Partial OT]{ \includegraphics[width=0.3\columnwidth]{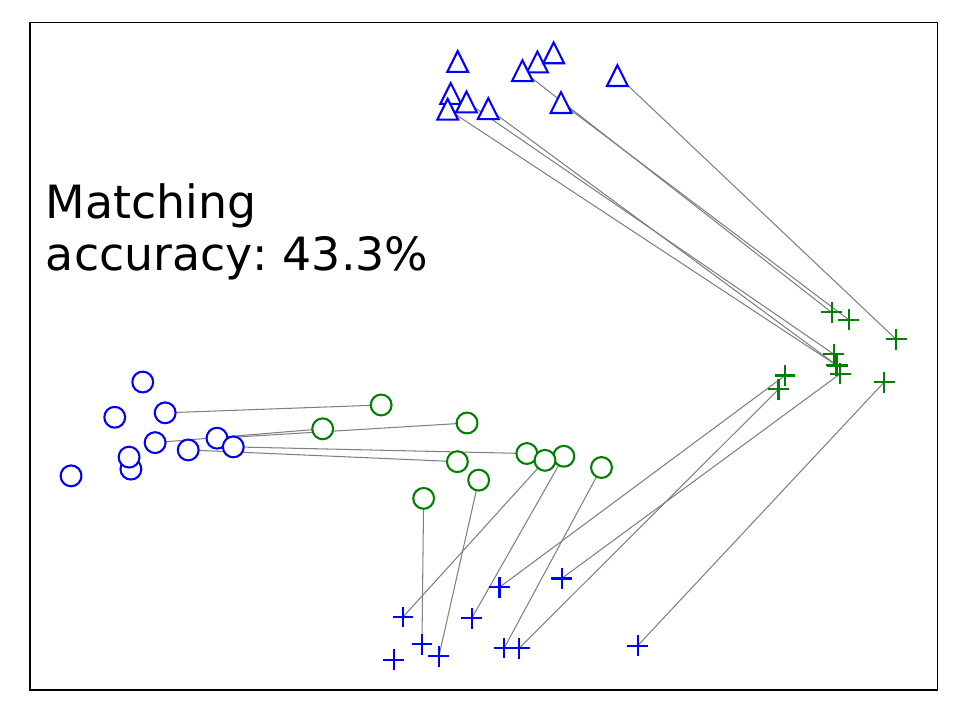} 
		\label{fig:toy_partial_OT}}
	\subfigure[Partial GW]{ \includegraphics[width=0.3\columnwidth]{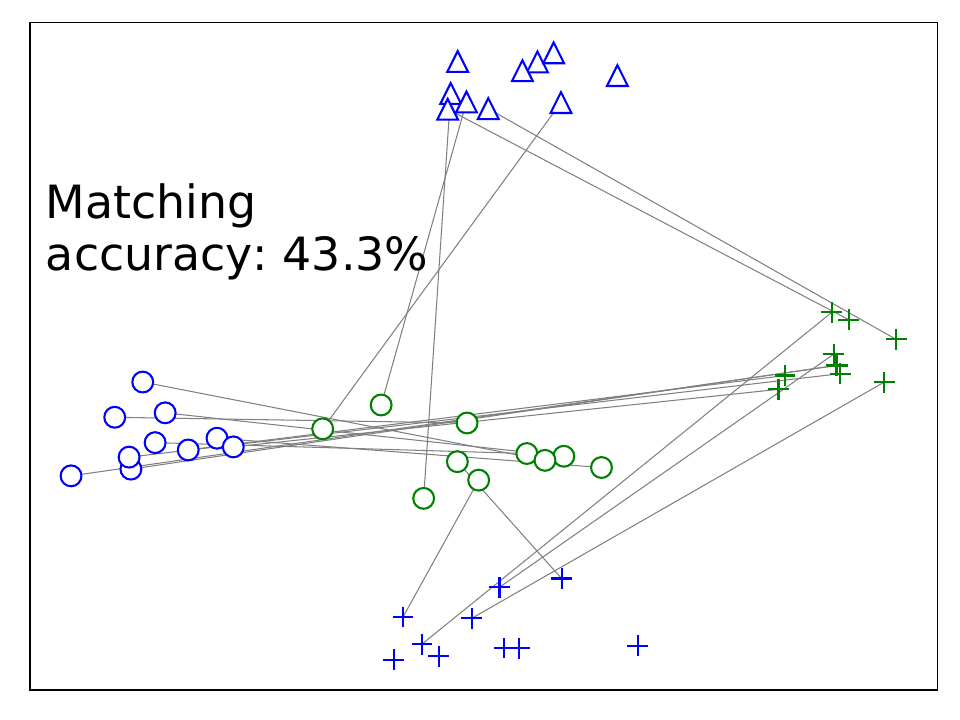} 
		\label{fig:toy_partial_GW}}
	\subfigure[Partial KPG-RL (ours)]{ \includegraphics[width=0.3\columnwidth]{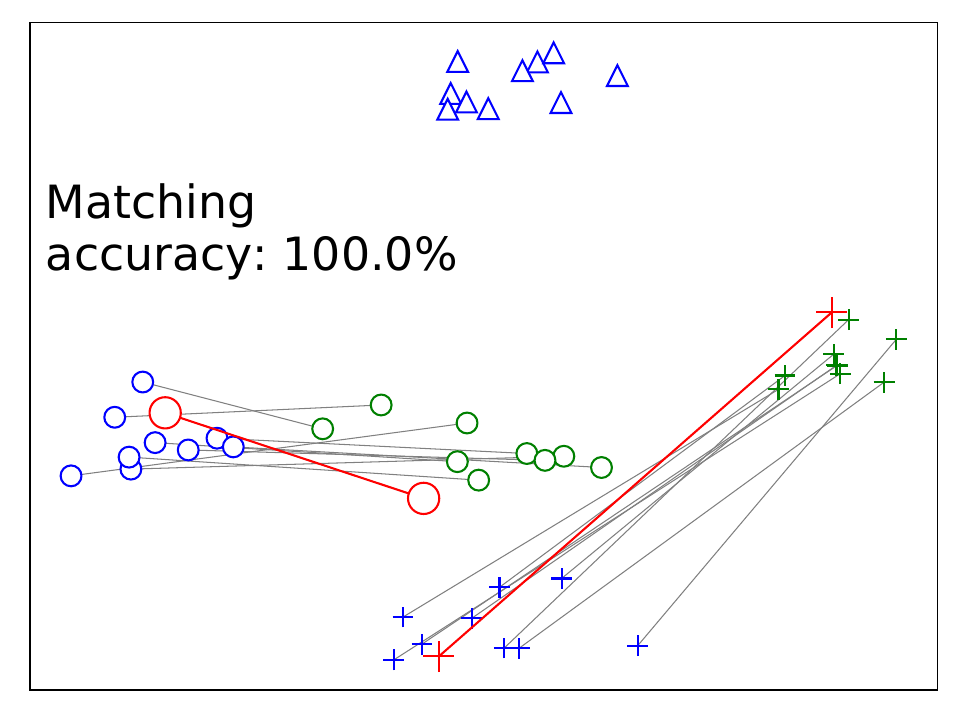} 
		\label{fig:toy_partial_kpg}}
	\caption{Matching produced by (a) partial OT model, (b) partial GW model, and (c) our proposed partial KPG-RL model.}
	\label{fig:toy_experiment}
\end{figure}

\vspace{0.4\baselineskip}\textit{Toy experiments for evaluating partial KPG-RL.}
Figure~\ref{fig:toy_experiment} illustrates the toy data experiment for evaluating the partial KPG-RL model.
In Fig.~\ref{fig:toy_experiment}, the source (blue) and target (green) distributions are Gaussian mixtures. The source (resp. target) distribution is composed of three (resp. two) distinct Gaussian components indicated by different shapes where the same shapes indicate the points of the same class. When conducting OT, the source class data represented by ``{\footnotesize{$\bigtriangleup$}}'' should not be transported. 
{The masses of all data points (including keypoints) are set to $1/m$ where $m$ is the number of source data points. Note that the total target mass is less than 1.}
In Figs.~\ref{fig:toy_experiment}(a) and~\ref{fig:toy_experiment}(b), we can observe that both the partial-OT model (defined in Eq.~\eqref{eq:pot}) and the partial-GW model~\citep{chapel2020partial} wrongly transport some source points of class ``{\footnotesize{$\bigtriangleup$}}'' to target domain and lead to lower matching accuracy. With the guidance of a few keypoints (red pairs), our proposed partial KPG-RL model does not transport the source points of class ``{\footnotesize{$\bigtriangleup$}}'' to target domain and apparently improves the matching accuracy, as in Fig.~\ref{fig:toy_experiment}(c).

\subsection{Heterogeneous Domain Adaptation}\label{sec:hda}
We discuss the experiments of heterogeneous domain adaptation (HDA) in this section, including closed-set and open-set HDA.
\subsubsection{Closed-Set Heterogeneous Domain Adaptation}
In closed-set HDA (in what follows, we refer to the closed-set HDA by HDA), we are given labeled source data $\{(x_i,t_i)\}_{i=1}^{m}$, a few labeled target data $\{y_j,\bar{t}_j\}_{j=1}^{n_l}$, and unlabeled target domain data $\{y_j\}_{j=n_l+1}^n$, where $m\gg n_l$, $n\gg n_l$, $x_i$ and $y_j$ are features, and $t_i$ and $\bar{t}_j$ are respectively the class labels of $x_i$ and $y_j$. The source and target data share the same label space (the set of classes).
The heterogeneity means that $x_i$ and $y_j$ are from different spaces/modalities.
% , \eg, $x_i$ is text and $y_j$ is image. 
The goal of HDA is to train a classification model using the given data to predict the label of unlabeled target domain data, leveraging the knowledge of labeled source domain data. 
The main challenge is that the domain gap between heterogeneous source and target distributions supported in distinct spaces hinders the direct employment of the source-trained model in the target domain. 

% Previous HDA methods~\citep{x} mainly map the source domain data to the target space and learn the classification model on the mapped data. Please refer to Appendix B2 for the review of the related HDA methods. It is important to enforce that the mapped source domain data are near the target domain data belonging to the same class. 
We tackle the problem of HDA using our proposed KPG-RL model as follows. We first transport the source domain data using our KPG-RL model to the target domain. More concretely, in KPG-RL, each labeled target domain sample and the source domain class center of the same class are taken as a matched keypoint pair, {implying the source class centers are repeated by $k$ times for $k$-shot setting. In this experiment, we resample source data such that $m=n-n_l$, and assign uniform mass $\frac{1}{m}$ to all samples, resulting in equal mass for paired keypoints.}
% We assign uniform mass ($\frac{1}{m+n_l}$) to source samples and centers. For the target domain, the labeled data (keypoints) mass is set to $\frac{1}{m+n_l}$, equal to source keypoints. The mass of unlabeled data is set to $\frac{m}{n(m+n_l)}$, ensuring equal total mass between domains.}
The KPG-RL model is then performed between empirical distributions of the source domain data along with class centers, and the target domain data (including labeled and unlabeled data). Based on the produced optimal transport plan, we transport the source domain data using the barycentric mapping~\citep{reich2013nonparametric} to the target domain. The barycentric mapping $B_{\pi}$ associated to transport plan $\pi$ is defined by $B_{\pi}(x_{i_0})=\frac{1}{\sum_{j=1}^n\pi_{i_0,j}}\sum_{j=1}^n\pi_{i_0,j}y_j.$ Note that our MBP (presented in Sect.~\ref{sec:large_scale}) learns transport map by training deep networks, which may be suitable for applications where we need to transport source samples outside the training set, \eg, I2I translation discussed in Sect.~\ref{sec:i2i_translation}. For simplicity, we directly adopt the barycentric mapping for the HDA experiments. 
Finally, we train a kernel SVM on the transported source domain data (using their class label before transport) and the labeled target domain data, which is applied to the target domain test data. In the kernel SVM, we use the Gaussian $k(x,y)=\exp(-\gamma \|x-y\|_2^2)$, where $\gamma$ is set to the reciprocal of the dimension of $x$.
$\epsilon$ is set to 0.005. 

% More details of the barycentric mapping, the kernel SVM, \textit{etc}, are given in Appendix~\ref{app:hda}. 

\begin{table}[t]
	\centering
        % \small
	\setlength{\tabcolsep}{2.0pt}
	\begin{tabular}{lccccccccc|>{\columncolor{mygray}}c}
		\toprule
		Method & A$\rightarrow$A & A$\rightarrow$D & A$\rightarrow$W  & D$\rightarrow$A & D$\rightarrow$D & D$\rightarrow$W & W$\rightarrow$A & W$\rightarrow$D & W$\rightarrow$W &  \bf Avg\\
        \midrule 
        % \multicolumn{11}{c}{DeCAF$_6\rightarrow$ResNet-50}\\
        % \midrule
        Labeled-target-only & 45.3 &69.2 &67.3 &45.3 &69.2 &67.3 &45.3 &69.2 &67.3&60.6\\
        STN&58.7&84.8&80.0&51.2&91.0&83.8&52.8&95.2&87.4&76.1\\
        SSAN&56.8&89.7&\bf87.1&54.2&82.6&\bf90.3&50.0&85.8&85.8&75.8\\
        DDACL&44.2&63.2&64.2&43.9&77.4&70.6&39.8&64.5&73.2&60.1\\
        CDSPP& 55.5 & 79.7 & 76.5 & 43.2 & 80.6 & 84.2 & 47.4 & 78.7 & 84.5 & 70.0 \\
        SGW& 49.7 & 77.7& 73.6& 49.4 & 78.4 & 73.6 & 48.7 & 80.3 & 74.5 & 67.3 \\
        \midrule
        GW& 33.6 & 41.6 & 35.5 & 39.7 & 40.0 & 31.7 & 34.8 & 34.5 & 29.0 & 35.6 \\
        \bf GW (w/ mask)& 41.3         & 71.6         & 69.7         & 41.9         & 71.2         & 69.8         & 40.3         & 71.6         & 69.7         & 60.8  \\
        HOT  & 39.0   & 44.8   & 40.0   & 31.3   & 52.6   & 44.8   & 29.7   & 60.0   & 56.5   & 44.3 \\
        \bf HOT (w/ mask) &45.2   & 60.3   & 57.4   & 48.9   & 63.5   & 59.2   & 40.3   & 67.1   & 61.4   & 55.9     \\
        TLB& 29.4   & 36.5   & 43.2   & 24.5   & 31.3   & 51.0   & 23.6   & 31.9   & 49.7   & 35.7     \\
        \bf TLB (w/ mask) &42.5   & 66.3   & 64.7   & 38.5   & 68.5   & 65.9   & 43.1   & 68.2   & 67.3   & 58.3     \\
        \midrule
        \bf KPG (w/ dist)& 55.2 & 60.7 & 71.6 & 51.3 & 71.9 & 77.1 & 48.7 & 70.0 & 77.7 & 64.9 \\
        \bf KPG-RL-GW & 58.7 & 92.9 & 84.2 & 57.4 & 95.5 & 87.1 & 55.5 & \bf95.5 &\bf 90.0 & 79.6\\
        \bf KPG-RL (LP)& 56.5 & \bf93.6 & 83.2 & \bf58.1 & 94.5 & 86.8 & 55.8 & 95.2 & 89.7 & 79.3 \\
        \bf KPG-RL (SH)& \bf 60.0 & 91.6 & 83.6 & 57.4 & \bf95.8& 87.7& \bf59.1 & 95.2 & 88.4 &\bf 79.9 \\

        \bottomrule
	\end{tabular}
    	\caption{Accuracy on Office-31 for HDA. ``A'', ``W'', and ``D'' are respectively the domains of amazon, webcam, and dslr. ``$\cdot\rightarrow*$'' denotes a heterogeneous adaptation task where $\cdot$ and $*$ are respectively source domain using DeCAF$_6$ feature and target domain using ResNet-50 feature.
	}
	\label{tab:result_office_hda}
\end{table}

{We compare our method with the following baseline methods, including 1) ``Labeled-target-only''  that trains the kernel SVM using the labeled target data; 2) the OT methods of ``GW'', ``HOT'', and ``TLB'' that transport source domain data using barycentric mapping induced by the transport plan of GW~\citep{memoli2011gromov}, Hierarchical OT~\citep{lee2019hierarchical}, and TLB~\citep{memoli2011gromov}, and then train the kernel SVM on the transported data and labeled target domain data;  3) the typical HDA methods of ``SGW''~\citep{yan2018semi} and ``STN''~\citep{yao2019heterogeneous}, and the recent HDA methods of ``SSAN''~\citep{li2020simultaneous}, ``DDACL''~\citep{yao2020discriminative} and ``CDSPP''~\citep{wang2022cross}.}
We conduct experiments on Office-31~\citep{saenko2010adapting} dataset. 
% The results on Office-31 dataset are reported in Table~\ref{tab:result_office_hda}. Due to space limit, please refer to Appendix B2 for the results on ImageCLEF-DA.
On Office-31, we use the DeCAF$_6$~\citep{donahue2014decaf} features and the features extracted by ResNet-50~\citep{He_2016_CVPR} pretrained on ImageNet~\citep{russakovsky2015imagenet} to respectively build source and target domains for constructing heterogeneous adaptation tasks. In each task, one labeled data (1-shot) for each class is given in the target domain. Note that all the methods use the same training data (including labeled source and target domain data, and unlabeled target domain data) and test data.

Table~\ref{tab:result_office_hda} reports the results on Offce-31 dataset.  In Table~\ref{tab:result_office_hda}, ``KPG-RL (SH)'' and ``KPG-RL (LP)'' denote our entropy-regularized KPG-RL model solved using Sinkhorn's algorithm and our KPG-RL model solved by linear programming, respectively. KPG-RL (SH) achieves marginally better average accuracy than KPG-RL (LP), which could be because the barycentric mapping based on the more dense transport plan optimized by Sinkhorn's algorithm is more robust to incorrect matching. We also observe that KPG-RL-GW achieves slightly lower average accuracy than KPG-RL (SH), and KPG-RL (SH) is more computationally efficient as discussed in Sect.~\ref{sec:kpg_ot}. 
``KPG (w/ dist)'' is the approach that imposes the guidance of keypoints using distance preservation in our framework, \ie, $R^s_{k,i_u}$ and $R^t_{l,j_u}$ in Eqs.~\eqref{eq:r} and \eqref{eq:r_bar} are taken as $C^s_{k,i_u}$ and $C^t_{l,j_u}$, respectively.  KPG-RL (SH) improves the accuracy of KPG (w/ dist) by 15\%, implying that our relation-preserving scheme is more effective for imposing the guidance of keypoints than the distance-preserving scheme. 
{Table~\ref{tab:result_office_hda} shows that the results of GW, HOT, and TLB are inferior to those of Labeled-target-only, indicating that GW, HOT, and TLB cause negative transfer. This is reasonable because GW, HOT, and TLB do not take advantage of the guidance of keypoints and can cause wrong matching.  We also use our mask-based constraint on the transport plan in GW, HOT, and TLB, of which the corresponding approach are denoted as GW (w/ mask), HOT (w/ mask), TLB (w/ mask). We can observe that with the mask-based constraint, the performances of GW, HOT, and TLB are improved but worse than KPG-RL (SH) and KPG-RL (LP), verifying the importance of the relation preservation scheme in KPG-RL.}
SGW~\citep{yan2018semi} aligns the class centers of labeled target domain data and transported source domain data in the GW model, and realizes positive transfer. Our proposed KPG-RL (SH) outperforms SGW by 12.2\%, confirming that our KPG-RL model preserving relation is more effective than SGW preserving class centers for HDA. Compared with the other HDA methods, KPG-RL (SH) achieves the best average accuracy. 

\begin{table}[t]
	\centering
	\setlength{\tabcolsep}{0.5pt}
	\footnotesize
	\begin{tabular}{l|ccccc>{\columncolor{mygray}}c|ccccc>{\columncolor{mygray}}c|ccccc>{\columncolor{mygray}}c}
		\toprule
		\multirow{2}*{Method} & \multicolumn{6}{c|}{1-shot}& \multicolumn{6}{c|}{2-shot}& \multicolumn{6}{c}{3-shot}\\
		~&S1&S2&S3&S4&S5&Avg&S1&S2&S3&S4&S5&Avg&S1&S2&S3&S4&S5&Avg\\
        \midrule 
        Labeled-target-only & 61.2 &58.4 &64.1 &60.6 &58.7 &60.6&77.8 &76.3 &78.3 &75.0 &74.2 &76.3 &80.7 &76.6&81.0&83.4&82.7 &80.9\\
        \bf KPG-RL (SH)  & \bf 77.6 &\bf 76.2&\bf 82.1 &\bf 82.9& \bf80.7&\bf 79.9 &\bf 86.5& \bf86.5& \bf86.8& \bf84.4 &\bf 82.6& \bf85.4 &\bf 86.5&\bf 88.5&\bf 87.0 & \bf88.3 & \bf88.2&\bf87.7\\
        % \multicolumn{11}{c}{DeCAF$_6\rightarrow$ResNet-50}\\
        % \midrule

        \bottomrule
	\end{tabular}
    \caption{Results of methods of Labeled-target-only and KPG given different shots (1, 2, and 3) of labeled target domain data (keypoints) per-class under five distinct samplings (S1,S2,S3,S4, and S5).	}
	\label{tab:result_office_hda_keypoints}
\end{table}

\begin{figure}[t]
	\centering
	\subfigure[]{ \includegraphics[width=0.4\columnwidth]{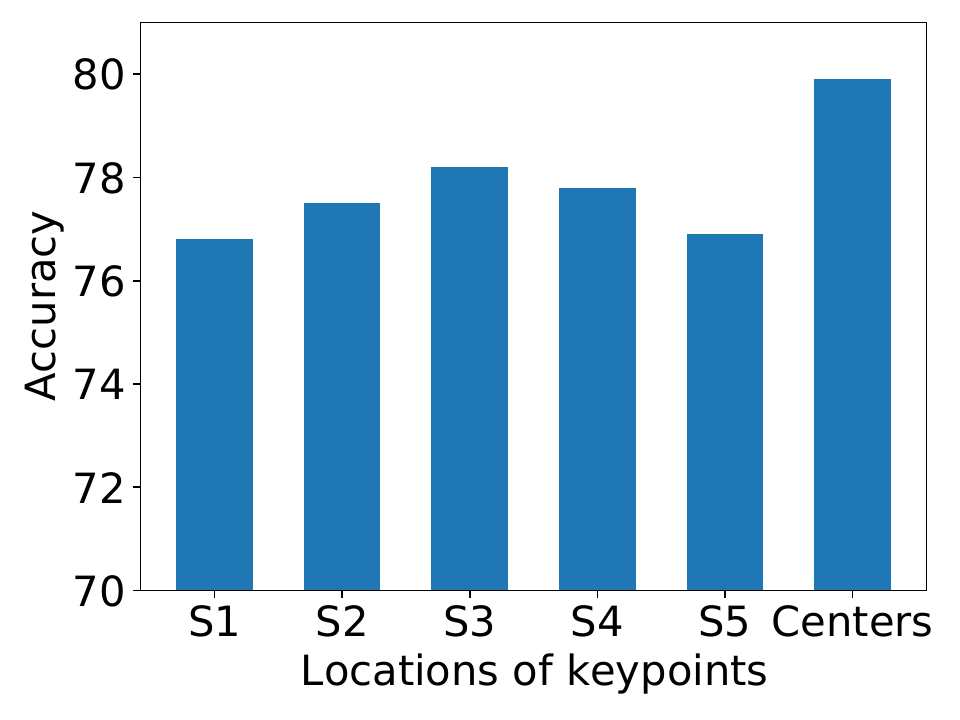} 
		\label{fig:locations_keypoint}}
	\subfigure[]{ \includegraphics[width=0.4\columnwidth]{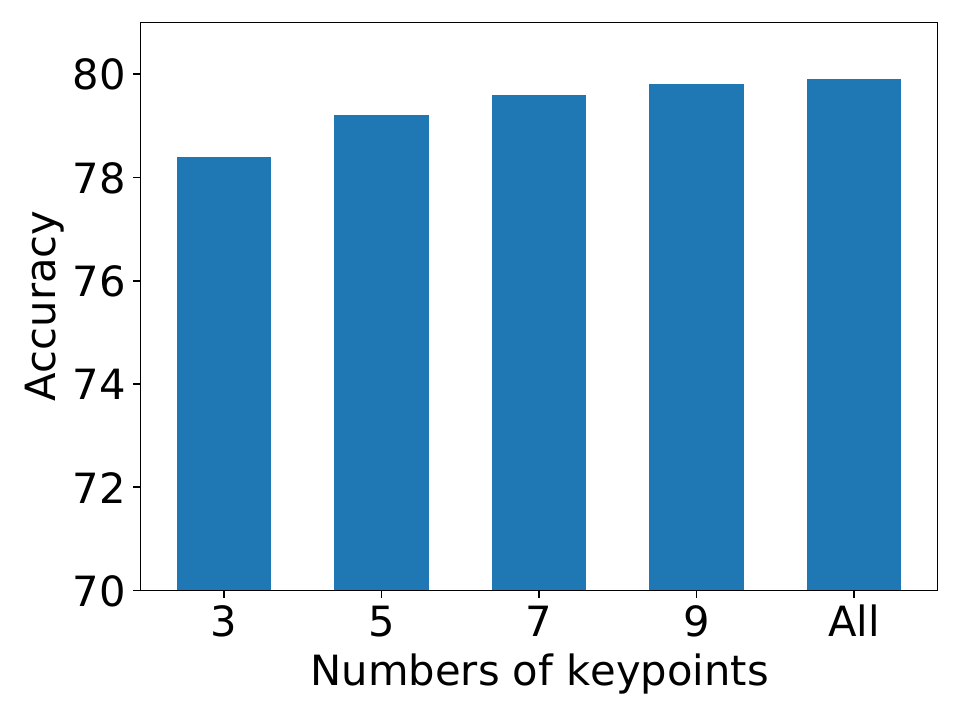} 
		\label{fig:numbers_keypoint}}
	\caption{(a) Results for different locations of source keypoints. (b)Results for different numbers of source keypoints.}
	% \label{fig:toy_experiment}
\end{figure}

\subsubsection{Ablation Studies}
We next investigate the effect of target/source keypoints, compare different choices of $d$, and study sensitivity to hyperparameters in closed-set HDA experiment, in this subsection.

\vspace{0.4\baselineskip}\textit{Effect of target keypoints.}
The keypoints are important to KPG-RL, since they are used to guide the matching. We show the results with varying keypoints by giving different numbers and samplings of labeled target domain data, in Table~\ref{tab:result_office_hda_keypoints}. It can be observed in Table~\ref{tab:result_office_hda_keypoints} that under different numbers and samplings of labeled target domain data, our approach  KPG-RL (SH) consistently outperforms the baseline Labeled-target-only, achieving positive transfer. We also find that as the number of labeled target domain data increases, the margin between the accuracy of Labeled-target-only and KPG-RL (SH) decreases. 
This indicates that given more labeled target domain data, the necessity of knowledge transferred from source domain becomes smaller. 
% Please refer to Appendix~\ref{app:ablation} for the effect of source keypoints.
% Due to space limit, we give more empirical analysis, \eg, sensitivity to hyper-parameters, in Appendix B.

\vspace{0.4\baselineskip}\textit{Effect of source keypoints.} 
In the paper, the source keypoints are taken as the source class centers. To study the sensitivity to the location of source keypoints, we randomly sample one data point from each class as a keypoint to construct the source keypoints. We run the experiments with five different samplings for constructing the source keypoints (these five runs are denoted as S1, S2, S3, S4, S5, respectively). The results are reported in Fig.~\ref{fig:locations_keypoint}.
We can see that using the class center as the keypoints achieves the best results, compared with randomly sampling one data point per class as the keypoints. This may be because the class centers are estimated using all the data of each class, and these centers can better represent each class than a randomly sampled data point of each class. 
% \begin{table}[H]
% 	\centering
% 	\caption{Results for different locations of source keypoints.}
% % 	\setlength{\tabcolsep}{3.3pt}
% 	\begin{tabular}{cccccccccc|>{\columncolor{mygray}}c}
% 		\toprule
% 		S1 &S2 &S3 &S4 &S5 &Centers   \\
%         \midrule 
%        76.8 &77.5 &78.2 &77.8 &76.9 &79.9\\
%         \bottomrule
% 	\end{tabular}
% 	\label{tab:sen_key_local}
% \end{table}
We next study the sensitivity to the number of source keypoints, of which the results are reported in Fig.~\ref{fig:numbers_keypoint}. 
We randomly sample 3/5/7/9 samples (keypoints) or use all the source samples (keypoints) for each class in the source domain to compute the source class centers, which are paired with labeled target samples for constructing the keypoint pairs. The results in Fig.~\ref{fig:numbers_keypoint} show that as the number of source keypoints increases, the accuracy gradually increases. The best result is obtained when all source samples are used to compute the class centers.
% \begin{table}[H]
% 	\centering
% 	\caption{Results for different numbers of source keypoints.}
% % 	\setlength{\tabcolsep}{3.3pt}
% 	\begin{tabular}{cccccccccc|>{\columncolor{mygray}}c}
% 		\toprule
% 		Number&	3&5	&7&9&All   \\
%         \midrule 
%        Accuracy&	78.4&79.2&79.6&79.8&79.9\\
%         \bottomrule
% 	\end{tabular}
% 	\label{tab:sen_key_num}
% \end{table}

% \vspace{0.4\baselineskip}\textit{On defining keypoints in other applications.}
% According to the results in Fig.~\ref{fig:locations_keypoint}, the class centers are better to be the keypoints than the randomly selected samples. For other practical applications, there may not be ``class labels'' available. We could first cluster the points and  then annotate the points near to the clustered centers as the keypoints.

\vspace{0.4\baselineskip}\textit{Comparison of different choices for $\bm{d}$.} Since $R_k^s$ and $R_l^t$ are in the probability simplex, it is reasonable to measure their difference by a distribution divergence/distance. The widely used distribution divergences/distances include the KL-divergence, JS-divergence, and Wasserstein distance. The KL-divergence is not symmetric, so we need to determine the order of inputs. For the Wasserstein distance, one should define the ground metric first. A possible strategy is to set the ground metric to 0 if the two keypoints are paired, otherwise 1.  Such a ground metric makes the Wasserstein distance equal to the $L_1$-distance. In this work, $d$ is taken as the JS-divergence. We compare the performance of different choices of $d$ in the experiment of HDA on Office-31, as in Table~\ref{tab:ablation_d}.
\begin{table}[t]
	\centering
	\setlength{\tabcolsep}{4.2pt}
	\begin{tabular}{lccccccccc|>{\columncolor{mygray}}c}
		\toprule
		Choices of $d$ & A$\rightarrow$A & A$\rightarrow$D & A$\rightarrow$W  & D$\rightarrow$A & D$\rightarrow$D & D$\rightarrow$W & W$\rightarrow$A & W$\rightarrow$D & W$\rightarrow$W &  \bf Avg\\
        \midrule 
         KL-ST & 59.0   & 89.7   & \bf 83.6    & 56.8   & 95.2   & \bf 89.0    & 57.7   & 93.6   & 88.1   & 79.2   \\
        KL-TS & 58.1   & 89.0   & 82.3   & 54.2   & 93.9   & 88.1   & 54.2   & 93.2   & \bf 89.4        & 78.0     \\
        $L_1$-distance & 57.4   & 85.8   & 79.0   & 58.0   & 85.8   & 82.9   & 58.4   & 92.6   & 83.6   & 75.9     \\
        $\chi^2$-distance & 52.3   & 85.8   & 81.3   & 53.2   & 91.3   & 82.3   & 52.6   & 90.3   & 82.9   & 74.7     \\
        GW    & 42.0   & 71.6   & 70.0   & 41.6   & 71.0   & 69.4   & 42.3   & 71.3   & 70.0   & 61.0    \\
        JS    & \bf 60.0    & \bf 91.6    & \bf 83.6    & \bf 57.4    & \bf 95.8    & 87.7   & \bf 59.1    & \bf 95.2    & 88.4   & \bf 79.9 \\
        \bottomrule
	\end{tabular}
    \caption{Results of different choices of $d$ in HDA experiment on Office-31. }
	\label{tab:ablation_d}
\end{table}
\begin{figure}[t]
	\centering
	\subfigure[]{ \includegraphics[width=0.31\columnwidth]{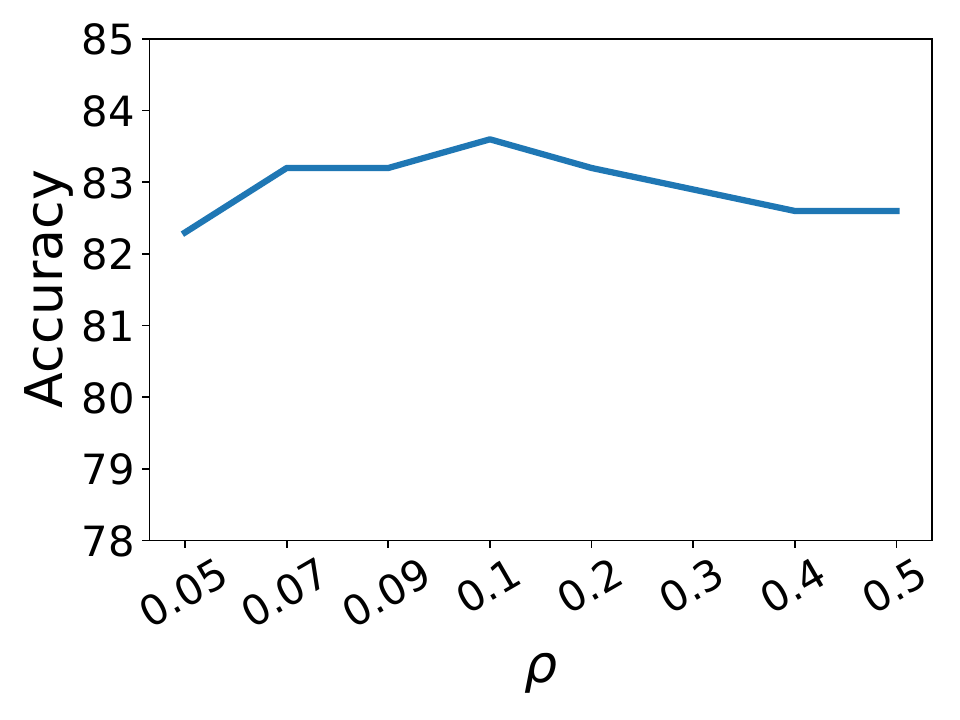} 
		\label{fig:sensitive_rho}}
	\subfigure[]{ \includegraphics[width=0.31\columnwidth]{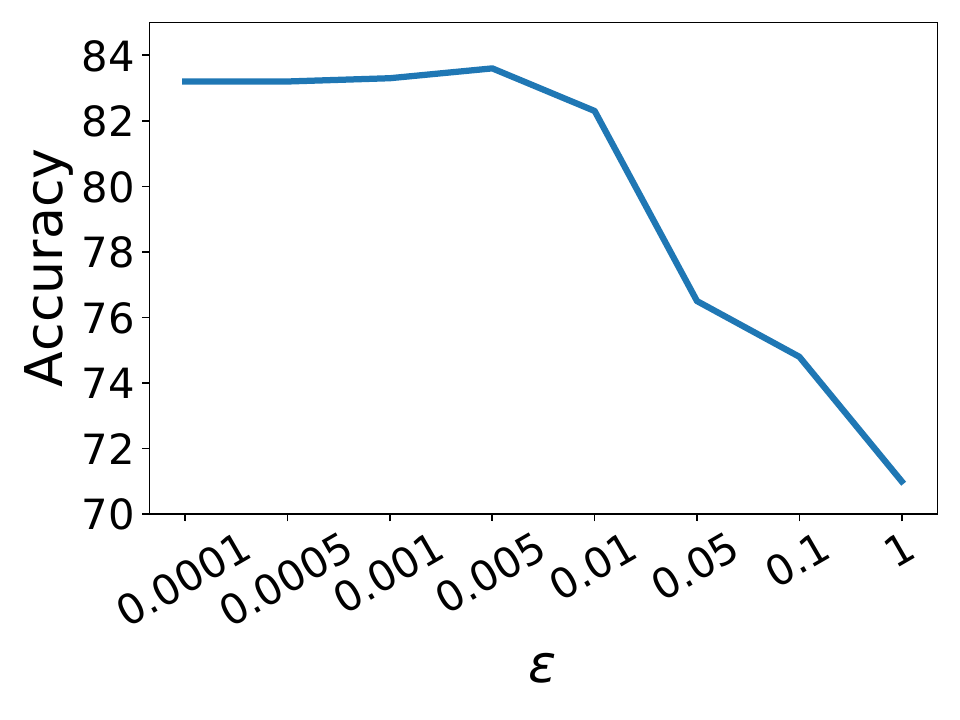} 
		\label{fig:sensitive_epsilon}}
        \subfigure[]{ \includegraphics[width=0.31\columnwidth]{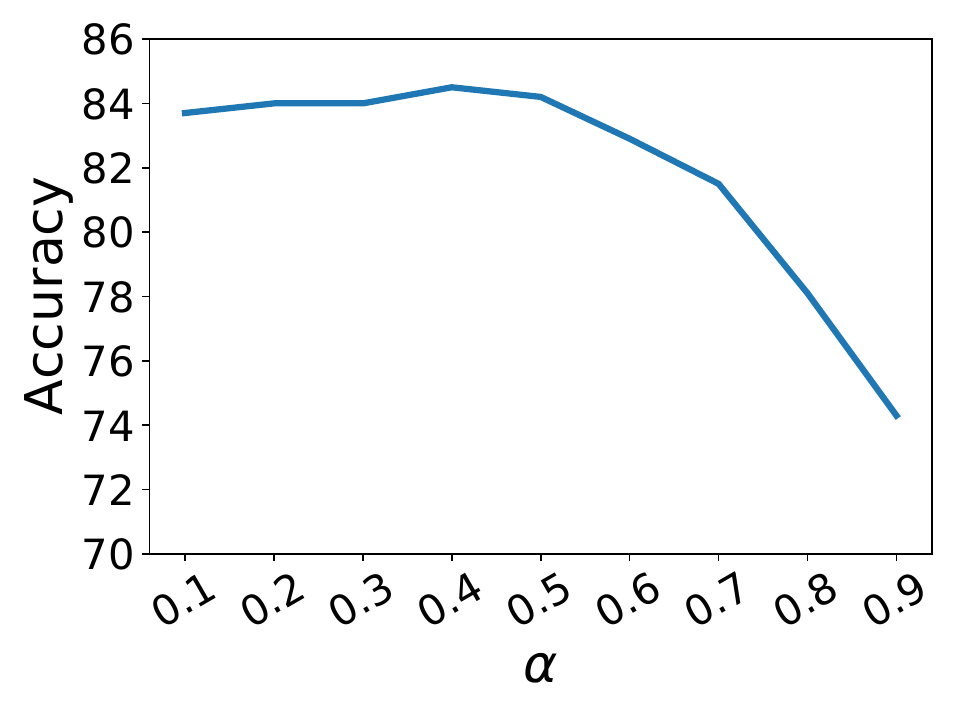} 
		\label{fig:sensitive_alpha}}
	\caption{ (a) Sensitivity of KPG-RL to hyper-parameter $\rho$.
 (b) Sensitivity of KPG-RL to hyper-parameter $\epsilon$. (c) Sensitivity of KPG-RL-GW to hyper-parameter $\alpha$. The results are for the Closed-Set HDA task A$\rightarrow$W.
	}
 \label{fig:sensitivity}
\end{figure}
In Table~\ref{tab:ablation_d}, KL-ST and KL-TS denote the KL-divergence $KL(R_k^s, R_l^t)$ and $KL(R_l^t, R_k^s)$, respectively. GW is the Gromov-Wasserstein distance between $ R_k^s$ and $R_l^t$ where the source/target cost is taken as the $\chi^2$-distance of source/target keypoints. We find that the JS-divergence achieves the best performance, compared with KL-ST, KL-TS, $L_1$-distance, $\chi^2$-distance, and Gromov-Wasserstein. 

\vspace{0.4\baselineskip}\textit{Sensitivity to hyper-parameters.}
We show the sensitivity of our method to hyperparameters $\rho$, $\epsilon$, and $\alpha$ in Fig.~\ref{fig:sensitivity}. $\epsilon$ is the coefficient of entropy regularization. $\rho$ is used to set $\tau$ and $\tau'$ for defining the relation in Eqs.~\eqref{eq:r} and~\eqref{eq:r_bar}.  $\alpha$ is in the KPG-RL-GW model. The best result for $\rho$ is obtained at 0.1. For $\epsilon$, the results are relatively stable in range [0.0001,0.005].
It can be observed that the best value of $\alpha$ is 0.4 in this task, and the results are relatively stable when $\alpha$ ranges in [0.2, 0.5]. 

\begin{table}[t]
\centering
\setlength{\tabcolsep}{8.6pt}
\begin{tabular}{lcccccccccc}
\toprule
Ratio  & 1\% & 2\% & 3\% & 5\% & 10\% & 15\% & 20\% & 30\% & 40\% & 50\% \\
\midrule
Accuracy  & 87.6 & 87.7 & 87.2 & 87.1 & 87.2 & 86.8 & 86.2 & 84.4 & 79.3 & 69.2 \\
\bottomrule
\end{tabular}
\caption{Accuracy (\%) of our method under different ratios of mismatched keypoint pairs.}
\label{tab:perturb_r}
\end{table}
{

\vspace{0.4\baselineskip}\textit{Performance under mismatched keypoint pairs.}
To assess robustness to noisy keypoint correspondences, we randomly corrupt a fraction of matched pairs by keeping the target keypoints fixed and replacing their paired source keypoints with class centers randomly sampled from an incorrect class. The results under 3-shot setting are reported in Table~\ref{tab:perturb_r}. 
We observe that KPG-RL is relatively stable under moderate mismatch (up to $20\%$, with accuracy around $87\%$), and degrades gradually at $30\%$ (84.4\%). When the mismatch becomes severe ($\ge 40\%$), the accuracy drops markedly (79.3\% at $40\%$ and 69.2\% at $50\%$), indicating that excessive cross-class mismatches can bias the keypoint guidance.

}

\subsubsection{Open-Set Heterogeneous Domain Adaptation}\label{sec:ophda}
We conduct open-set HDA experiments to evaluate our proposed partial KPG-RL model. The problem setting of open-set HDA is the same as HDA, except that the unlabeled target domain data contain samples of unknown classes absent in the categories of labeled data. The goal of open-set HDA is to correctly classify the common class samples and to detect the unknown class samples. We first consider the case that the proportion $\eta$ of unknown class samples in the unlabeled target domain data is given. We then extend the approach to the case 
 with unknown $\eta$.
% For the case that the proportion is unknown, we can first estimate this proportion and then apply our method. 
% We provide this experimental results for this case in Appendix B3.

\begin{table}[t]
	\centering
	\setlength{\tabcolsep}{2.0pt}
	\footnotesize
	\begin{tabular}{l|cc>{\columncolor{mygray}}c|cc>{\columncolor{mygray}}c|cc>{\columncolor{mygray}}c|cc>{\columncolor{mygray}}c|cc>{\columncolor{mygray}}c}
		\toprule
		 \multirow{2}*{Method} & \multicolumn{3}{c|}{A$\rightarrow$A} &\multicolumn{3}{c|}{ A$\rightarrow$D }& \multicolumn{3}{c|}{A$\rightarrow$W}  &\multicolumn{3}{c|}{ D$\rightarrow$A}& \multicolumn{3}{c}{D$\rightarrow$D}\\
% 		 \cmidrule{2-4} \cmidrule{5-7}
		  ~& OS$^*$&UNK&HOS& OS$^*$&UNK&HOS& OS$^*$&UNK&HOS& OS$^*$&UNK&HOS&OS$^*$&UNK&HOS\\
% 		 & D$\rightarrow$D & D$\rightarrow$W & W$\rightarrow$A & W$\rightarrow$D & W$\rightarrow$W &  Avg\\
        \midrule 
        Baseline &40.0&71.0&51.2& 37.3&85.3&51.9&29.1&82.7&43.0 &40.0&71.0&51.2&37.3&85.3&51.9\\
        STN &\bf48.2 &\bf80.6&\bf60.3 &\bf67.3&\bf86.6&\bf75.7 &54.6 &78.4& 64.3&46.3&76.6 &57.8 &63.6 &84.4&72.6 \\
        SSAN&25.4 &66.7 &36.8 &29.1 &68.4 &40.8 &\bf64.5 &\bf83.1 &\bf72.7 &34.6 &70.6 &46.4 &22.7&64.1&33.6 \\
        \bf Partial KPG-RL&41.8&77.1&54.2&48.2&74.9&58.6 &38.2&73.2&50.2&\bf52.7&\bf82.3&\bf64.3&\bf80.0&\bf92.6&\bf85.9\\
        \midrule
        \midrule
         \multirow{2}*{Method} & \multicolumn{3}{c|}{D$\rightarrow$W} &\multicolumn{3}{c|}{ W$\rightarrow$A }& \multicolumn{3}{c|}{W$\rightarrow$D}  &\multicolumn{3}{c|}{ W$\rightarrow$W}& \multicolumn{3}{c}{\bf Avg}\\
		  ~& OS$^*$&UNK&HOS& OS$^*$&UNK&HOS& OS$^*$&UNK&HOS& OS$^*$&UNK&HOS&OS$^*$&UNK&HOS\\
        \midrule 
        Baseline &29.1&82.7&43.0&40.0&71.0&51.2&37.3&85.3&51.9&29.1&82.7&43.0&35.5&79.7 &48.7\\
        STN &54.6 &79.2 &64.6&	49.9 &78.4 &60.4 &60.0 &82.6 &69.5 &55.5 &79.2&	65.2 &55.6 &80.7 &65.6\\
        SSAN&29.1 &66.4 &40.4 &31.8&68.4&43.3 &19.1&64.5&29.4&26.4&67.9&37.9&31.4&68.9&42.4 \\
        \bf Partial KPG-RL& \bf73.6&\bf89.2&\bf80.7&\bf52.7&\bf82.7&\bf64.4&\bf78.2&\bf90.9&\bf84.1&\bf71.8&\bf88.3&\bf79.2&\bf59.7&\bf83.5&\bf69.1\\

        \bottomrule
	\end{tabular}
    \caption{Accuracy of common class sample (OS$^*$), Accuracy of unknown class sample (UNK), and their harmonic mean (HOS) on Office-31 for open-set HDA ($\eta = 0.67$). 
% 	``A'', ``W'', and ``D'' is respectively the domains of amazon, webcam, and dslr. ``$\cdot\rightarrow*$'' denotes an adaptation task where $\cdot$/$*$ are source/target domains.
	}
	\label{tab:result_office_openset_hda_main}
\end{table}

\vspace{0.4\baselineskip}\textit{Open-set HDA with given $\bm{\eta}$.}
To apply the partial KPG-RL model defined in Eq.~\eqref{eq:partial_kpp_main} to open-set HDA, for each labeled target domain data, we take its corresponding source class center to construct a keypoint pair. 
We then resample the source domain data such that the total number of resampled source domain data and the source keypoints is $m'=(1-\eta) n$.  We define the source distribution as $\bm{p}=\frac{1-\eta}{m'}(\sum_{j=1}^{n_l}\delta_{c_j} + \sum_{j=n_l+1}^{m'}\delta_{x_j'})$, where $x_j'$ is a resampled source domain sample and $c_j$ is the source class center corresponding to the target labeled sample $y_j$. The target distribution is defined as $\bm{q} = \frac{1}{n}\sum_{j=1}^n\delta_{y_j}$. The partial KPG-RL model is conducted to transport mass from $\bm{p}$ to $\bm{q}$ with $s=1-\eta$. {Such a mass construction strategy results in equal mass across corresponding keypoints in the two domains.} After transport, the $\eta$-fraction unlabeled target data receiving smallest mass from source domain are detected as unknown class, and the rest unlabeled target data are taken as common class ones. Finally, we train the kernel SVM on the transported source domain data and labeled target domain data to classify the unlabeled target domain common class data.
% For applying partial-KPG-RL to open-set HDA, the keypoint pairs are constructed using labeled target domain data and their corresponding source class centers, same as KPG-RL for HDA in Sect.~\ref{sec:hda}.  We use the partial-KPG-RL model in Eq.~\eqref{eq:partial_kpp_main} to transport all the labeled source domain data to partially match the target domain data. The unmatched target domain data are detected as unknown class. We then train a kernel SVM on the transported source domain data and labeled target domain data to classify the common class samples detected by partial-KPG-RL. 

The natural baseline for open-set HDA is the approach that rejects the $\eta$-proportion unlabeled samples with the largest distance to labeled target domain data as unknown, and trains a kernel SVM on the labeled target domain data to classify common class data. We also evaluate the HDA methods STN~\citep{yao2019heterogeneous} and SSAN~\citep{li2020simultaneous} in the open-set HDA task. For STN~\citep{yao2019heterogeneous} and SSAN~\citep{li2020simultaneous}, we reject the $\eta$-proportion unlabeled samples with the lowest prediction confidence as unknown class. 
Table~\ref{tab:result_office_openset_hda_main} reports the results. We use the open-set evaluation metrics~\citep{bucci2020effectiveness}, including accuracy of common class sample (OS$^*$), accuracy of unknown class sample (UNK), and their harmonic mean HOS = $2\frac{\mbox{OS}^*\times\mbox{UNK}}{\mbox{OS}^*+\mbox{UNK}}$. We can observe in Table~\ref{tab:result_office_openset_hda_main} that our proposed partial KPG-RL achieves the best results in terms of average OS$^*$, UNK, and HOS, indicating that the partial KPG-RL is effective for both classifying common class data and identifying unknown class data. Compared with the Baseline,  partial KPG-RL outperforms it by 20.4\% in terms of average HOS, confirming the positive transfer achieved by our method. 
% More implementation details and the solution for Open-Set HDA with unknown $\eta$ are given in Appendix~\ref{app:openset_hda}.
% The results verify the effectiveness of partial-KPG-RL for open-set HDA.

% outperforms the Baseline by 20.4\% in terms of average HOS that balances the recognition of common and detection of unknown class data. It is observed that partial-KPG-RL achieves better average OS$^*$ and UNK than the Baseline, indicating that the partial-KPG-RL is effective for both classifying common class data and identifying unknown class data. Notably, partial-KPG-RL outperforms the Baseline in 8 among total 9 tasks in terms of HOS.
\vspace{0.2cm}
% {Due to space limit, we include the experiment of our approach for deep unsupervised domain adaptation in Appendix~\ref{app:uda}. More empirical analysis and ablation studies, \eg, sensitivity to hyper-parameters, the effect of $d$, the discussion on how to define the keypoints in more general practical applications, the time and memory cost, \textit{etc.}, are given in Appendix~\ref{app:ablation}.}

\begin{table}[t]
	\centering
	\setlength{\tabcolsep}{2.0pt}
	\footnotesize
	\begin{tabular}{l|cc>{\columncolor{mygray}}c|cc>{\columncolor{mygray}}c|cc>{\columncolor{mygray}}c|cc>{\columncolor{mygray}}c|cc>{\columncolor{mygray}}c}
		\toprule
		 \multirow{3}*{Method} & \multicolumn{3}{c|}{A$\rightarrow$A} &\multicolumn{3}{c|}{ A$\rightarrow$D }& \multicolumn{3}{c|}{A$\rightarrow$W}  &\multicolumn{3}{c|}{ D$\rightarrow$A}& \multicolumn{3}{c}{D$\rightarrow$D}\\
% 		 \cmidrule{2-4} \cmidrule{5-7}
        ~& \multicolumn{3}{c|}{($\hat{\eta}=0.57$)} &\multicolumn{3}{c|}{ ($\hat{\eta}=0.48$)}& \multicolumn{3}{c|}{ ($\hat{\eta}=0.62$)}  &\multicolumn{3}{c|}{ ($\hat{\eta}=0.57$)} & \multicolumn{3}{c}{ ($\hat{\eta}=0.48$)}\\
		  ~& OS$^*$&UNK&HOS& OS$^*$&UNK&HOS& OS$^*$&UNK&HOS& OS$^*$&UNK&HOS&OS$^*$&UNK&HOS\\
% 		 & D$\rightarrow$D & D$\rightarrow$W & W$\rightarrow$A & W$\rightarrow$D & W$\rightarrow$W &  Avg\\
        \midrule 
        Baseline &38.2&61.9&47.2&20.0&\bf69.3&31.0&28.2&\bf80.1&41.7&38.2&61.9&47.2&20.0&\bf69.3&31.0\\
        % STN~\cite{yao2019heterogeneous}&\bf48.2 &\bf80.6&\bf60.3 &\bf67.3&\bf86.6&\bf75.7 &54.6 &78.4& 64.3&46.3&76.6 &57.8 &63.6 &84.4&72.6 \\
        % SSAN~\cite{li2020simultaneous}&25.4 &66.7 &36.8 &29.1 &68.4 &40.8 &\bf64.5 &\bf83.1 &\bf72.7 &34.6 &70.6 &46.4 &22.7&64.1&33.6 \\
        \bf Partial KPG-RL&\bf49.1&\bf70.1&\bf57.8&\bf61.8&59.3&\bf60.5&\bf54.5&73.2&\bf62.5&\bf59.1&\bf73.6&\bf65.5&\bf83.6&66.7&\bf74.2\\
        \midrule
        \midrule
         \multirow{3}*{Method} & \multicolumn{3}{c|}{D$\rightarrow$W} &\multicolumn{3}{c|}{ W$\rightarrow$A }& \multicolumn{3}{c|}{W$\rightarrow$D}  &\multicolumn{3}{c|}{ W$\rightarrow$W}& \multicolumn{3}{c}{\multirow{2}*{\bf Avg}}\\
         ~& \multicolumn{3}{c|}{($\hat{\eta}=0.62$)} &\multicolumn{3}{c|}{ ($\hat{\eta}=0.57$)}& \multicolumn{3}{c|}{ ($\hat{\eta}=0.48$)}  &\multicolumn{3}{c|}{ ($\hat{\eta}=0.62$)} & \multicolumn{3}{c}{~}\\
		  ~& OS$^*$&UNK&HOS& OS$^*$&UNK&HOS& OS$^*$&UNK&HOS& OS$^*$&UNK&HOS&OS$^*$&UNK&HOS\\
        \midrule 
        Baseline &28.2&80.1&41.7&38.2&61.9&47.2&20.0&\bf69.3&31.0&28.2&80.1&41.7&28.8&70.4&40.0\\
        % STN~\cite{yao2019heterogeneous} &54.6 &79.2 &64.6&	49.9 &78.4 &60.4 &60.0 &82.6 &69.5 &55.5 &79.2&	65.2 &55.6 &80.7 &65.6\\
        % SSAN~\cite{li2020simultaneous}&29.1 &66.4 &40.4 &31.8&68.4&43.3 &19.1&64.5&29.4&26.4&67.9&37.9&31.4&68.9&42.4 \\
        \bf Partial KPG-RL&\bf78.2&\bf87.4&\bf82.6&\bf60.9&\bf74.5&\bf67.0&\bf81.8&66.7&\bf73.5&\bf78.2&\bf87.0&\bf82.4&\bf67.5&\bf73.2&\bf69.5\\

        \bottomrule
	\end{tabular}
    \caption{Results on Office-31 for open-set HDA with unknown $\eta$. $\hat{\eta}$ is the estimate of $\eta$ (the true $\eta = 0.67$). 
% 	``A'', ``W'', and ``D'' is respectively the domains of amazon, webcam, and dslr. ``$\cdot\rightarrow*$'' denotes an adaptation task where $\cdot$/$*$ are source/target domains.
	}
	\label{tab:result_office_openset_hda}
\end{table}
\begin{table}[t]
	\centering
	\begin{tabular}{ccccccccccc}
		\toprule
		${\eta}$&0.50&0.55&0.60&0.65&0.70&0.75&0.80\\
        \midrule 
        Average HOS&67.2&69.2&69.9&68.7&69.2&68.5&65.5\\
        \bottomrule
	\end{tabular}
    \caption{Average HOS of partial KPG-RL using varying magnitude of ${\eta}$ (the unknown true value of ${\eta}$ is 0.67).
	}
	\label{tab:sen_eta}
\end{table}

\vspace{0.4\baselineskip}\textit{Open-set HDA with unknown $\bm{\eta}$.}
For the more practical open-set HDA setting that $\eta$ is unknown, researchers can design methods to estimate $\eta$ and then apply our method using the estimate of $\eta$, or take $\eta$ as a hyper-parameter and design methods to tune it. We directly use the positive-unlabeled learning~\citep{bekker2020learning} method~\citep{zeiberg2020fast} to estimate the fraction of common class data among the target domain unlabeled data, by taking the labeled target data as positive samples. The results of different methods for open-set HDA using the estimate $\hat{\eta}$ of $\eta$ are given in Table~\ref{tab:result_office_openset_hda}. According to Table~\ref{tab:result_office_openset_hda}, the positive transfer is achieved by our method.  We can see that $\hat{\eta}$ in all tasks is lower than the true $\eta$, implying that fewer unknown class samples are detected.  Correspondingly, the UNK value (73.2\%) achieved by partial KPG-RL using $\hat{\eta}$ in Table~\ref{tab:result_office_openset_hda} is smaller than that (83.5\%)  using ${\eta}$ in Table~\ref{tab:result_office_openset_hda_main}. Surprisingly, the OS$^*$ value (67.5\%) of partial KPG-RL in Table~\ref{tab:result_office_openset_hda} is higher than that (59.7\%) in Table~\ref{tab:result_office_openset_hda_main}. As a balance, the HOS value (69.5\%) achieved by partial KPG-RL using $\hat{\eta}$ is similar to the HOS value (69.1\%) of partial KPG-RL using the true $\eta$.
In Table~\ref{tab:sen_eta}, we take ${\eta}$ as a hyper-parameter and show the average HOS achieved by partial KPG-RL using varying magnitudes of ${\eta}$. It is observed that the average HOS is relatively stable to ${\eta}$ in a relatively large range of $[0.55,0.75]$.

{
\subsection{Multi-Omic Single-Cell Alignment}\label{sec:single_cell}
We discuss the multi-omic single-cell alignment experiment in this section. In single-cell analysis, different technologies can interrogate different molecular aspects of the cell, resulting in multi-modal cell data, \eg, gene expression, protein synthesis, chromatin accessibility, \textit{etc}.  Aside from a few recent co-assay procedures that simultaneously isolate separate molecular material for each measurement, applying multiple assays on the same single cell is impossible. The goal of multi-omic single-cell alignment task is to align the different modalities of cells.  We consider the setting that along with a large number of unpaired cross-modal cells, a few paired cross-modal cells with 1-1 correspondence are provided. We take the paired cross-modal cells as keypoints, and then employ our keypoint-guided OT models to find the matching/correspondence of cells of different modalities, to realize the alignment further.

\vspace{0.4\baselineskip}\textit{Datasets.} Following~\citep{demetci2022scot}, we use two sets of single-cell multi-omics data to evaluate the usefulness of our methods for real data sets. Both data sets are generated by co-assays, and thus, we have known cell-level correspondence information for benchmarking. The first data set is generated using the \textit{scGEM} assay~\citep{cheow2016single}, which simultaneously profiles gene expression and DNA methylation. The data set (Sequence Read Archive accession SRP077853) is derived from human somatic cell samples undergoing conversion to induced pluripotent stem cells and shows a continuous trajectory. We preprocessed the data as described in~\citep{cheow2016single,demetci2022scot} and ended up with dimensions 177$\times$34 for the gene expression data and 177$\times$27 for the DNA methylation data. We take the gene expression as source domain, and the DNA methylation as target domain. The second data set is generated by the \textit{SNAREseq} assay~\citep{chen2019high}, which links chromatin accessibility with gene expression. The data (Gene Expression Omnibus accession GSE126074) is derived from a mixture of human cell lines: BJ, H1, K562, and GM12878, and shows distinct cell type clusters. We preprocess the data sets following~\citep{chen2019high,demetci2022scot}. The resulting data matrices for the SNARE-seq data set were of size 1047$\times$19 and 1047$\times$10 for ATAC-seq and RNA-seq, respectively. We take the ATAC-seq as source domain, and the RNA-seq as target domain.

\vspace{0.4\baselineskip}\textit{Compared methods.} We compare the following methods, including the typical multi-omic single-call alignment methods of UnionCom~\citep{cao2020unsupervised} and MMD-MA~\citep{singh2020unsupervised}, the OT-based methods of Pamona~\citep{10.1093/bioinformatics/btab594} and SCOTv2~\citep{demetci2022scot} that are based on partial and unbalanced Gromov-Wasserstein models, respectively, the recent method of scTopoGAN~\citep{10.1093/bioadv/vbad171}, and our proposed partial KPG-RL, partial KPG-RL-GW, unbalanced KPG-RL, and unbalanced KPG-RL-GW. {MMD-MA and scTopoGAN align distributions in a latent space using an MMD loss and a GAN objective, respectively. For a fair comparison, we augment them with an additional MSE loss on matched cells, adapting them to the semi-supervised setting. In contrast, incorporating matched-cell supervision into the other baseline methods is non-trivial.} 
% For unbalanced KPG-RL and unbalanced KPG-RL-GW, $\mu$ is set to 1 in experiments.

\vspace{0.4\baselineskip}\textit{Implementation details.} To match the real-world scenario that different modalities are often unbalanced, we consider two kinds of imbalance in experiments, namely ``MissType'' and ``Subsample''. For MissType, we discard one type of cells in each domain. For Subsample, we discard 50 percent of cells from both the first half of types in source domain and the last half of types in target domain. In both kinds, we select one matched cell pair for each shared type across domains as annotated data. {We use matched cells as cross-domain keypoints. Each source sample is assigned mass \(1/m\), where \(m\) is the number of source samples. For each matched pair, the target keypoint is assigned the same mass as its corresponding source keypoint, while the remaining (unpaired) target samples are assigned mass \(1/n\), where \(n\) is the number of target samples. Note that the total mass may differ between the source and target domains.} We then apply our keypoint-guided OT models to search for the transport plan. With the transport plan, we employ the embedding technique in~\citep{cao2020unsupervised} to realize the alignment of cells in an embedding space. For partial KPG-RL and partial KPG-RL-GW, $s$ is set to 0.9, same as Pamona~\citep{10.1093/bioinformatics/btab594}. 

\vspace{0.4\baselineskip}\textit{Evaluation metric.} For ideal alignment, the source data in the embedding space should be aligned with target data from the same cell type. To testify the correctness of alignment, we train a k-NN classifier on the target-aligned data (\ie, target data in embedding space after alignment) and apply it to the source-aligned data (\ie, source data in embedding space after alignment). The classification accuracy is taken as the evaluation metric.

\begin{table}[t]
	\centering
        \setlength{\tabcolsep}{2.6pt}
	\begin{tabular}{lcccccc>{\columncolor{mygray}}c}
		\toprule
		\multirow{2}*{Method} &\multirow{2}*{Type} &\multicolumn{2}{c}{scGEM}&&\multicolumn{2}{c}{SNARE}&\\
        \cmidrule{3-4}
        \cmidrule{6-7}
           ~&~&MissType&Subsample&&MissType&Subsample&Avg\\
           \midrule
           UnionCom&U&27.9&37.8&&56.7&23.1&36.4 \\
           Pamona&U&11.7&48.1&&2.8&2.4&16.3 \\
           SCOTv2&U&11.7&12.4&&31.7&12.5&17.1\\
           MMD-MA&S&31.6&25.1&&36.9&38.9&33.1\\
           scTopoGAN&S&29.1&25.2&&24.4&38.2&29.2 \\
           \midrule
           \bf Partial KPG-RL&S&58.6&47.4&&61.2&94.5&65.4 \\
           \bf Partial KPG-RL-GW&S&46.8&44.5&&58.4&94.0&60.9 \\
           \bf Unbalanced KPG-RL&S&56.7&\bf 49.6&&\bf 73.2&\bf 94.6&\bf 68.5 \\
           \bf Unbalanced KPG-RL-GW&S&\bf 61.3&46.7&&68.9&94.4&67.8\\
        \bottomrule
	\end{tabular}
    \caption{Accuracy (\%) of multi-omic single-cell alignment methods on scGEM and SNARE datasets. Methods are categorized as unsupervised (U) or semi-supervised (S), where S uses paired cross-modal cells while U does not.}

	\label{tab:results_single_cell}
\end{table}

\vspace{0.4\baselineskip}\textit{Results.} Table~\ref{tab:results_single_cell} reports the alignment accuracy of different methods. It can be observed that with the guidance of a few paired data, our proposed partial KPG-RL, partial KPG-RL-GW, unbalanced KPG-RL, and unbalanced KPG-RL-GW improve the accuracy of the baseline methods by more than 24\%. Partial KPG-RL-GW outperforms Pamona, which uses the Gromov-Wasserstein model by 44.6\%, confirming the effectiveness of keypoint guidance imposed by our approach in partial Gromov-Wasserstein model. Similarly, the performance improvement (50.1\%) achieved by unbalanced KPG-RL-GW over SCOTv2 based on unbalanced Gromov-Wasserstein verifies the usefulness of keypoints guidance realized by our approach in unbalanced Gromov-Wasserstein model. 
{Compared with the semi-supervised variants of MMD-MA and scTopoGAN, our proposed approaches achieve substantially higher performance, demonstrating the effectiveness of our keypoint-based strategy.}
We also find that the unbalanced KPG-RL/KPG-RL-GW performs slightly better than partial KPG-RL/KPG-RL-GW, respectively. 
}

\subsection{Image-to-Image Translation}\label{sec:i2i_translation}
The I2I translation experiments are for evaluating the proposed KPG-RL-MBP and KPG-RL-MSP discussed in Sect.~\ref{sec:large_scale}.
We consider the ``semi-paired'' I2I translation task that a large number of unpaired along with a few paired cross-domain images are given for training. 
% Currently, existing I2I translation are mainly supervised or unpaired. The supervised I2I translation requires a paired target image for each source image, which is expensive in annotation. 
% While in the unpaired I2I translation, no paired images are given, which may require additional knowledge to guide the meaningful translation. 
% Our considered task is the ``semi-supervised'' setting. 
We aim to leverage the paired cross-domain images to guide the desired translation.
% , that preserves the class information for the unpaired source images. 
We take the paired images as keypoints and use our proposed KPG-RL-MBP and KPG-RL-MSP to translate the source images to the target domain. To do this, we first learn the optimal transport plan based on Theorem~\ref{thm:dual}, and then learn the transport map through Eqs.~\eqref{eq:gan_based} and~\eqref{eq:msp}. In experiments, we set the difference measure $\bar{d}$ in Eq.~\eqref{eq:gan_based} as the squared $L_2$-distance in feature space. The experimental details are given in Appendix F.

\begin{figure}[t]
	\centering
        \includegraphics[width=0.8\columnwidth]{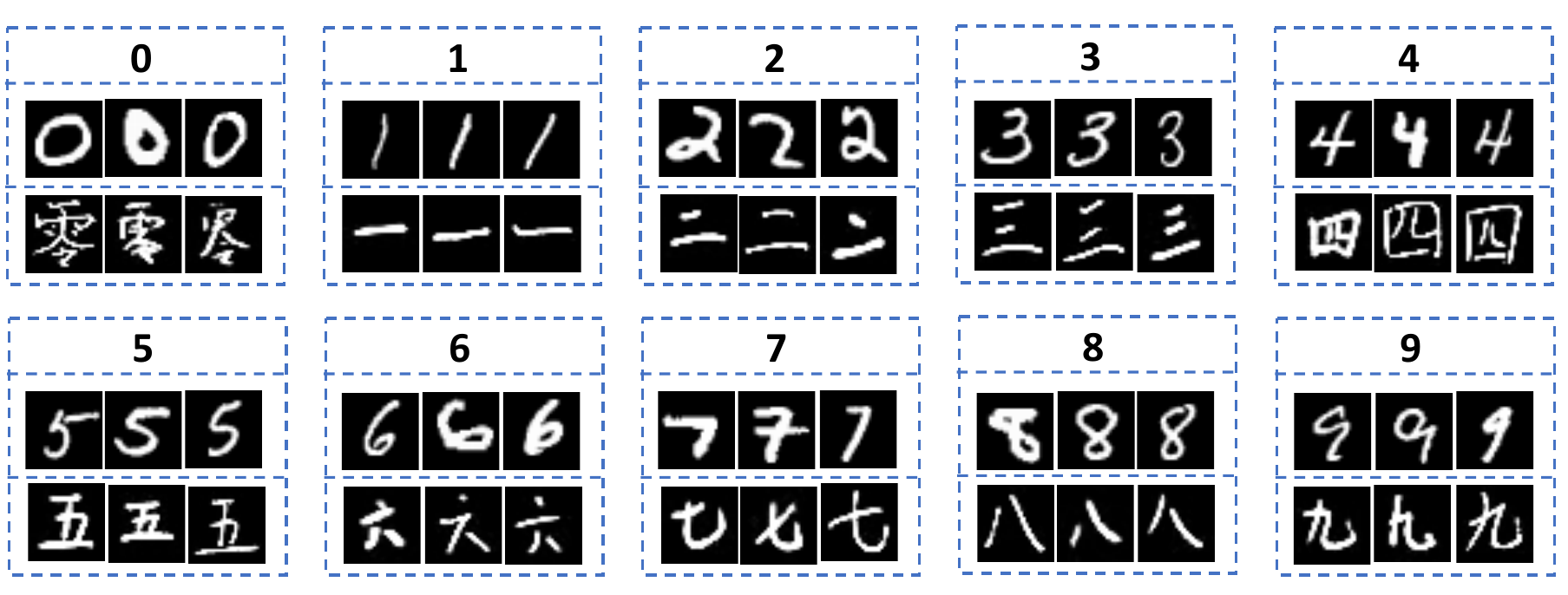}
        
	\caption{ Examples of images from MNIST and Chinese-MNIST. In each dashed box, the first line is the digit corresponding to the images on the second and last lines. The second and last lines respectively show images from MNIST and Chinese-MNIST.
	}
 \label{fig:samples_mnist_cmnist}
\end{figure}

\begin{figure}[t]
	\centering
        \subfigure[Keypoints pairs]{ \includegraphics[width=0.46\columnwidth]{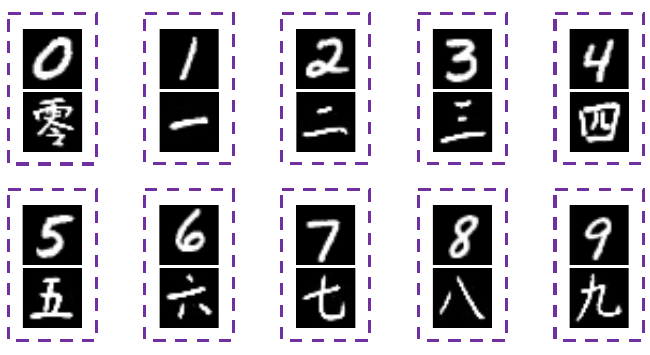} \label{fig:keypoints}}
        % \subfigure[Cycle-GAN (w/ keypoints)]{ \includegraphics[width=0.47\columnwidth]{figure//cycle-gan.pdf}\label{fig:Cycle-GAN (w/ keypoints)}}
        \subfigure[KPG-RL-BP]{ \includegraphics[width=0.47\columnwidth]{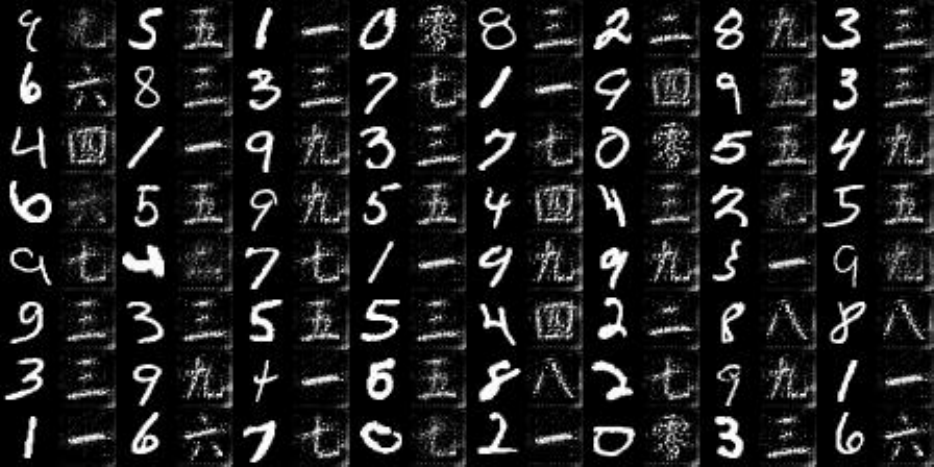}\label{fig:BP}}
        \subfigure[KPG-RL-MBP]{ \includegraphics[width=0.47\columnwidth]{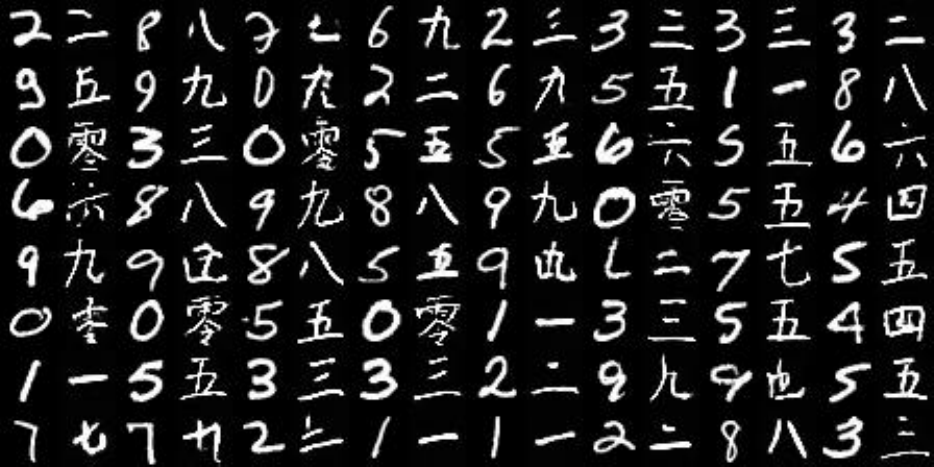}\label{fig:T_GAN}}
        \subfigure[KPG-RL-MSP]{ \includegraphics[width=0.47\columnwidth]{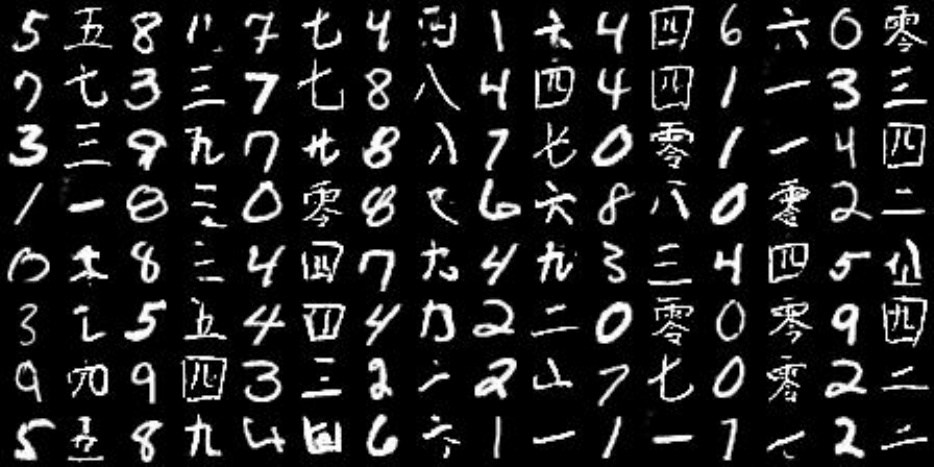}\label{fig:T_MSP}}
	\caption{ (a) Annotated keypoint pairs for digits. Each keypoint pair is in a box. (b-d) Original and transported source images by (b) KPG-RL-BP, (c) KPG-RL-MBP, and (d) KPG-RL-MSP on digits. In (b-d), the odd columns are the source images, and their right side are the corresponding transported images.
	}
 \label{fig:transported_images}
\end{figure}

\vspace{0.4\baselineskip}\textit{Datasets.} The experiments are conducted on digits and natural animal images. For \textit{Digits}, we take the MNIST~\citep{lecun1998gradient} and Chinese-MNIST\footnote{\url{https://www.kaggle.com/datasets/gpreda/chinese-mnist}}  datasets as source and target distributions, respectively. The MNIST and Chinese-MNIST contain the digits (from 0 to 9) in different modalities, respectively. 
Examples of images from MNIST and Chinese-MNIST are illustrated in Fig.~\ref{fig:samples_mnist_cmnist}. 
In experiments, we annotate 10 keypoint pairs, each corresponding to a digit, as shown in Fig.~\ref{fig:keypoints}. We expect that with the guidance of the 10 keypoint pairs, the source images can be transported to the target images that represent the same digit. For \textit{Natural Animal Images}, we take three species (cat, fox, leopard) of animals from AFHQ~\citep{choi2020stargan} as source distribution, and another three species (lion, tiger, wolf) as target distribution. We randomly choose 1000 images for each species. We resize the images to 256$\times$256.  Three keypoint pairs are given, as shown in Fig.~\ref{fig:keypoints_animal}. By the guidance of the keypoint pairs, we expect that the cat, fox, and leopard images are transported to the images of lion, tiger, and wolf, respectively. 

\vspace{0.4\baselineskip}\textit{Metrics.} We adopt two evaluation metrics, \ie, Frechet Inception Distance (FID)~\citep{heusel2017gans} and Accuracy (ACC). ``FID'' is a commonly adopted metric in deep generative models for measuring how well the transported images resemble the target images. Lower FID indicates better image quality. The metric of ``ACC'' measures how well the source images are transported to be target images of ground-truth transported classes. 
For digit images, the ground-truth transported classes are defined as the digits of original source images.
For animal images, the ground-truth transported classes for source images of cat, fox, and leopard are defined as lion, tiger, and wolf, respectively.
Specifically, we train a classifier on the target data to recognize their class labels (\ie, digits or animal species). We then predict the class labels of the transported source images using the trained classifier, and calculate the accuracy of the predictions against corresponding ground-truth transported class labels. Higher accuracy indicates that the source images are better transported to ground-truth transported classes, and thus the guidance of keypoints is better realized.

{
\vspace{0.4\baselineskip}\textit{Compared methods.} We compare the following I2I translation methods (see Table~\ref{tab:quantitative_results_digits}), including the typical GAN methods of Cycle-GAN~\citep{zhu2017unpaired}, TCR~\citep{mustafa2020transformation}\footnote{Note that TCR~\citep{mustafa2020transformation} is assumes that besides the paired images, the unpaired images are only from the source domain. Since unpaired target data are given in our setting, we additionally employ the adversarial training loss on unpaired data for TCR for a fair comparison.}, and W2GAN~\citep{korotin2021wasserstein}; the recent GAN methods of KNOT~\citep{korotin2023kernel}; the diffusion methods of EGSDE~\citep{zhao2022egsde}, DSBM~\citep{shi2023diffusion} and DDIB~\citep{su2023dual}; the flow methods of ReFlow~\citep{liu2023flow}, OT-CFM~\citep{tong2024improving}, and SF2M~\citep{pmlr-v238-tong24a}.  For the GAN methods, we additionally use a MSE loss on the paired data in training. Among these methods, W2GAN, KNOT, ReFlow, OT-CFM, and SF2M are based on OT. In particular, OT-CFM and SF2M use the mini-batch OT to guide the training of flow matching. The compared methods also include our proposed KPG-RL-BP, KPG-RL-MBP, and  KPG-RL-MSP. We additionally utilize the mini-batch version of our KPG-RL to replace the mini-batch OT in OT-CFM and SF2M to guide the training of flow matching, and the corresponding methods are denoted as KPG-RL-CFM and KPG-RL-SF2M, respectively.
}

\begin{table}[t]
	\centering
	\setlength{\tabcolsep}{9.0pt}
	\begin{tabular}{lcccccc}
		\toprule
		\multirow{2}*{Method} &\multirow{2}*{Type}& \multicolumn{2}{c}{Digits}&&\multicolumn{2}{c}{Natural animal images}\\
            \cmidrule{3-4}
            \cmidrule{6-7}
           ~&~&FID $\downarrow$& ACC (\%) $\uparrow$&&FID $\downarrow$ & ACC (\%)  $\uparrow$\\
           \midrule
           Cycle-GAN &GAN&6.99&22.72&&78.56&30.27\\
           % TCR&129.43&26.57&&342.48&33.33\\
           TCR &GAN&6.90&36.21&&74.13&32.60\\
           % *WGAN-QC&OT,GAN&- -&- -&&- -&- -\\
           W2GAN&OT, GAN&12.04&34.21&&121.86&28.40\\
           % OT-ICNN&OT,GAN &14.37&29.12&&126.43&34.67\\
           KNOT&OT, GAN&3.18&9.25&&118.26&27.33\\
           EGSDE&Diffusion&34.72&11.78&&52.11&29.33\\
           DDIB&Diffusion&9.47&9.15&&28.45&32.44\\
           DSBM&Diffusion&6.52&12.34&&24.53&27.33\\
           ReFlow&OT, Flow&32.68&11.57&&56.04&29.33\\
           OT-CFM&OT, Flow&13.27&10.09&&23.78&31.33\\
           SF2M&OT, Flow&12.83&9.47&&22.46&28.67\\
           \midrule
           \bf KPG-RL-CFM&OT, Flow& 13.86&68.62&&23.58&80.27\\
           \bf KPG-RL-SF2M&OT, Flow&12.96&65.13&&23.15&78.33\\
           \bf KPG-RL-BP&OT&98.56&74.51&&305.43&58.00\\  
           \bf KPG-RL-MBP&OT, GAN&{5.36}&\textbf{76.28}&&17.34&\textbf{87.60}\\
           \bf KPG-RL-MSP&OT, GAN&\textbf{3.05}&{72.36}&&\bf 16.62&{87.27}\\
        \bottomrule
	\end{tabular}
    \caption{FID and accuracy of the transported source images. Smaller FID indicates higher quality of transported source images. Higher accuracy implies better imposing the guidance of the keypoints in transport.
 % *WGAN-QC does not produce reasonable images in I2I translation.
	}
	\label{tab:quantitative_results_digits}
\end{table}

{
\vspace{0.4\baselineskip}\textit{Results on digits for I2I translation}
% We take the training images of MNIST~\citep{lecun1998gradient} as source distribution and the Chinese-MNIST~\footnote{\url{https://www.kaggle.com/datasets/gpreda/chinese-mnist}} as target distribution. The MNIST and Chinese-MNIST contain the numbers (from 0 to 9) in different modalities, respectively. 
% Several sampled images are illustrated in Fig.~\ref{fig:samples_mnist_cmnist}. 
% In experiments, we annotate 10 keypoint pairs, each corresponding to a number, as shown in Fig.~\ref{fig:keypoints}. We expect that with the guidance of the 10 keypoint pairs, the source images can be transported to the target ones representing the same numbers, \ie, the number is preserved after transport.
% Note that in this experiment, we learn the transport plan in feature space, in which we train auto-encoders for MNIST and Chinese-MNIST respectively, and the encoder part is used to extract features. 
In Figs.~\ref{fig:BP},~\ref{fig:T_GAN}, and~\ref{fig:T_MSP}, we visualize the transported images by KPG-RL-BP, KPG-RL-MBP, and KPG-RL-MSP. 
% We can observe that the transported source images by the Cycle-GAN (w/ keypoints) are clear, but there seem to be many source images incorrectly transported to be the target images of other digits. 
% This implies that Cycle-GAN (w/ keypoints) does not well transport the source images to ground-truth transported classes (\ie, digit of the original source images).
% This implies that the class label (\ie, digit) of the source images is not well preserved by Cycle-GAN (w/ keypoints) after transport.
Figure~\ref{fig:BP} shows that most transported source images by KPG-RL-BP belong to corresponding ground-truth transported classes, but are blurry. In Figs.~\ref{fig:T_GAN} and \ref{fig:T_MSP}, it can be observed that most transported source images in even columns by KPG-RL-MBP and KPG-RL-MSP share the same digits (class labels) as the original source images in odd columns. Meanwhile,  KPG-RL-MBP and KPG-RL-MSP produce transported images with less blur than KPG-RL-BP, which is attributed to our manifold constraints $\mathcal{L}_M$ in Eq.~\eqref{eq:gan_based} and manifold sampling~\eqref{eq:msp}, respectively. We give the quantitative results in Table~\ref{tab:quantitative_results_digits}. In the left half of Table~\ref{tab:quantitative_results_digits}, we can observe that our approaches, \ie, KPG-RL-CFM, KPG-RL-SF2M, KPG-RL-BP, KPG-RL-MBP, and  KPG-RL-MSP, outperform the compared methods by more than 25\% in terms of accuracy, demonstrating that our keypoint-guided models indeed realize the guidance of paired data to produce better translation. Meanwhile, our approaches except for KPG-RL-BP achieve comparable FID compared with the baselines. The KPG-RL-CFM and KPG-RL-SF2M respectively achieve comparable FID compared with OT-CFM and SF2M, but improve the accuracy of OT-CFM and SF2M by more than 50\%, indicating that our keypoint-guidance techniques can improve the performance of OT-CFM and SF2M.

% we can observe that our KPG-RL-MBP achieves the highest accuracy and the lowest FID on digits among the compared approaches, indicating that our KPG-RL-MBP can better transport source images to the ground-truth transported classes and produce transported images of high quality. Note that in Table~\ref{tab:quantitative_results_digits}, WGAN-QC\footnote{The original WGAN-QC maps noises to images. To apply WGAN-QC to the I2I translation task, we replace the generator and discriminator with those of our approach and take source images as the inputs of the generator. The implementation is based on the official code of WGAN-QC.} generates the same noisy output for all input source images, and we do not report its accuracy and FID.

% \begin{table}[t]
% 	\centering
% 	\caption{FID and accuracy of target classifier on the transported source images.
% 	}
% % 	\setlength{\tabcolsep}{3.3pt}
% 	\begin{tabular}{ccccc}
% 		\toprule
% 		Method&Cycle-GAN (w/ keypoints)&$T_{BP}$&$T_{GAN}$\\
%         \midrule 
%         FID $\downarrow$& 9.99 & 157.38 &\textbf{6.54}\\
%         Accuracy (\%) $\uparrow$&22.72&74.51&\textbf{76.14}\\
%         \bottomrule
% 	\end{tabular}
% 	\label{tab:quantitative_results_digits}
% \end{table}

\begin{figure}[t]
	\centering
         \includegraphics[width=1.0\columnwidth]{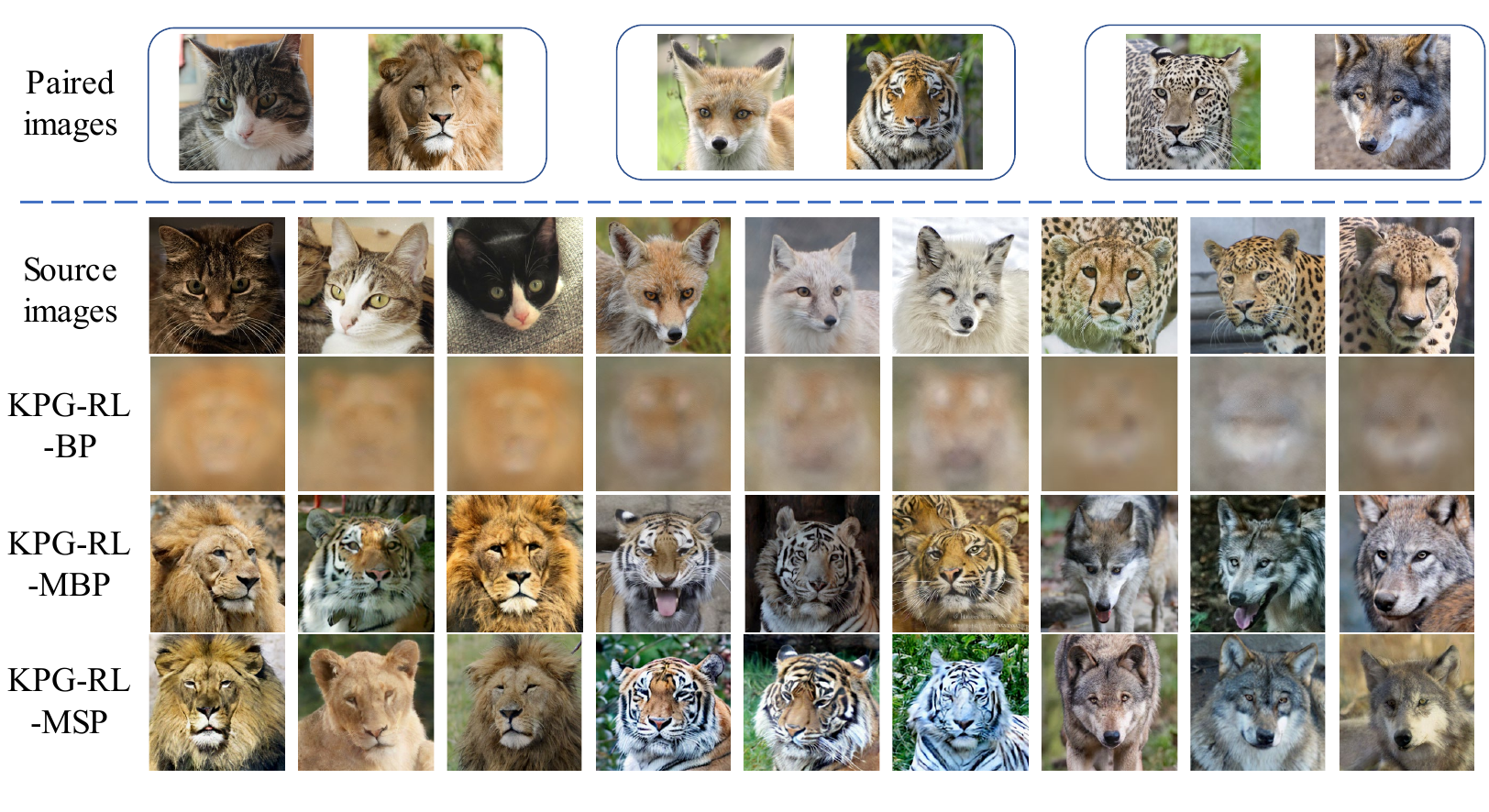}
	\caption{ Top: annotated paired images (keypoints). Bottom: transported images of corresponding source images by KPG-RL-BP, KPG-RL-MBP, and KPG-RL-MSP. By the guidance of paired keypoints, the images of cat/fox/leopard are expected to be transported to images of lion/tiger/wolf. 
	}
 \label{fig:keypoints_animal}
\end{figure}

% \begin{figure}[t]
% 	\centering
%         \subfigure[Transported source images of cat.]{
%          \includegraphics[width=1\columnwidth]{figure//transported_cat.pdf}
%          \label{fig:transported_cat}
%          }
%          \subfigure[Transported source images of fox.]{
%          \includegraphics[width=1\columnwidth]{figure//transported_fox.pdf}
%          \label{fig:transported_fox}
%          }
%          % \subfigure[]{
%          % \includegraphics[width=0.9\columnwidth]{figure//transported_lepod.pdf}}
         
% 	\caption{ Transported source images of  cat and fox by Cycle-GAN (w/ keypoints), KPG-RL-BP, and KPG-RL-MBP. By the guidance of paired keypoints (see Fig.~\ref{fig:keypoints_animal}), the images of cat/fox are expected to be transported to images of lion/tiger.
% 	}
%  \label{fig:transported_images_animal1}
% \end{figure}

% \begin{figure}[t]
% 	\centering
%         % \subfigure[]{
%         %  \includegraphics[width=0.9\columnwidth]{figure//transported_cat.pdf}}
%         %  \subfigure[]{
%         %  \includegraphics[width=0.9\columnwidth]{figure//transported_fox.pdf}}
%          % \subfigure[]{
%          \includegraphics[width=1\columnwidth]{figure//transported_lepod.pdf}
%          % }
         
% 	\caption{ Transported source images of  leopard by Cycle-GAN (w/ keypoints), KPG-RL-BP, and KPG-RL-MBP.  By the guidance of paired keypoints (see Fig.~\ref{fig:keypoints_animal}), the images of leopard are expected to be transported to images of wolf.
% 	}
%  \label{fig:transported_images_animal2}
% \end{figure}

\vspace{0.4\baselineskip}\textit{Results on natural animal images for I2I translation.}
% We conduct this experiment on AFHQ dataset~\citep{choi2020stargan}. We take three species (cat, fox, leopard) of animals from AFHQ as source, and another three species (lion, tiger, wolf) as target. We randomly choose 1000 images for each specie. To reduce the computational cost, we resize the images to 32$\times$32. Three keypoint pairs are given, as shown in Fig.~\ref{fig:keypoints_animal}. By the guidance of the keypoint pairs, we expect that the cat, fox, and leopard images are transported to the images of lion, tiger, and wolf, respectively. The experimental setup is the same as that for the digits, except for that the feature extractor is taken as the image encoder of CLIP~\citep{radford2021learning}, and the number of input/output channels of $T'$/generator is 3. The quantitative and qualitative results are given in  Table~\ref{tab:quantitative_results_digits} and Fig.~\ref{fig:transported_images_animal}, respectively. 
The results on natural animal images for I2I translation are shown in the right half of Table~\ref{tab:quantitative_results_digits} and Fig.~\ref{fig:keypoints_animal}. Table~\ref{tab:quantitative_results_digits} shows that our proposed approach KPG-RL-MBP and KPG-RL-MSP achieves a comparable FID, compared with the other methods. The accuracy achieved by our KPG-RL-MBP/KPG-RL-MSP is higher than that of the previous compared methods by more than 50\% as in Table~\ref{tab:quantitative_results_digits}. 
The results indicate that our proposed KPG-RL-MBP and KPG-RL-MSP better transport the source images to corresponding ground-truth transported classes\footnote{The ground-truth transported classes for source images of cat, fox, and leopard are lion, tiger, and wolf, respectively.}. 
% This is also verified by the qualitative results in Figs.~\ref{fig:transported_images_animal1} and~\ref{fig:transported_images_animal2}. 
We explicitly model the guidance of paired keypoints to the unpaired data in our approaches, while the other approaches do not model the relation of samples to keypoints. This may account for the higher accuracy of KPG-RL-MBP and KPG-RL-MSP than that of the previous approaches.
The KPG-RL-BP produces blurry images as in Fig.~\ref{fig:keypoints_animal}. KPG-RL-MBP and KPG-RL-MSP produce clearer images than KPG-RL-BP, as shown in Fig.~\ref{fig:keypoints_animal}, verifying the effectiveness of the manifold barycentric projection and manifold sampling. 
Note that ideally, most transported images by KPG-RL-BP though should be of the ground-truth transported classes according to Eq.~\eqref{eq:barycentric_projection}, are blurry as in Fig.~\ref{fig:keypoints_animal}. Hence the classifier may not recognize the transported images by KPG-RL-BP correctly. This leads to lower accuracy than  KPG-RL-MBP and KPG-RL-MSP in Table~\ref{tab:quantitative_results_digits}. We can also observe that KPG-RL-CFM and KPG-RL-SF2M achieve comparable FID compared with OT-CFM and SF2M, and improve their accuracy by more than 40\%. This further verifies the importance of the keypoint guidance in OT-CFM and SF2M.

\begin{table}[t]
    \centering
    \setlength{\tabcolsep}{13.0pt}
    \begin{tabular}{lc|ccccc}
        \toprule
        \multicolumn{2}{c}{Number} & 3 & 6 & 9 & 12 & Full \\
        \midrule
        \multirow{2}*{KPG-RL-MBP} & FID $\downarrow$ & 17.34&16.69&15.53&15.69&14.72\\
        \cmidrule{2-7}
        ~&ACC (\%) $\uparrow$&87.60&89.40&91.33&92.67&98.13\\
        \midrule
        \multirow{2}*{KPG-RL-MSP} & FID $\downarrow$ &16.62&16.48&16.23&16.15&15.81  \\
        \cmidrule{2-7}
        ~&ACC (\%) $\uparrow$&87.27&88.44&89.40&90.67&96.67\\
        \bottomrule
    \end{tabular}
\caption{Results of KPG-RL-MBP and KPG-RL-MSP on natural animal images with varying numbers of keypoint pairs.}
 \label{tab:effect_of_numbers}
\end{table}

\vspace{0.4\baselineskip}\textit{Effect of the number of keypoints.} We study the effect of the number of paired images in semi-paired I2I translation experiments. Table~\ref{tab:effect_of_numbers} shows that for both KPG-RL-MBP and KPG-RL-MSP, as the number of paired data increases, the accuracy increases, and the FID marginally decreases. This implies that our approach can impose the guidance of different amounts of paired data to translate source images to desired classes.
}

\section{Conclusion}\label{sec:conclusion}
This paper proposes a novel KPG-RL model that leverages keypoints to guide the correct matching (\ie, transport plan) in OT. We devise a mask-based constraint to preserve the matching of keypoints pairs, and propose to preserve the relation of each point to the keypoints to impose the guidance of these keypoints in OT. The keypoint-guided OT is built in KP/GW formulations for both balanced and partial/unbalanced transport settings. We further propose neural transport strategies of manifold barycentric projection and manifold sampling to transport source data to target domain, based on the developed duality of KPG-RL.
The effectiveness of the proposed KPG-RL model is verified in the application of HDA, multi-omic single-cell alignment, and I2I translation.
In the future, we are interested in more applications, \eg, Multimodal Large Language Models, of keypoint-guided OT.

\acks{We thank the editors and reviewers for the valuable suggestions, enabling us to improve the quality of this work. This work was supported by National Key R\&D Program 2021YFA1003000, NSFC (12501709, 12125104, 12426313, 623B2084), the Major Key Project of PCL under Grant PCL2024A06, China National Postdoctotal Program for Innovative Talents (BX20240276), China Fundamental Research Funds for the Central Universities (xzy022025047), China Postdoctoral Science Foundation (2025M773058), and Shaanxi Province ``Sanqin Bochuang'' Talent Support Program (2024SQBC021).}

%%%%%%%%%%%%%%%%%%%%%%%%%%%%%%%%%%%%%%%%%%%%%%%%%%%%%%%%%%%%

% \acks{This work was supported by ???.}


% \newpage
% \renewcommand{\thetable}{A-\arabic{table}}
% \renewcommand{\theequation}{A-\arabic{equation}}
% \renewcommand{\thetheorem}{A-\arabic{theorem}}
% \renewcommand{\theproposition}{A-\arabic{proposition}}
% \renewcommand{\thedefinition}{A-\arabic{definition}}
% \renewcommand{\thefigure}{A-\arabic{table}}
% %\renewcommand{\thesubtable}{\Roman{subtable}}
% \renewcommand{\thefigure}{A-\arabic{figure}}

% \setcounter{proposition}{0}
% \setcounter{equation}{0}
% \setcounter{figure}{0}
% \setcounter{table}{0}
% \setcounter{theorem}{0}

\appendix
\section{Proof and Generalization of Proposition~\ref{thm:thm_mask_main}}\label{sec:app_A}
\renewcommand{\theproposition}{A-\arabic{proposition}}
Proposition~\ref{thm:thm_mask_main} in the paper is for the case that  $p_i = q_j$, for all $(i,j)\in\mathcal{K}$. As stated in the paper, the mask-based modeling of the transport plan is applicable even for the case that there exist some $(i,j)\in\mathcal{K}$ such that $p_i \neq q_j$. To see this, we first mathematically give the definition of preserving the matching of keypoint pairs and then prove Proposition~\ref{thm:thm_mask}, a generalization of Proposition~\ref{thm:thm_mask_main}. 
\begin{definition}\label{def:preserve}
Given the marginal distributions $\bm{p}$ and $\bm{q}$, we say that the transport plan $\pi\in\Pi(\bm{p},\bm{q})$ preserves the matching of a keypoint pair with index $(i,j)\in\mathcal{K}$, if $\pi$ satisfies one of the following conditions:
\begin{enumerate}
    \item If $p_i = q_j$, $\pi$ satisfies that $\pi_{i,j'}=0, \forall j'\neq j; \pi_{i',j}=0, \forall i'\neq i; \pi_{i,j} = p_i = q_j.$
    \item If $p_i > q_j$, $\pi$ satisfies that $ \pi_{i',j}=0, \forall i'\neq i; \pi_{i,j} = q_j.$
    \item If $p_i < q_j$, $\pi$ satisfies that $\pi_{i,j'}=0, \forall j'\neq j;  \pi_{i,j} = p_i.$
\end{enumerate}
\end{definition}

\begin{figure}[t]
	\centering
	\includegraphics[width=0.98\columnwidth]{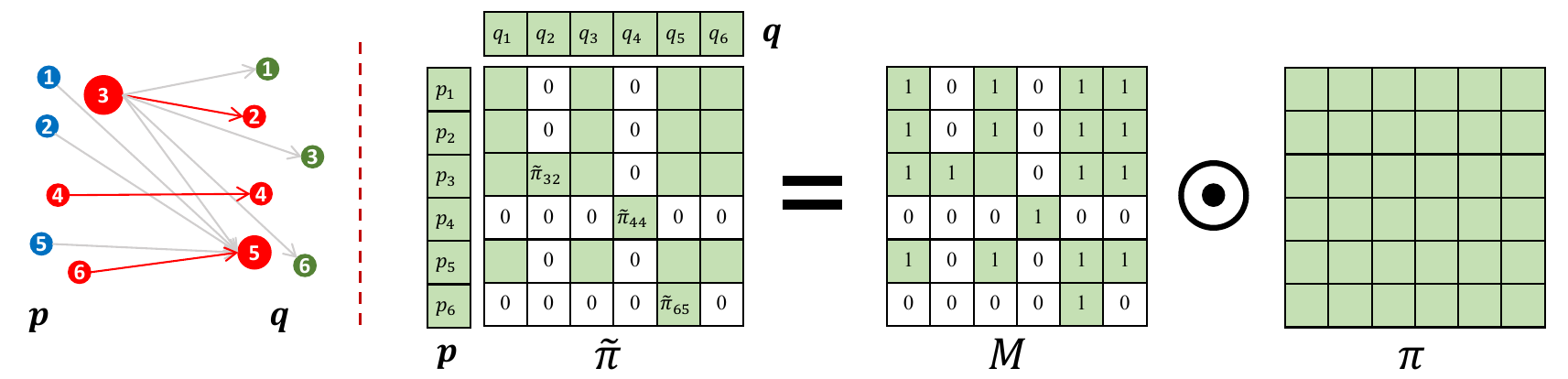} 
	\caption{Example of modeling the matching of keypoints (red) using mask, where $\mathcal{K}=\{(3,2),(4,4),(6,5)\}$ with $p_3>q_2,p_4=q_4,p_6<q_5$.}
	\label{fig:mask_modeling_app}
\end{figure}

The left part of Fig.~\ref{fig:mask_modeling_app} illustrates these conditions. Specifically, the first condition implies that if $p_i = q_j$ (\eg, $(i,j)$ is taken as (4, 4) in Fig.~\ref{fig:mask_modeling_app}), the all mass $p_i$ of $x_i$ will be transported to $y_j$ and  $y_j$ can only receive mass from $x_i$. 
The second condition implies that if $p_i > q_j$ (\eg, $(i,j)$ is taken as (3, 2) in Fig.~\ref{fig:mask_modeling_app}), $y_j$ can only receive mass from $x_i$ and consequently the partial mass $p_i-q_j$ of $x_i$ is allowed to be transported to the target points apart from $y_j$.
The third condition indicates that if $p_i < q_j$ (\eg, $(i,j)$ is taken as (6, 5) in Fig.~\ref{fig:mask_modeling_app}),
the all mass $p_i$ of $x_i$ will be transported to $y_j$ and $y_j$ is enabled to receive partial mass $q_j-p_i$ from the source points apart from $x_i$. For the convenience of description, for each pair $(i,j)\in\mathcal{K}$, we denote $j=\kappa(i)$ and $i=\kappa'(j)$.
\begin{proposition}\label{thm:thm_mask}
Suppose that the mask matrix $M$ satisfies that
\begin{equation}\label{eq:construct_m}
    M_{i,j}=\begin{cases} 1, & \mbox{ if } (i,j)\in\mathcal{K},\\
    0, & \mbox{ if } i \in \mathcal{I}, \mbox{ }p_i\leq q_{\kappa(i)}, \mbox{ and }(i,j)\notin \mathcal{K},\\
    0, & \mbox{ if } j \in \mathcal{J}, \mbox{ } p_{\kappa'(j)}\geq q_j, \mbox{ and }(i,j)\notin \mathcal{K},\\
    1, & \mbox{ if } i \in \mathcal{I}, \mbox{ }p_i> q_{\kappa(i)}, \mbox{ and }(i,j)\notin \mathcal{K},\\
    1, & \mbox{ if } j \in \mathcal{J}, \mbox{ } p_{\kappa'(j)}< q_j, \mbox{ and }(i,j)\notin \mathcal{K},\\
    1, &  \mbox{ otherwise (\ie, $i \notin \mathcal{I}, j\notin\mathcal{J} $)}.
    \end{cases}
\end{equation}
% $M_{i,j} = 1$, for $(i,j)\in\mathcal{K}$; $M_{i,j} = 0$, for $i \in \mathcal{I}$ and $(i,j)\notin \mathcal{K}$; $M_{i,j} = 0$, for $j \in \mathcal{J}$ and $(i,j)\notin \mathcal{K}$; $M_{i,j} = 1$, for $i \notin \mathcal{I}$ and $j \notin \mathcal{J}$. 
% and $p_i = q_j$, for $(i,j)\in\mathcal{K}$. 
Then, the transport plan $\Tilde{\pi}=M\odot\pi$ with $\pi\in\Pi(\bm{p},\bm{q};M)$ preserves the matching of keypoint pairs with index in $\mathcal{K}$.
\end{proposition}
According to the definition of $M$, $M_{i,j} = 1$ for the keypoint pair $(i,j)\in\mathcal{K}$, implying that $\Tilde{\pi}_{i,j}$ could take non-zero value. For $i \in \mathcal{I}$, $(i,j)\notin \mathcal{K}$ and $p_i\leq q_{\kappa(i)}$, $M_{i,j}$ is set to 0, enforcing that the $i$-th row of $\Tilde{\pi}$ are zeros except for the location $\kappa(i)$ of the target keypoint paired with $i$ (\eg, the 4-th and 6-th rows of $\Tilde{\pi}$ in Fig.~\ref{fig:mask_modeling_app}). Similarly, for $j \in \mathcal{J}$, $(i,j)\notin \mathcal{K}$, and $p_{\kappa'(j)}\leq q_{j}$, we set $M_{i,j} = 0$, enforcing that the $j$-th column of $\Tilde{\pi}$ are zeros except for the location $\kappa'(j)$ of the source keypoint paired with $j$ (\eg, the 2-th and 4-th columns of $\Tilde{\pi}$ in Fig.~\ref{fig:mask_modeling_app}). For the other points (corresponding to the last three cases in Eq.~\eqref{eq:construct_m}), we set $M_{i,j} = 1$, indicating that there is no additional constraint on $\Tilde{\pi}_{i,j}$.
If $p_i = q_j$, for all $(i,j)\in\mathcal{K}$ (\ie, $p_i = q_{\kappa(i)}, \forall i \in \mathcal{I}$), Proposition~\ref{thm:thm_mask} degenerates to Proposition~\ref{thm:thm_mask_main} in the paper.
% Since two keypoints with index pair $(i,j)\in\mathcal{K}$ are matched, the $i$-th row and $j$-th column of the transport plan $\Tilde{\pi}$ must be zeros except $\Tilde{\pi}_{i,j}$

\vspace{0.4\baselineskip}\textit{Proof of Proposition~\ref{thm:thm_mask}:}
For any $(i,j)\in\mathcal{K}$, we next prove that $\Tilde{\pi}$ preserves the matching of keypoint pair $(i,j)$.
\begin{itemize}
    \item If $p_i=q_j$, from the definition of $M$, we have $M_{i,j'} = 0$ for all $j'\neq j$ and $M_{i,j}=1$. Then, we have $\Tilde{\pi}_{i,j'} = M_{i,j'}\pi_{i,j'} = 0$ for all $j'\neq j$. Since $\sum_{j'=1}^n \Tilde{\pi}_{i,j'}  = p_i$, we have $\Tilde{\pi}_{i,j}=p_i$. Similarly, we have  $M_{i',j} = 0$ for all $i'\neq i$. Then, we have $\Tilde{\pi}_{i',j} = M_{i',j}\pi_{i',j} = 0$ for all $i'\neq i$, and $\Tilde{\pi}_{i,j}= \sum_{i'=1}^m \Tilde{\pi}_{i',j}  = q_j$.
    \item If $p_i>q_j$, from the definition of $M$, we have $M_{i',j} = 0$ for all $i'\neq i$ and $M_{i,j}=1$. Then, we have  $\Tilde{\pi}_{i',j} = M_{i',j}\pi_{i',j} = 0$ for all $i'\neq i$, and $\Tilde{\pi}_{i,j}= \sum_{i'=1}^m \Tilde{\pi}_{i',j}  = q_j$.
    \item $p_i < q_j$, from the definition of $M$, we have $M_{i,j'} = 0$ for all $j'\neq j$ and $M_{i,j}=1$. Then $\Tilde{\pi}_{i,j'} = M_{i,j'}\pi_{i,j'} = 0$ for all $j'\neq j$, and $\Tilde{\pi}_{i,j}= \sum_{j'=1}^n \Tilde{\pi}_{i,j'}  = p_i$. 
\end{itemize}
Thus, for any keypoint pair with index $(i,j)\in\mathcal{K}$, $\Tilde{\pi}$ satisfies the conditions in Definition~\ref{def:preserve}. This means that $\Tilde{\pi}$ preserves the matching of keypoint pairs with index in $\mathcal{K}$.

\hfill$\blacksquare$

\section{Algorithms}
{
\subsection{Linear Programming for Solving KPG-RL}
We cast the matrix $G$ ({resp.} $M, \pi$) as the vector $\bm{c}\text{ ({resp.} $\bm{m},\bm{x}$)}\in\mathbb{R}^{mn}$, such that the ($i+m(j-1)$)-th element of $\bm{c}$ is $G_{ij}$. By denoting 
\begin{equation}
    A = \left[
    \begin{matrix}
    \mathbbm{1}_n^{\top}\otimes I_{m}\\
    I_n\otimes \mathbbm{1}_m
    \end{matrix}\right], \quad
    \bm{h} = \left[
    \begin{matrix}
    \bm{p}\\
    \bm{q}
    \end{matrix}\right], %\mbox{ and } \tilde{\bm{c}} = \bm{c}\odot\bm{m},
\end{equation}
where $I_m$ is the identity matrix of size $n$ and $\otimes$ is the Kronecker product, 
the KPG-RL model in Eq.~\eqref{eq:kpg} in the paper reads
\begin{equation}\label{eq:lp_form}
    \begin{split}
        &\min_{\bm{x}} {\bm{c}}^{\top}\bm{x}\\
        &s.t. \hspace{0.2cm} \bm{x}\geq 0 ,\\
        & \hspace{0.6cm} A\bm{x} = \bm{h}, \\
        & \hspace{0.6cm} \mbox{diag}(\mathbbm{1}_{mn}-\bm{m})\bm x=0.
    \end{split}
\end{equation}
With the standard form of linear programming in Eq.~\eqref{eq:lp_form}, the Simplex algorithm can be directly used to solve the KPG-RL model.

\subsection{Sinkhorn's Algorithm for Solving KPG-RL}\label{app:Sinkhorn}
The entropy-regularized model for KPG-RL is 
\begin{equation}\label{original_reg}
    \begin{split}
        &\min_{\pi} \langle M\odot \pi, G \rangle_F - \epsilon E(M\odot \pi)\\
        &s.t.  \hspace{0.2cm}\pi\geq 0, (M\odot \pi)\mathbbm{1}_{n} = \bm{p},
        (M\odot \pi)^{\top}\mathbbm{1}_{m} = \bm{q},(\mathbbm{1}_{m\times n}-M)\odot\pi=0,
    \end{split}
\end{equation}
where $E(M\odot \pi)=-(\langle M\odot \pi, \log(M\odot \pi) \rangle_F - \mathbbm{1}_m^{\top}(M\odot\pi)\mathbbm{1}_n)$ is the entropy of the transport plan $M\odot \pi$.
The Lagrangian function is
\begin{equation}
\begin{split}
     L(\pi,\bm{f},\bm{g}) = &\langle M\odot \pi, G \rangle_F + \epsilon \left(\langle M\odot \pi, \log(M\odot \pi)\rangle_F - \mathbbm{1}_m^{\top}(M\odot\pi)\mathbbm{1}_n\right) \\
      &-  \langle \bm{f}, (M\odot\pi) \mathbbm{1}_n -\bm{p}\rangle_F - \langle \bm{g}, (M\odot\pi)^{\top} \mathbbm{1}_m -\bm{q}\rangle_F,
\end{split}
\end{equation}
where $\bm{f}\in \mathbb{R}^{m}$ and $\bm{g}\in \mathbb{R}^{n}$.
The first-order conditions then yield
\begin{equation}
    \frac{\partial L}{\partial \pi_{i,j}} = M_{i,j}G_{i,j} + \epsilon M_{i,j}\log(M_{i,j}\pi_{i,j}) -M_{i,j}f_i -M_{i,j}g_j = 0.
\end{equation}
If $M_{i,j}=0$, we have $\pi_{i,j}=0$ according to the constraints, and if $M_{i,j}=1$, we have 
$\pi_{i,j}=e^{f_i/\epsilon}e^{-G_{i,j}/\epsilon}e^{g_j/\epsilon}$. Therefore, we can unify the expression as $\pi_{ij}=M_{ij}e^{f_i/\epsilon}e^{-C_{ij}/\epsilon}e^{g_j/\epsilon}$. The matrix form is $\pi=\mbox{diag}(\bm{u})K\mbox{diag}(\bm{v})$ where $\bm{u}=e^{\bm{f}/\epsilon}, K = M\odot e^{-G/\epsilon}, \mbox{ and } \bm{v}=e^{\bm{g}/\epsilon}$. The constraints are 
\begin{equation}
    \mbox{diag}(\bm{u})K\mbox{diag}(\bm{v})\mathbbm{1}_{n} = \bm{p}, \hspace{0.4cm} (\mbox{diag}(\bm{u})K\mbox{diag}(\bm{v}))^{\top}\mathbbm{1}_{m} = \bm{q}.
\end{equation}
Since the entries of $K$ are non-negative, the Sinkhorn's algorithm can be applied~\citep{sinkhorn1967concerning}. The iteration formulas are 
% Let $\tilde{K}=M\odot K$, then the Sinkhorn iteration can be applied as 
\begin{equation}\label{eq:sinkhorn}
    \bm{u}^{(l+1)} = \frac{\bm{p}}{K\bm{v}^l}, \hspace{0.2cm}\bm{v}^{(l+1)} = \frac{\bm{q}}{K^{\top}\bm{u}^{(l+1)}}.
\end{equation}
The division operator used above is entry-wise.
}

\vspace{0.4\baselineskip}\textit{Log-domain Sinkhorn iteration.} For KP, the Sinkhorn iteration in the log-domain is more stable~\cite{schmitz2018wasserstein}. We next deduce the log-domain Sinkhorn iteration for our KPG-RL. In the log-domain, the left equation in Eq.~\eqref{eq:sinkhorn} is
\begin{equation}\label{log}
    \begin{split}
        & \frac{1}{\epsilon}f_i^{(l+1)} = \log(p_i) - \log\left(\sum_{j=1}^n M_{i,j}e^{\frac{-G_{i,j} +g_j^{(l)}}{\epsilon}}\right).\\
        % & g^{l+1}/\epsilon = \log(q) - (\log(M) - C/\epsilon +f^{l}/\epsilon).
    \end{split}
\end{equation}
Let 
\begin{equation}
    H_G(\bm{f},\bm{g})=\frac{1}{\epsilon}(-G+\bm{f}\mathbbm{1}^{\top}_n+\bm{g}\mathbbm{1}_m^{\top}),
\end{equation} then 
\begin{equation}\label{log1}
    \begin{split}
       f_i^{(l+1)} = &  \epsilon\log(p_i) - \epsilon\log\left(\sum_{j=1}^n M_{i,j}e^{H_G(\bm{f}^{(l)},\bm{g}^{(l)})_{i,j}}e^{-f^{(l)}_i/\epsilon}\right)\\
       = &  \epsilon\log(p_i) - \epsilon\log\left(e^{-f^{(l)}_i/\epsilon}\sum_{j=1}^n M_{i,j}e^{H_G(\bm{f}^{(l)},\bm{g}^{(l)})_{i,j}}\right)\\
       = &  \epsilon\log(p_i) - \epsilon\log\left(\sum_{j=1}^n M_{i,j}e^{H_G(\bm{f}^{(l)},\bm{g}^{(l)})_{i,j}} \right)+f^{(l)}_i\\
       = &  \epsilon\log(p_i) - \epsilon\log\left(\sum_{j=1}^ne^{\log(M_{i,j})H_G(\bm{f}^{(l)},\bm{g}^{(l)})_{i,j}} \right)+f^{(l)}_i.
        % & g^{l+1}/\epsilon = \log(q) - (\log(M) - C/\epsilon +f^{l}/\epsilon).
    \end{split}
\end{equation}
 If $M_{i,j}=0$, $\log(M_{i,j})H_G(\bm{f}^{(l)},\bm{g}^{(l)})_{i,j}=-\infty$. We define $\bar{H}(\bm{f},\bm{g})$ as 
\begin{equation}
    \bar{H}_G(\bm{f},\bm{g})_{i,j} = \begin{cases}
    H_G(\bm{f},\bm{g})_{i,j} & \mbox{ if } M_{i,j}=1,\\
    -\infty & \mbox{ if } M_{i,j}=0,
    \end{cases}
\end{equation}
and define the log-sum-exp function $\mbox{LogSumExp}:\mathbb{R}^{m\times n}\rightarrow \mathbb{R}^{m}$ as
\begin{equation}
    \mbox{LogSumExp}(A) = \left(\log(\sum_je^{A_{1, j}}),\log(\sum_je^{A_{2, j}}),\cdots, \log(\sum_je^{A_{m, j}})\right)^{\top}.
\end{equation}
Then, the matrix form of Eq.~\eqref{log1} becomes
\begin{equation}\label{eq:log_f}
    \bm{f}^{(l+1)} = \epsilon\log(\bm{p}) -\epsilon \mbox{LogSumExp}\left(\bar{H}_G(\bm{f}^{(l)},\bm{g}^{(l)})\right)+\bm{f}^{(l)}.
\end{equation}
Similarly, the corresponding iteration formula in the log-domain of the right equation in \eqref{eq:sinkhorn} is
\begin{equation}\label{eq:log_g}
    \bm{g}^{(l+1)} = \epsilon\log(\bm{q}) -\epsilon \mbox{LogSumExp} \left(\bar{H}_G(\bm{f}^{(l+1)},\bm{g}^{(l)})^{\top}\right)+\bm{g}^{(l)}.
\end{equation}
Equations~\eqref{eq:log_f} and~\eqref{eq:log_g} form the formulas of the Sinkhorn iteration in the log-domain.
{
\subsection{Convergence of Sinkhorn Iteration}
We note that $\Pi(\bm{p},\bm{q};M)$ is not empty for masks in both Proposition~\ref{thm:thm_mask_main} and Proposition~\ref{thm:thm_mask}. Thus, Proposition~\ref{thm:convergence_sinkhorn} shows the convergence of the Sinkhorn iteration for masks defined in Proposition~\ref{thm:thm_mask_main} and Proposition~\ref{thm:thm_mask}. 
We next prove Proposition~\ref{thm:convergence_sinkhorn} and empirically validate this convergence through experiments.

\vspace{0.4\baselineskip}\textit{Proof of Proposition~\ref{thm:convergence_sinkhorn}.} 
By assumption, \(\Pi(\bm p,\bm q;M)\neq\emptyset\). Since \(\Pi(\bm p,\bm q;M)\) is a convex polytope and the objective is strictly convex on the admissible entries, the entropy-regularized KPG-RL problem admits a unique global minimizer.

We next show that the Sinkhorn iteration is an alternating KL/Bregman projection scheme.
Define
\begin{equation*}
\begin{split}
      \mathcal R=&
\{\pi\in\mathbb R_+^{m\times n}:\pi\bm 1_n=\bm p,\ (\bm 1-M)\odot\pi=0\},\\
\mathcal C=&
\{\pi\in\mathbb R_+^{m\times n}:\pi^\top\bm 1_m=\bm q,\ (\bm 1-M)\odot\pi=0\},  
\end{split}
\end{equation*}
so that \(\Pi(\bm p,\bm q;M)=\mathcal R\cap\mathcal C\).
Starting from \(\pi^{(0)}=K\), alternating KL projections onto \(\mathcal R\) and \(\mathcal C\) generate
$$
\pi^{(l+1/2)} = 
P_{\mathcal R}^{\mathrm{KL}}(\pi^{(l)})
=
\arg\min_{\pi\in\mathcal R}\mathrm{KL}(\pi\mid \pi^{(l)}) = \operatorname{diag}\!\left(\frac{\bm p}{\pi^{(l)}\bm 1_n}\right)\pi^{(l)},
$$
and $$
\pi^{(l+1)}
=
P_{\mathcal C}^{\mathrm{KL}}(\pi^{(l+1/2)})
=
\arg\min_{\pi\in\mathcal C}\mathrm{KL}(\pi\mid \pi^{(l+1/2)}) =
\pi^{(l+1/2)}
\operatorname{diag}\!\left(\frac{\bm q}{(\pi^{(l+1/2)})^\top\bm 1_m}\right),
$$
where the division is element-wise. Since every iterate has the form
$$
\pi^{(l)}=\operatorname{diag}(\bm u^{(l)})K\operatorname{diag}(\bm v^{(l)}),
$$
the above updates reduce to
$$
\bm u^{(l+1)}=\frac{\bm p}{K\bm v^{(l)}},
\qquad
\bm v^{(l+1)}=\frac{\bm q}{K^\top \bm u^{(l+1)}},
$$
which are exactly the Sinkhorn iterations in Eq.~\eqref{eq:sinkhorn_main}.

By the convergence theorem for cyclic Bregman projections onto closed convex sets with nonempty intersection, the transport plans \(\pi^{(l)}\) converge to the KL projection of \(K\) onto \(\mathcal R\cap\mathcal C\), namely
$$
\pi^\star
=
\arg\min_{\pi\in\Pi(\bm p,\bm q;M)}
\mathrm{KL}(\pi\mid K).
$$
This completes the proof.

\hfill$\blacksquare$

\begin{table}[t]
\centering
\setlength{\tabcolsep}{6pt}
\renewcommand{\arraystretch}{1.15}
\begin{tabular}{l|cccccc} % 建议去掉竖线，首列通常左对齐(l)或居中(c)
\toprule
& \multicolumn{2}{c}{$\texttt{Thresh}=0.1$}
& \multicolumn{2}{c}{$\texttt{Thresh}=0.01$}
& \multicolumn{2}{c}{$\texttt{Thresh}=0.001$} \\
\cmidrule(lr){2-3} \cmidrule(lr){4-5} \cmidrule(lr){6-7} % (lr)让线条左右收缩，视觉分隔不同组
& Arange & Rand & Arange & Rand & Arange & Rand \\
\midrule
$U=3$  & $2.00\pm0.00$ & $2.00\pm0.00$ & $3.22\pm0.42$ & $3.23\pm0.42$ & $4.57\pm0.61$ & $4.53\pm0.63$ \\
$U=5$  & $2.02\pm0.14$ & $2.00\pm0.00$ & $3.26\pm0.44$ & $3.24\pm0.43$ & $4.61\pm0.58$ & $4.73\pm0.58$ \\
$U=10$ & $2.01\pm0.10$ & $2.00\pm0.00$ & $3.18\pm0.39$ & $3.20\pm0.40$ & $4.70\pm0.59$ & $4.76\pm0.61$ \\
\bottomrule
\end{tabular}
\caption{Number of iterations of the Sinkhorn's algorithm (mean$\pm$std over 100 runs) for the mask defined in Proposition~\ref{thm:thm_mask_main}. ``Thresh'' is the stopping tolerance for Sinkhorn (based on $\|\bm{u}^{(l+1)}-\bm{u}^{(l)}\|_2$). ``Arange'' places the randomly selected keypoint indices contiguously at the front, while ``Rand'' keeps the keypoint indices in random positions.}
\label{tab:kpg_sinkhorn_convergence}
\end{table}

\begin{figure}
    \centering
    \includegraphics[width=1\linewidth]{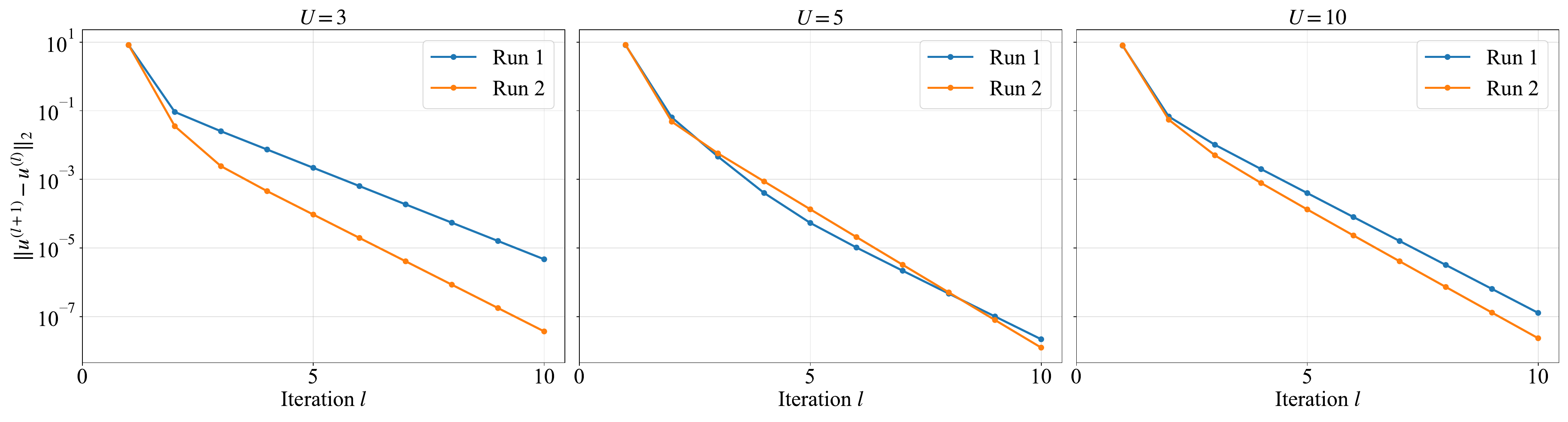}
    \caption{Example curves of $\|\bm{u}^{(l+1)}-\bm{u}^{(l)}\|_2$ in two random runs under each number of keypoint pairs, for the mask defined in Proposition~\ref{thm:thm_mask_main}.}
    \label{fig:curves}
\end{figure}

\vspace{0.4\baselineskip}\textit{Empirical convergence verification.} To empirically validate the convergence of Sinkhorn’s algorithm, we repeated the following procedure 100 times under different settings. In each run, we (i) sampled 100 points from a two-dimensional Gaussian distribution for both the source and target domains, (ii) randomly selected \(U\) cross-domain pairs as keypoints, and (iii) applied our keypoint-guided OT method (KPG-RL) using Sinkhorn’s algorithm with \(\epsilon=0.1\). Table~\ref{tab:kpg_sinkhorn_convergence} reports the iteration counts under different stopping tolerances, for varying numbers of keypoints and two kinds of keypoint index orderings. The results show that Sinkhorn reaches the prescribed tolerance in a finite number of iterations, and that the convergence behavior is insensitive to the number and ordering of keypoints. Figure~\ref{fig:curves} presents representative error curves, demonstrating a monotonic decrease as iterations proceed.

We further examine the convergence of the Sinkhorn iteration under the general mask in Proposition~\ref{thm:thm_mask}. We use the unequal-mass setting illustrated in Fig.~\ref{fig:mask_modeling_app}, where the source and target distributions are supported on six points, respectively, and the keypoint pairs are $\mathcal K=\{(3,2),(4,4),(6,5)\}$.
The prescribed marginals are set as
$\bm{p}=(0.16,0.14,0.22,0.18,0.15,0.15) \mbox{ and } q=(0.15,0.12,0.15,0.18,0.18,0.22),$
so that \(p_3>q_2\), \(p_4=q_4\), and \(p_6<q_5\). The corresponding mask is generated according to Proposition~\ref{thm:thm_mask}. 
For each trial, the source and target locations are independently sampled from a two-dimensional standard Gaussian distribution, and the cost matrix is given by the squared Euclidean distance. We run the log-domain Sinkhorn iteration with \(\epsilon=0.1\), and report the same diagnostics as in Table~\ref{tab:converge_general} and Fig.~\ref{fig:curves_general}, including the number of iterations required to reach prescribed tolerances for \(\|\bm{u}^{(l+1)}-\bm{u}^{(l)}\|_2\) and the final marginal violation. The experiment is repeated 100 times. Table~\ref{tab:converge_general} and Fig.~\ref{fig:curves_general} indicate the convergence of Sinkhorn iteration.

\begin{table}[t]
\centering
\begin{tabular}{lccc}
\toprule
\texttt{Thresh} & $0.1$ & $0.01$ & $0.001$ \\
\midrule
 $U=3$ & $2.22\pm0.44$ & $8.17\pm3.24$ & $18.25\pm5.16$ \\
\bottomrule
\end{tabular}
\caption{Number of iterations of the Sinkhorn's algorithm (mean$\pm$std over 100 runs) for the general mask defined in Proposition~\ref{thm:thm_mask}. ``Thresh'' is the stopping tolerance for Sinkhorn (based on $\|\bm{u}^{(l+1)}-\bm{u}^{(l)}\|_2$). }
\label{tab:converge_general}
\end{table}

\begin{figure}
    \centering
    \includegraphics[width=0.39\linewidth]{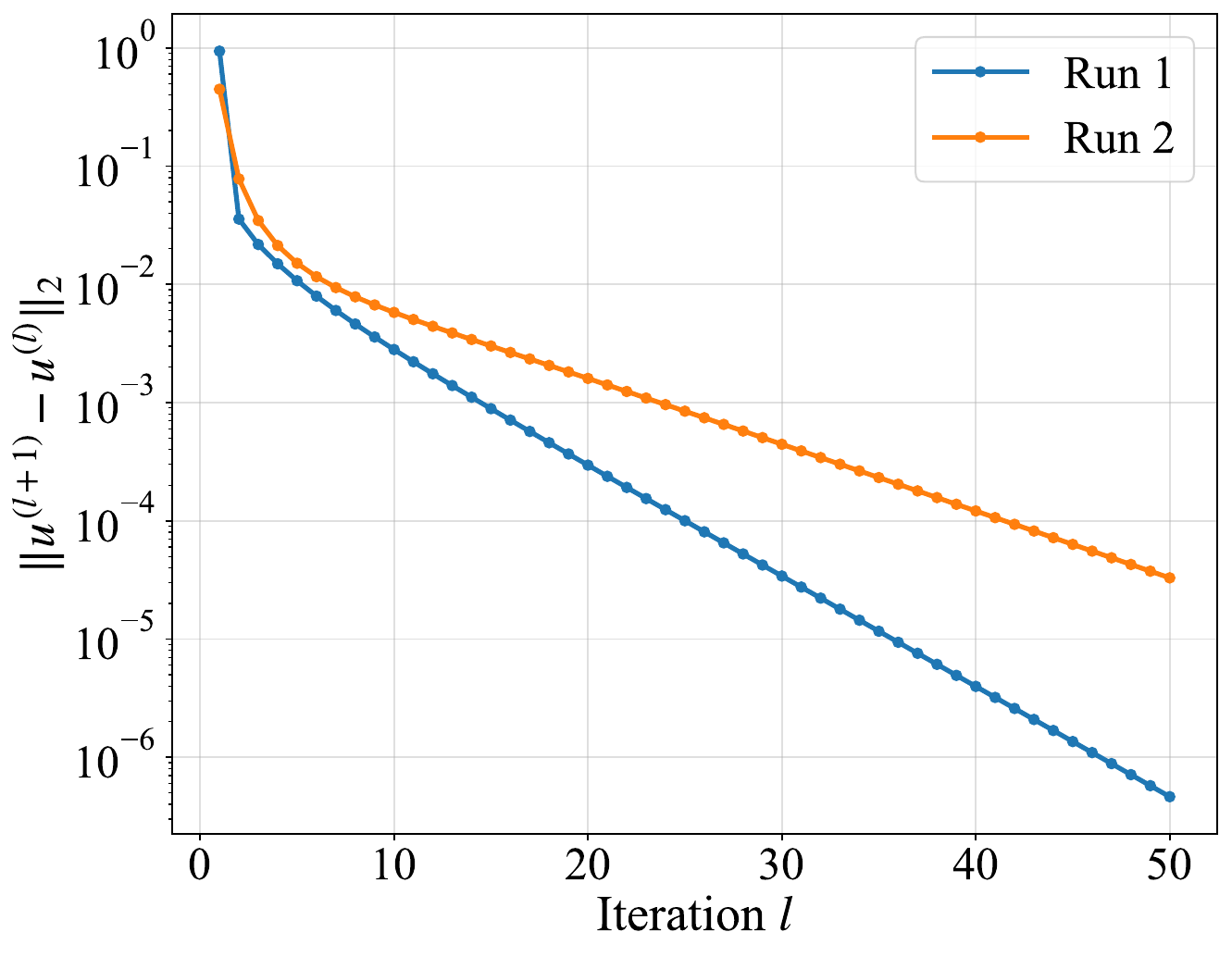}
    \caption{Example curves of $\|\bm{u}^{(l+1)}-\bm{u}^{(l)}\|_2$ in two random runs for the general mask defined in Proposition~\ref{thm:thm_mask}.}
    \label{fig:curves_general}
\end{figure}

% Note that if there exists a keypoint pair \((i,j)\in\mathcal K\) with \(p_i\neq q_j\) (w.l.o.g., assume \(p_i>q_j\)), the total-support condition may be violated. To enable the use of Sinkhorn’s algorithm, we introduce an additional source point \(x_{m+1}\) as a copy of \(x_i\) (i.e., \(x_{m+1}=x_i\)). We then split the source mass \(p_i\) by assigning mass \(q_j\) to \(x_i\) and the remaining mass \(p_i-q_j\) to \(x_{m+1}\). With this construction, Sinkhorn’s algorithm becomes applicable.
}
\subsection{Frank-Wolfe Algorithm for Solving (Partial) KPG-RL-GW}
\textit{Frank-Wolfe algorithm for solving KPG-RL-GW.} We define the 4-order tensor $L=(L_{i,j,k,l})\in \mathbb{R}^{m\times n\times m\times n}$ by $L_{i,j,k,l} = (C_{i,k}^s-C^t_{j,l})^2$, and define the tensor-matrix product $L\circ \pi = ((L\circ \pi)_{i,j})\in \mathbb{R}^{m\times n}$ by $(L\circ \pi)_{i,j} = \sum_{k,l}L_{i,j,k,l}\pi_{k,l}$.
% \begin{equation}
% (L\otimes \pi)_{i,j} = \sum_{k,l}L_{i,j,k,l}\pi_{k,l}.
% \end{equation}
Then, the KPG-RL-GW model in Eq.~\eqref{eq:KPG_RL_GW} in the paper reads 
\begin{equation}
\begin{split}
    &\min_{\pi\in\Pi(\bm{p},\bm{q};M)} \langle(M\odot\pi), \alpha L\circ (M\odot\pi)+ (1-\alpha)G\rangle_F \triangleq \mathcal{L}(\pi) \\
    % &s.t.  \hspace{0.2cm}\pi\geq 0, (M\odot \pi)\mathbbm{1}_{n} = \bm{p},
    %     (M\odot \pi)^{\top}\mathbbm{1}_{m} = \bm{q}.
\end{split}
\end{equation}
The gradient of the objective function is
\begin{equation}
    \nabla \mathcal{L}(\pi) = M\odot(2\alpha L\circ (M\odot\pi) + (1-\alpha) G).
\end{equation}
In the $k$-th iteration of the Frank-Wolfe algorithm, it runs the following three steps:

\textbf{Step 1.} Compute a linear minimization oracle over the set $\Pi(\bm{p},\bm{q};M)$, \ie, 
\begin{equation}\label{eq:project_grad}
    \hat{\pi}\leftarrow\mathop{\mbox{argmin}}\limits_{\pi\in\Pi(\bm{p},\bm{q};M)} \langle \nabla \mathcal{L}(\pi^{(k)}), \pi\rangle_F.
\end{equation}
Equation~\eqref{eq:project_grad} can be rewritten as 
\begin{equation}\label{eq:project_grad1}
    \hat{\pi}\leftarrow\mathop{\mbox{argmin}}\limits_{\pi\in\Pi(\bm{p},\bm{q};M)} \langle M\odot\pi, 2\alpha L\circ(M\odot\pi^{(k)}) + (1-\alpha) G \rangle_F,
\end{equation}
Equation~\eqref{eq:project_grad1} is a KPG-RL-like problem and can be solved using linear programming or Sinkhorn's algorithm.

\textbf{Step 2.} Determine optimal step-size $\beta^{(k)}$ subject to
\begin{equation}\label{eq:step_size}
    \beta^{(k)}\leftarrow\mathop{\mbox{argmin}}\limits_{\beta\in [0,1]} \mathcal{L}((1-\beta)\pi^{(k)}+\beta\hat{\pi}).
\end{equation}
$\beta^{(k)}$ can be obtained by the line search method as in~\citep{titouan2019optimal}.

\textbf{Step 3.} Update 
\begin{equation}
    \pi^{(k+1)} = (1-\beta^{(k)})\pi^{(k)}+\beta^{(k)}\hat{\pi}.
\end{equation}

\vspace{0.4\baselineskip}\textit{Frank-Wolfe algorithm for solving partial KPG-RL-GW.}
The Frank-Wolfe algorithm for solving partial KPG-RL-GW (discussed in Sect.~\ref{sec:ex_partial}) is the same as that for KPG-RL-GW discussed above, except that the solution set in Step 1 is $\Pi^s(\bm{p},\bm{q};M)$. Therefore, $\hat{\pi}$ is obtained by solving a partial KPG-RL-like model, \ie,
\begin{equation}\label{eq:project_grad2}
    \hat{\pi}\leftarrow\mathop{\mbox{argmin}}\limits_{\pi\in\Pi^s(\bm{p},\bm{q};M)} \langle M\odot\pi, 2\alpha L\circ(M\odot\pi^{(k)}) + (1-\alpha) G \rangle_F,
\end{equation}
which is solved similarly to problem~\eqref{eq:partial_kpp_main}.

\subsection{Algorithm for Solving Unbalanced KPG-RL-GW}
Similar to the log-domain Sinkhorn iteration in Appendix B.2, we provide the generalized Sinkhorn iteration (Eqs.~\eqref{eq:generalized_sinkhorn}) for unbalanced KPG-RL (Eq.~\eqref{eq:unbalanced_kpg_rl}) in log-domain as follows:
\begin{equation}\label{eq:log_fg_u}
\begin{split}
    \bm{f}^{(l+1)} =& \bm{\rho}\odot\left[\epsilon\log(\bm{p}) -\epsilon\mbox{LogSumExp}\left(\bar{H}_G(\bm{f}^{(l)},\bm{g}^{(l)})\right)+\bm{f}^{(l)}\right]\\
    \bm{g}^{(l+1)} =& \bm{\rho}\odot\left[\epsilon\log(\bm{q}) -\epsilon\mbox{LogSumExp} \left(\bar{H}_G(\bm{f}^{(l+1)},\bm{g}^{(l)})^{\top}\right)+\bm{g}^{(l)}\right].
\end{split}
\end{equation}

% For convenience, we denote $L_{kgw}(\pi) = \alpha L_{gw}(\tilde{\pi}) + (1-\alpha)L_{kpg}(\pi) + \mu_1 \Bar{\mathcal{D}}^{\otimes}(\tilde{\pi}_1\times\tilde{\pi}_1|\bm{p}\times\bm{p}) + \mu_2 \Bar{\mathcal{D}}^{\otimes}(\tilde{\pi}_2\times\tilde{\pi}_2|\bm{q}\times\bm{q})$. 
Following unbalanced GW~\citep{sejourne2021unbalanced}, we optimize the following relaxation of unbalanced KPG-RL-GW with entropic regularization:
\begin{equation}
    \begin{split}
        F_{kgw}(\pi,\gamma) = &\alpha \sum_{i,k=1}^m\sum_{j,l=1}^n \tilde{\pi}_{i,j}\tilde{\gamma}_{k,l}\vert C^s_{i,k}-C^t_{j,l}\vert^2 + (1-\alpha)\sum_{i,j}G_{i,j}\tilde{\pi}_{i,j} \\ 
        &+ \mu_1 \Bar{\mathcal{D}}^{\otimes}(\tilde{\pi}_1\times\tilde{\gamma}_1|\bm{p}\times\bm{p}) + \mu_2 \Bar{\mathcal{D}}^{\otimes}(\tilde{\pi}_2\times\tilde{\gamma}_2|\bm{q}\times\bm{q}) + \epsilon E(\tilde{\pi}\times\tilde{\gamma}),
    \end{split}
\end{equation}
where $\tilde{\pi}=M\odot\pi$ and $\tilde{\gamma}=M\odot\gamma$. Inspired by~\citep{sejourne2021unbalanced}, we alternately update $\pi$ and $\gamma$ as in Algorithm~\ref{alg:kpg-rl-gw}. We show empirically the effectiveness of the algorithm in Sect.~\ref{sec:single_cell}.   
\begin{algorithm}[t]
 \caption{Algorithm for Unbalanced KPG-RL-GW.}\label{alg:kpg-rl-gw}
    \begin{algorithmic}
    \STATE Let $\zeta(\pi)=\sum_{i,j}\pi_{i,j}$ be the total mass of $\pi$
    \STATE Initialization: $\tilde{\pi} = \tilde{\gamma} = \bm{p}\times \bm{q}$
    \WHILE{$\tilde{\gamma}$ not converged}
    \STATE $\tilde{\gamma} = \tilde{\pi}$
    % \STATE $\tilde{\pi} = M\odot\pi, \tilde{\gamma} = M\odot\gamma$
    \STATE $\tilde{\epsilon} = \zeta(\tilde{\gamma})\epsilon, \tilde{\bm{\rho}} = \zeta(\tilde{\gamma})\bm{\rho}$
    \STATE Compute $\tilde{C} = (\tilde{C}_{i,j})\in\mathbb{R}^{m\times n},  \tilde{C}_{i,j}= \alpha \sum_{k,l}|C^s_{i,k}-C^t_{j,l}|^2\tilde{\gamma}_{k,l} + (1-\alpha)G_{i,j} + \mu\bar{\mathcal{D}}(\tilde{\gamma}\mathbbm{1}_n|\bm{p})+\mu\bar{\mathcal{D}}(\tilde{\gamma}^\top\mathbbm{1}_m|\bm{q}) + \epsilon \bar{\mathcal{D}}(\tilde{\gamma}|\bm{p}\times\bm{q})$
    \FOR{$l=1,2,\cdots$}
    \STATE $\bm{f}^{(l+1)} = \bm{\tilde{\rho}}\odot\left[\epsilon\log(\bm{p}) -\epsilon\mbox{LogSumExp}\left(\bar{H}_{\tilde{C}}(\bm{f}^{(l)},\bm{g}^{(l)})\right)+\bm{f}^{(l)}\right]$
    \STATE $\bm{g}^{(l+1)} = \bm{\tilde{\rho}}\odot\left[\epsilon\log(\bm{q}) -\epsilon\mbox{LogSumExp} \left(\bar{H}_{\tilde{C}}(\bm{f}^{(l+1)},\bm{g}^{(l)})^{\top}\right)+\bm{g}^{(l)}\right]$
    \ENDFOR
    \STATE $\bm{u}= e^{\bm{f}/\tilde{\epsilon}},\bm{v} = e^{\bm{g}/\tilde{\epsilon}}, K = M\odot e^{-\tilde{C}/\tilde{\epsilon}}, \tilde{\pi} = \mbox{diag}(\bm{u})K\mbox{diag}(\bm{v})$
    \STATE $\tilde{\pi} = \sqrt{\zeta(\tilde{\gamma})/\zeta(\tilde{\pi})}\tilde{\pi}$
    \ENDWHILE
    \RETURN $\tilde{\gamma}$
\end{algorithmic}
\end{algorithm}

% \begin{algorithm}[H]
%     \SetAlgoLined
%     \KwData{this text}
%     \KwResult{how to write algorithm with \LaTeX2e }
%     initialization\;
%     \While{not at end of this document}{
%         read current\;
%         \eIf{understand}{
%             go to next section\;
%             current section becomes this one\;
%             }{
%             go back to the beginning of current section\;
%         }
%     }
% \caption{Algorithm for Unbalanced KPG-RL-GW}
% \end{algorithm}

% the unbalanced KPG-RL-GW can be solved by Algorithm~??. Note that Algorithm~?? optimize a lower bound of unbalanced KPG-RL-GW as in~\citep{sejourne2021unbalanced}, of which the tightness is left as future work. 
% \begin{algorithm}[t]
%     \caption{Algorithm for Unbalanced KPG-RL-GW}
    
% \end{algorithm}

% \begin{algorithm}[H]\label{alg:alg0}
%   \caption{Algorithm for Unbalanced KPG-RL-GW}
%   \SetAlgoLined
%   \KwIn{Distributions $\bm{p},\bm{q}$, }
%   \KwOut{Learned potentials $u_{\hat{\omega}},v_{\hat{\omega}}$}
%   \For{$ {\rm iter} = 1,\cdots,N_{\rm iter}'$}{
%     Sampling mini-batch data $\bm{X}=\{\x_b\}_{b=1}^{B'}$ from $p$, $\bm{Y}=\{\y_b\}_{b=1}^{B'}$ from $q$\;
%     \eIf{paired data are available}
%     {Computing the loss of semi-supervised OT in Eq.~(6) on $\bm{X}$ and $\bm{Y}$\;}
%     {Computing the loss of unsupervised OT in Eq.~(6) on $\bm{X}$ and $\bm{Y}$\;}
%     Backward propagation to compute the gradient and update $\omega$ using Adam algorithm\;
%     }
%   $\hat{\omega} = \omega$. 
% \end{algorithm}

\section{Proof of Theorems~\ref{thm:kpg_rl_kp} and~\ref{thm:kpg_rl_gw}}
In this Appendix, for convenience in description, we denote the mask matrix for distributions $\bm{p}$ and $\bm{q}$ as $M^{\bm{p}\bm{q}}$. Note that Theorems~\ref{thm:kpg_rl_kp} and~\ref{thm:kpg_rl_gw} are for the case where $p_{i}=q_j$ for all $(i,j)\in\mathcal{K}$.
\subsection{Proof of Theorem~\ref{thm:kpg_rl_kp}}
We verify the conditions of the proper metric as follows.

\vspace{0.4\baselineskip}\textit{(1) Prove that $\mathcal{S}_{krk}(\bm{p},\bm{q})=0$ if and only if $\bm{p}=\bm{q}$. }
\quad 

(a) If $\bm{p}=\bm{q}$, we have $x_i=y_i$ and $p_i=q_i$ for any $i\in[m]$ (since the permutation of support points does not change the distribution). Hence, $C_{i,i}=c(x_i,y_i)=0$, and $C^s_{i,i_u}=c(x_i,x_{i_u})=c(y_i,y_{i_u})=C^t_{i,i_u}, \forall i\in[m] \mbox{ and }\forall i_u\in\mathcal{I}$, which implies that $R^s_i=R^t_i$. Then, we have $G_{i,i}=d(R^s_i,R^t_i)=0$. We define $\pi$ by $\pi_{i,j}=p_i$ if $i=j$, otherwise 0. Obviously, $M^{\bm{p}\bm{q}}\odot\pi$ is in $\Pi(\bm{p},\bm{q};M^{\bm{p}\bm{q}})$ and $\sum_{i,j}M^{\bm{p}\bm{q}}_{i,j}\pi_{i,j}(\alpha C_{i,j}+(1-\alpha)G_{i,j}) = 0$. Therefore, $\mathcal{S}_{krk}(\bm{p},\bm{q})=0$. 

(b) We denote $\pi^*$ as the optimal solution of problem~\eqref{eq:s_krk}. If $\mathcal{S}_{krk}(\bm{p},\bm{q})=0$, we have $\langle M^{\bm{p}\bm{q}}\odot\pi^* ,C \rangle_F=0$. This means that the KP problem $\min_{\pi\in\Pi(\bm{p},\bm{q})}\langle \pi ,C \rangle_F = 0$. Using the Proposition 2.2 in \citep{peyre2019computational}, we have $\bm{p}=\bm{q}$.

\vspace{0.4\baselineskip}\textit{(2) Prove that $\mathcal{S}_{krk}(\bm{p},\bm{q})=\mathcal{S}_{krk}(\bm{q},\bm{p})$. }\quad
% Since the paired keypoints across domains share the same index, the mask matrix $M^{\bm{p}\bm{q}}$ is symmetric. Thus, $M^{\bm{p}\bm{q}} = (M^{\bm{p}\bm{q}})^{\top}= M^{\bm{q}\bm{p}}$.

From the definition of mask matrix in Proposition~\ref{thm:partial_kpp_main}, we have $M^{\bm{p}\bm{q}}_{i,j}= M^{\bm{q}\bm{p}}_{j,i}$.
$C$ and $G$ are symmetric because $c$ and $d$ are distances.  For any $\pi \in \Pi(\bm{p},\bm{q};M^{\bm{p}\bm{q}})$, we define $\pi'$ as $\pi'_{i,j}=\pi_{j,i}$, and then $\pi' \in \Pi(\bm{q},\bm{p};M^{\bm{q}\bm{p}})$.
Then, we have 
\begin{equation}
    \begin{split}
    \mathcal{S}_{krk}(\bm{p},\bm{q})=
    &\min_{\pi \in \Pi(\bm{p},\bm{q};M^{\bm{p}\bm{q}})}\sum_{i,j}{M^{\bm{p}\bm{q}}_{i,j}\pi_{i,j}(\alpha C_{i,j}+(1-\alpha)G_{i,j})}\\ =&\min_{\pi \in \Pi(\bm{p},\bm{q};M^{\bm{p}\bm{q}})}\sum_{i,j}{M^{\bm{q}\bm{p}}_{j,i}\pi_{i,j}(\alpha C_{j,i}+(1-\alpha)G_{j,i})}\\
= &\min_{\pi' \in \Pi(\bm{q},\bm{p};M^{\bm{q}\bm{p}})}\sum_{j,i}{M^{\bm{q}\bm{p}}_{j,i}\pi'_{j,i}(\alpha C_{j,i}+(1-\alpha)G_{j,i})}\\
=&\mathcal{S}_{krk}(\bm{q},\bm{p}).
    \end{split}
\end{equation}
% Therefore, $\mathcal{S}_{krk}(\bm{p},\bm{q})=\mathcal{S}_{krk}(\bm{q},\bm{p})$.
% $. By defining $\pi'$ as $\pi'_{i,j}=\pi_{j,i}$, we have $\sum_{i,j}{M^{\bm{p}\bm{q}}_{i,j}\pi_{i,j}(\alpha C_{i,j}+(1-\alpha)G_{i,j})} 

\vspace{0.4\baselineskip}\textit{(3) Prove that $\mathcal{S}_{krk}(\bm{p},\bm{q})\leq\mathcal{S}_{krk}(\bm{p},\bm{r})+\mathcal{S}_{krk}(\bm{r},\bm{q})$ for any $\bm{r}=\sum_{k=1}^mr_k\delta_{z_k}\in \mathcal{P}_{\mathcal{I}}^\mathcal{X}$.} \quad 

Let $M^{\bm{p}\bm{r}}\odot\pi^{\bm{p}\bm{r}}$ and $M^{\bm{r}\bm{q}}\odot\pi^{\bm{r}\bm{q}}$ be the optimal transport plans corresponding to $\mathcal{S}_{krk}(\bm{p},\bm{r})$ and $\mathcal{S}_{krk}(\bm{r},\bm{q})$, respectively. We define \begin{equation}
    \tilde{\gamma}=(M^{\bm{p}\bm{r}}\odot\pi^{\bm{p}\bm{r}})\mbox{diag}\left(\frac{1}{\tilde{\bm{r}}}\right) (M^{\bm{r}\bm{q}}\odot\pi^{\bm{r}\bm{q}}),
\end{equation}
where the element $\tilde{r}_j$ of $\tilde{\bm{r}}$ is $r_j$ if $r_j>0$, and 1 otherwise. We notice that 
\begin{equation}
    \tilde{\gamma}\mathbbm{1}_m =(M^{\bm{p}\bm{r}}\odot\pi^{\bm{p}\bm{r}})\mbox{diag}\left(\frac{1}{\tilde{\bm{r}}}\right)\bm{r}= (M^{\bm{p}\bm{r}}\odot\pi^{\bm{p}\bm{r}})\left(\frac{\bm{r}}{\tilde{\bm{r}}}\right) = (M^{\bm{p}\bm{r}}\odot\pi^{\bm{p}\bm{r}})\tilde{\mathbbm{1}}_{m},
\end{equation}
where the $j$-th location of $\tilde{\mathbbm{1}}_{m}$ is 1 if $r_j>0$, otherwise 0. Note that for $j$ such that $r_j=0$, we have $\sum_{i}(M^{\bm{p}\bm{r}}\odot\pi^{\bm{p}\bm{r}})_{i,j}=r_j=0$, which implies $(M^{\bm{p}\bm{r}}\odot\pi^{\bm{p}\bm{r}})_{i,j}=0$ for any $i$. Hence,
\begin{equation}
    \tilde{\gamma}\mathbbm{1}_m = (M^{\bm{p}\bm{r}}\odot\pi^{\bm{p}\bm{r}})\tilde{\mathbbm{1}}_{m} = (M^{\bm{p}\bm{r}}\odot\pi^{\bm{p}\bm{r}}){\mathbbm{1}}_{m} = \bm{p}.
\end{equation}
Similarly, $\tilde{\gamma}^\top\mathbbm{1}_m=\bm{q}$. Since the indexes of paired keypoints across any two distribution in $\mathcal{P}^{\mathcal{X}}_{\mathcal{I}}$ are the same, we have for any $i_u\in\mathcal{I}$, the $i_u$-th row and column of $M^{\bm{p}\bm{r}}$ and $M^{\bm{r}\bm{q}}$ are zeros except for that $M^{\bm{p}\bm{r}}_{i_u,i_u} = M^{\bm{r}\bm{q}}_{i_u,i_u} =1$. So the $i_u$-th row and column of $\tilde{\gamma}$ are zeros except for $\tilde{\gamma}_{i_u,i_u}$.
Then, we can write $\tilde{\gamma}=M^{\bm{p}\bm{q}}\odot\gamma$ with $\gamma\in\mathbb{R}_+^{m\times m}$. Therefore, we have $\gamma\in\Pi(\bm{p},\bm{q};M^{\bm{p}\bm{q}})$. 
% We denote $D^ = \alpha C + (1-\alpha)G$. We can verify that 
% \begin{equation}
% \begin{split}
%         D_{i,j} =& \alpha C_{i,j}+ (1-\alpha)G_{i,j}
%         =\alpha c(x_i,y_j)+ (1-\alpha)d(R_i^s,R_j^t)\\
%         \leq&\alpha c(x_i,z_ky_j)+ (1-\alpha)d(R_i^s,R_j^t)
% \end{split}
% \end{equation}
The triangle inequality then follows from
\begin{equation}
    \begin{split}
       &\mathcal{S}_{krk}(\bm{p},\bm{q}) = \min_{\pi \in \Pi(\bm{p},\bm{q};M^{\bm{p}\bm{q}})}  \sum_{i,j}{M^{\bm{p}\bm{q}}_{i,j}\pi_{i,j}(\alpha c(x_i,y_j)+(1-\alpha)d(R^s_i,R^t_j))}\\
       \leq&\sum_{i,j}{\tilde{\gamma}_{i,j}(\alpha c(x_i,y_j)+(1-\alpha)d(R^s_i,R^t_j))}\\
    %   \leq&\sum_{i,k,j}{\tilde{\gamma}(\alpha (c(x_i,z_k)+c(z_k,y_j)) +(1-\alpha)(d(R^s_i,R^r_k)+d(R^r_k,R^t_j))}\\
       =&\sum_{i,j}(\alpha c(x_i,y_j) + (1-\alpha)d(R^s_i,R^t_j))
       \sum_{k}\frac{(M^{\bm{p}\bm{r}}\odot\pi^{\bm{p}\bm{r}})_{i,k}(M^{\bm{r}\bm{q}}\odot\pi^{\bm{r}\bm{q}})_{k,j}}{\tilde{r}_k}\\
%     \end{split}
%     \end{equation}
% \begin{equation}
%     \begin{split}
    %   (M^{\bm{p}\bm{r}}\odot\pi^{\bm{p}\bm{r}})\mbox{diag}\left(\frac{1}{\tilde{\bm{r}}}\right) (M^{\bm{r}\bm{q}}\odot\pi^{\bm{r}\bm{q}})]_{i,j}\\
    %   +&\sum_{i,k,j}(M^{\bm{p}\bm{r}}\odot\pi^{\bm{p}\bm{r}})\mbox{diag}\left(\frac{1}{\tilde{\bm{r}}}\right) (M^{\bm{r}\bm{q}}\odot\pi^{\bm{r}\bm{q}})]_{i,j}(\alpha c(z_k,y_j) + (1-\alpha)d(R^r_k,R^t_j))\\
       \leq&\sum_{i,k,j}(\alpha (c(x_i,z_k)+c(z_k,y_j)) +(1-\alpha)(d(R^s_i,R^r_k)+d(R^r_k,R^t_j))
       \frac{(M^{\bm{p}\bm{r}}\odot\pi^{\bm{p}\bm{r}})_{i,k}(M^{\bm{r}\bm{q}}\odot\pi^{\bm{r}\bm{q}})_{k,j}}{\tilde{r}_k}\\
       =&\sum_{i,k,j}(\alpha c(x_i,z_k) + (1-\alpha)d(R^s_i,R^r_k))\frac{(M^{\bm{p}\bm{r}}\odot\pi^{\bm{p}\bm{r}})_{i,k}(M^{\bm{r}\bm{q}}\odot\pi^{\bm{r}\bm{q}})_{k,j}}{\tilde{r}_k}\\
       +& \sum_{i,k,j}(\alpha c(z_k,y_j) + (1-\alpha)d(R^r_k,R^t_j))\frac{(M^{\bm{p}\bm{r}}\odot\pi^{\bm{p}\bm{r}})_{i,k}(M^{\bm{r}\bm{q}}\odot\pi^{\bm{r}\bm{q}})_{k,j}}{\tilde{r}_k}\\
       =& \sum_{i,k}(\alpha c(x_i,z_k) + (1-\alpha)d(R^s_i,R^r_k))(M^{\bm{p}\bm{r}}\odot\pi^{\bm{p}\bm{r}})_{i,k}\\
       +& \sum_{k,j}(\alpha c(z_k,y_j) + (1-\alpha)d(R^r_k,R^t_j))(M^{\bm{r}\bm{q}}\odot\pi^{\bm{r}\bm{q}})_{k,j}\\
       = &\mathcal{S}_{krk}(\bm{p},\bm{r}) + \mathcal{S}_{krk}(\bm{r},\bm{q}),
    \end{split}
\end{equation}
where $z_k$ is the support point of $\bm{r}$ and $R^r_k$ is the relation of $z_k$ to the keypoints of $\bm{r}$.

\hfill$\blacksquare$

\subsection{Proof of Theorem~\ref{thm:kpg_rl_gw}}\quad

(a) If $\bm{p}$ and $\bm{q}$ are isomorphic, for any $i\in [m]$ and any $i_u\in\mathcal{I}$, we have $c(x_i,x_{i_u})=c'(y_{\sigma(i)},y_{\sigma(i_u)})=c'(y_{\sigma(i)},y_{i_u})$, implying that $R_i^s=R_{\sigma(i)}^t$. We define $\pi$ as $\pi_{i,j}=p_i$ if $j=\sigma(i)$, otherwise 0. 
We then have 
\begin{equation}
    \begin{split}
        &\sum_{i,j}\alpha\Big(\sum_{k,l}(M^{\bm{p}\bm{q}}\odot\pi)_{i,j}(M^{\bm{p}\bm{q}}\odot\pi)_{k,l}|C^s_{i,k}-C^t_{j,l}|^2\Big)+(1-\alpha)(M^{\bm{p}\bm{q}}\odot\pi)_{i,j}G_{i,j}\\
        =&\sum_{i}\Big[\alpha\Big(\sum_{k}(M^{\bm{p}\bm{q}}\odot\pi)_{i,\sigma(i)}(M^{\bm{p}\bm{q}}\odot\pi)_{k,\sigma(k)}|C^s_{i,k}-C^t_{\sigma(i),\sigma(k)}|^2\Big)\\
        &\hspace{0.8cm}+(1-\alpha)(M^{\bm{p}\bm{q}}\odot\pi)_{i,\sigma(i)}G_{i,\sigma(i)}\Big]\\
%         \end{split}
%     \end{equation}
% \begin{equation}
%     \begin{split}
        =&\sum_{i}\Big[\alpha\Big(\sum_{k}(M^{\bm{p}\bm{q}}\odot\pi)_{i,\sigma(i)}(M^{\bm{p}\bm{q}}\odot\pi)_{k,\sigma(k)}|c(x_i,x_k)-c'(x_{\sigma(i)},x_{\sigma(k)})|^2\Big)\\
        &\hspace{0.8cm}+(1-\alpha)(M^{\bm{p}\bm{q}}\odot\pi)_{i,\sigma(i)}d(R^s_i,R^t_{\sigma(i)})\Big] \\
        =& 0.
    \end{split}
\end{equation}
This implies $\mathcal{S}_{krg}(\bm{p},\bm{q})=0$.

(b) Let $(M^{\bm{p}\bm{q}})\odot \pi^*$ be the optimal transport plan corresponding to $\mathcal{S}_{krg}(\bm{p},\bm{q})$. If $\mathcal{S}_{krg}(\bm{p},\bm{q})=0$, we have \begin{equation}
    \sum_{i,j,k,l}(M^{\bm{p}\bm{q}}\odot\pi^*)_{i,j}(M^{\bm{p}\bm{q}}\odot\pi^*)_{k,l}|C^s_{i,k}-C^t_{j,l}|^2 = 0. 
\end{equation}
This indicates that the Gromov-Wasserstein distance 
\begin{equation}
\min_{\pi\in\Pi(\bm{p},\bm{q})}\sum_{i,j,k,l}\pi_{i,j}\pi_{k,l}|C^s_{i,k}-C^t_{j,l}|^2 = 0. 
\end{equation}
By virtue to Gromov-Wasserstein properties in \cite{memoli2011gromov}, there exists a bijection $\sigma:[m]\longmapsto[m]$ such that $c(x_i,x_k)=c'(y_{\sigma(i)},y_{\sigma(k)})$, and $p_i=q_{\sigma(i)}$.

\hfill$\blacksquare$

\section{Proof and Generalization of Theorem~\ref{thm:partial_kpp_main}}
For convenience of understanding, Theorem~\ref{thm:partial_kpp_main} in the paper is for the case that $p_i=q_j$ for all $(i,j)\in\mathcal{K}$. In this appendix, we provide and prove Theorem~\ref{thm:partial_kpp}, a generalization of Theorem~\ref{thm:partial_kpp_main} in the paper, for the general case that there could exist some $(i,j)\in\mathcal{K}$ such that $p_i\neq q_j$. 
% Before that, we rewrite the partial-KPG-RL model first.

% \textbf{Partial-KPG-RL model:}
% \begin{equation}\label{eq:partial_kpp}
% \min_{\pi \in \Pi^{s}(\bm{p},\bm{q};M)}\left\{L_{kpg}(M\odot\pi) = \langle M\odot\pi , G \rangle_F\right\},
% \end{equation}
% where $\Pi^{s}(\bm{p},\bm{q};M) = \{ \pi \in \mathbb{R}^{m\times n}_+ \vert (M\odot\pi) \mathbbm{1}_{n} \leqslant \bm{p}, (M\odot\pi)^{\top} \mathbbm{1}_{m} \leqslant \bm{q}, \mathbbm{1}_{m}^{\top}(M\odot\pi) \mathbbm{1}_{n} = s; (M\odot\pi)_{i,:}\mathbbm{1}_n = p_i, \forall i \in \mathcal{I}; \mathbbm{1}^{\top}_m(M\odot\pi)_{:,j} = q_j, \forall j \in \mathcal{J}\}$.
% As stated in the paper, the above partial-KPG-RL model is for the case that $p_i=q_j$ for all $(i,j)\in\mathcal{K}$. For the case that there exist some $(i,j)\in\mathcal{K}$ such that $p_i=q_j$, its corresponding partial-KPG-RL model is left as our future work.
% \vspace{0.2\baselineskip}

\begin{theorem}\label{thm:partial_kpp}
Suppose $A>0$, $\xi>0$, $\sum_{i\in \mathcal{I}} p_i+\max\{q_{\kappa(i)}-p_i,0\} <s$, and $\sum_{j\in \mathcal{J}} q_j+\max\{p_{\kappa'(j)}-q_j,0\} < s$, then the optimal transport plan $M\odot\pi^*$ of the partial KPG-RL model in Eq.~\eqref{eq:partial_kpp_main} is the $m$-by-$n$ block in the upper left corner of the optimal transport plan $\bar{M}\odot\bar{\pi}^*$ of problem $\mbox{min}_{\bar{\pi} \in \Pi(\bar{\bm{p}},\bar{\bm{q}};\bar{M})} \langle \bar{M}\odot\bar{\pi} , \bar{G} \rangle_F$.
\end{theorem}

The definitions of $\kappa(i)$ and $\kappa'(j)$ are given in Appendix A. The condition $\sum_{i\in \mathcal{I}} p_i+\max\{q_{\kappa(i)}-p_i,0\} <s$ implies that the sum of the mass ($\sum_{i\in \mathcal{I}} p_i$) of the source key points and the mass ($\sum_{i\in \mathcal{I}}\max\{q_{\kappa(i)}-p_i,0\}$) of the other source points apart from the key points that should be transported to the target key points is less than $s$. The condition $\sum_{j\in \mathcal{J}} q_j+\max\{p_{\kappa'(j)}-q_j,0\} < s$ implies that the sum of the mass ($\sum_{j\in \mathcal{J}}q_j$) of target keypoints and the mass ($\sum_{j\in \mathcal{J}} \max\{p_{\kappa'(j)}-q_j,0\}$) of the other target points apart from keypoints received from source keypoints is less than $s$. The two conditions are reasonable to guarantee the admissible solutions of problem~\eqref{eq:partial_kpp_main}. If $p_i=q_j$ for all $(i,j)\in\mathcal{K}$ (\ie, $p_i=q_{\kappa(i)},\forall i\in\mathcal{I}$ and $q_j=p_{\kappa'(j)},\forall j\in\mathcal{J}$), Theorem~\ref{thm:partial_kpp} degenerates into Theorem~\ref{thm:partial_kpp_main}.

\subsection{Proof of Theorem~\ref{thm:partial_kpp}} 
We denote $\ddot{\pi}=\bar{\pi}^*_{1:m,1:n}$ and $t = \bar{\pi}^*_{m+1,n+1}$. 
To prove Theorem~\ref{thm:partial_kpp}, we first make some preparations and then perform the following three steps. In Step 1, we prove that $t =0$. In Step 2, we prove that $\ddot{\pi} \in \Pi^{s}(\bm{p},\bm{q};M)$, which means that $\ddot{\pi}$ is a feasible solution to problem~\eqref{eq:partial_kpp_main}. In Step 3, we prove that $M\odot\ddot{\pi} $ is the optimal transport plan of problem~\eqref{eq:partial_kpp_main}. We next detail these steps.

\vspace{0.4\baselineskip}\textit{Preparations.}

Since $\bar{\pi}^* \in \Pi(\bar{\bm{p}},\bar{\bm{q}};\bar{M})$ and $t=\bar{\pi}^*_{m+1,n+1}$, we have 
\begin{equation}
\begin{split}
  \mathbbm{1}_{m+1}^{\top}(\bar{M}\odot\bar{\pi}^*)\mathbbm{1}_{n+1}=&
\left[
\begin{matrix}
    \mathbbm{1}_m^{\top} &1
\end{matrix}\right] \left(\left[
\begin{matrix}
    M & \bm{a}\\
    \bm{b} & 1
\end{matrix}\right]\odot\left[
\begin{matrix}
    \ddot{\pi} & \bar{\pi}^*_{1:m,n+1}\\
    \bar{\pi}^*_{m+1,1:n} & \bar{\pi}^*_{m+1,n+1}
\end{matrix}\right]\right)\left[\begin{matrix}
    \mathbbm{1}_n\\ 1
\end{matrix}\right]\\ = &\mathbbm{1}_m^{\top}(M\odot\ddot{\pi})\mathbbm{1}_n + \sum_{i=1}^m{a_i\bar{\pi}^*_{i,n+1}} + \sum_{j=1}^n{b_j\bar{\pi}^*_{m+1,j}} + t.
\end{split}
\end{equation}
Meanwhile, we have
\begin{equation}
    \mathbbm{1}_{m+1}^{\top}(\bar{M}\odot\bar{\pi}^*)\mathbbm{1}_{n+1}= \mathbbm{1}_{m+1}^{\top}\bar{\bm{p}} = \Vert \bar{\bm{p}} \Vert_1 = \Vert \bm{p}\Vert_1 + \Vert \bm{q}\Vert_1 -s,
\end{equation}
\begin{equation}
     \sum_{i=1}^m{a_i\bar{\pi}^*_{i,n+1}} + t = \Vert \bm{p}\Vert_1 -s,
\end{equation}
and 
\begin{equation}
    \sum_{j=1}^n{b_j\bar{\pi}^*_{m+1,j}} + t = \Vert \bm{q}\Vert_1 -s.  
\end{equation}
Combining the above four equations, we have
\begin{equation}
    \mathbbm{1}_m^{\top}(M\odot\ddot{\pi})\mathbbm{1}_n + \Vert \bm{p}\Vert_1 + \Vert \bm{q}\Vert_1 -2s -t = \Vert \bm{p}\Vert_1 + \Vert \bm{q}\Vert_1 -s.
\end{equation}
Therefore,
\begin{equation}
    \mathbbm{1}_m^{\top}(M\odot\ddot{\pi})\mathbbm{1}_n = s+t.
\end{equation}

\vspace{0.4\baselineskip}\textit{Step 1: prove that ${t=\bar{\pi}^*_{m+1,n+1} = 0}$.}

First, we have 
\begin{equation}\label{eq:expand_2}
    \begin{split}
        \langle\bar{M}\odot\bar{\pi}^*,\bar{G}\rangle_F = & \sum_{i=1}^{m}\sum_{j=1}^n M_{i,j}\bar{\pi}^*_{i,j}G_{i,j}
        + \xi\sum_{i=1}^ma_i\bar{\pi}^*_{i,n+1}\\ + &\xi \sum_{j=1}^nb_j\bar{\pi}^*_{m+1,j} + (2\xi+A)\bar{\pi}^*_{m+1,n+1}\\
        = & \sum_{i=1}^{m}\sum_{j=1}^n M_{i,j}\bar{\pi}^*_{i,j}G_{i,j} + \xi (\Vert \bm{p}\Vert_1 + \Vert \bm{q}\Vert_1 -2s -2t) + (2\xi+A)t\\
        = & \sum_{i=1}^{m}\sum_{j=1}^n M_{i,j}\bar{\pi}^*_{i,j}G_{i,j} + \xi (\Vert \bm{p}\Vert_1 + \Vert \bm{q}\Vert_1 -2s) + At.
    \end{split}
\end{equation}
Suppose $\bar{\pi}^*_{m+1,n+1}>0$, we next construct a solution $\gamma$ such that $\gamma_{m+1,n+1}=0$ and leads to conflict.
We randomly select a set $S = \{(i,j)|\bar{\pi}^*_{i,j}>0, i\leq m, j\leq n, i\notin \mathcal{I}, j\notin \mathcal{J}\}$ and a index pair $(i_0,j_0)$ satisfying the constraints of elements in $S$, such that $\sum_{(i,j)\in S}\bar{\pi}^*_{i,j}\leq t$ and $\sum_{(i,j)\in S}\bar{\pi}^*_{i,j} + \bar{\pi}^*_{i_0,j_0} > t$. In the rest part of this section, the involved $i,j$ satisfy $i\leq m$ and $j\leq n$.
Such non-empty $S$ and $(i_0,j_0)$ always exist, because 
\begin{equation}
\begin{split}
    \mathbbm{1}_m^{\top}(M\odot\ddot{\pi})\mathbbm{1}_n = &\sum_{i=1}^m\sum_{j=1}^n{M_{i,j}\bar{\pi}^*_{i,j}}
    = \sum_{i\in \mathcal{I},j}\bar{\pi}^*_{i,j} + \sum_{i\notin \mathcal{I},j\in \mathcal{J}}M_{i,j}\bar{\pi}^*_{i,j} + \sum_{i\notin \mathcal{I}, j\notin \mathcal{J}}\bar{\pi}^*_{i,j}  \\
    = &\sum_{i\in \mathcal{I},j}\bar{\pi}^*_{i,j} + \sum_{i\notin \mathcal{I},j\in \mathcal{J}}M_{i,j}\bar{\pi}^*_{i,j} + \sum_{i\notin \mathcal{I}, j\notin \mathcal{J}}\bar{\pi}^*_{i,j}  \\
    % = &\sum_{i\in \mathcal{I},j}\bar{\pi}^*_{i,j} + \sum_{i\notin \mathcal{I},j\in \mathcal{J}}\mathbb{I}(M_{i,j}=1)\bar{\pi}^*_{i,j} + \sum_{i\notin \mathcal{I}, j\notin \mathcal{J}}\bar{\pi}^*_{i,j}  \\
    = &\sum_{i\in \mathcal{I},j}\bar{\pi}^*_{i,j} + \sum_{i\notin \mathcal{I},i'\in \mathcal{I}}M_{i,\kappa(i')}\bar{\pi}^*_{i,\kappa(i')} + \sum_{i\notin \mathcal{I}, j\notin \mathcal{J}}\bar{\pi}^*_{i,j}  \\
    \leq &\sum_{i\in I}p_i + \sum_{i'\in \mathcal{I},i \neq i'}M_{i,\kappa(i')}\bar{\pi}^*_{i,\kappa(i')}+\sum_{i\notin \mathcal{I}, j\notin \mathcal{J}}\bar{\pi}^*_{i,j} \\
    = &\sum_{i\in I}p_i + \sum_{i'\in \mathcal{I}}\max\{q_{\kappa(i')}-p_{i'},0\}+\sum_{i\notin \mathcal{I}, j\notin \mathcal{J}}\bar{\pi}^*_{i,j}, 
\end{split}
\end{equation}
$\mathbbm{1}_m^{\top}(M\odot\ddot{\pi})\mathbbm{1}_n = s+t$,
and $\sum_{i\in \mathcal{I}}p_i + \max\{q_{\kappa(i)}-p_i,0\} < s$, we have $\sum_{i\notin \mathcal{I}, j\notin \mathcal{J}}\bar{\pi}^*_{ij}>t$. We now move the mass of index pairs in $S$ and $(i_0,j_0)$ to their marginal such that a total mass of $t$ is moved. Specifically, 
for $(i,j)\in S$, we set $\gamma_{i,j}  = 0, \gamma_{i,n+1} = \bar{\pi}^*_{i,n+1} + \bar{\pi}^*_{i,j},  \gamma_{m+1,j} = \bar{\pi}^*_{m+1,j} + \bar{\pi}^*_{i,j}$.
For $(i_0,j_0)$, we set $\gamma_{i_0,j_0} = \bar{\pi}^*_{i_0,j_0} - (t-\sum_{i\notin \mathcal{I}, j\notin \mathcal{J}}\bar{\pi}^*_{i,j}), \gamma_{i_0,n+1} = \bar{\pi}^*_{i_0,n+1} + (t-\sum_{i\notin \mathcal{I}, j\notin \mathcal{J}}\bar{\pi}^*_{i,j}),  \gamma_{m+1,j_0} = \bar{\pi}^*_{m+1,j_0}-(t-\sum_{i\notin \mathcal{I}, j\notin \mathcal{J}}\bar{\pi}^*_{i,j})$.
For $(i,j)\notin S$, we set $\gamma_{i,j} = \bar{\pi}^*_{i,j}, \gamma_{i,n+1} = \bar{\pi}^*_{i,n+1},  \gamma_{m+1,j} = \bar{\pi}^*_{m+1,j}$. It is easy to verify that $\gamma\in\Pi(\bar{\bm{p}},\bar{\bm{q}};\bar{M})$. 
Similar to Eq.~\eqref{eq:expand_2}, we have 
\begin{equation}
    \langle\bar{M}\odot\gamma,\bar{G}\rangle_F = \sum_{i=1}^{m}\sum_{j=1}^n M_{i,j}\gamma_{i,j}G_{i,j} + \xi (\Vert \bm{p}\Vert_1 + \Vert \bm{q}\Vert_1 -2s) .
\end{equation}
Using the optimality of $\bar{M}\odot\bar{\pi}^*$, we have
\begin{equation}\label{eq:11}
    \langle\bar{M}\odot\gamma,\bar{G}\rangle_F- \langle\bar{M}\odot\bar{\pi}^*,\bar{G}\rangle_F = \sum_{i=1}^{m}\sum_{j=1}^n M_{i,j}(\gamma_{i,j}-\bar{\pi}^*_{i,j})G_{i,j} - At > 0.
\end{equation}
From the definition of $\gamma$, we can see that $\gamma_{i,j}\leq\bar{\pi}^*_{i,j}$, and thus $\sum_{i=1}^{m}\sum_{j=1}^n M_{i,j}(\gamma_{i,j}-\bar{\pi}^*_{i,j})G_{i,j} \leq 0$. Hence, from Eq.~\eqref{eq:11}, we have $A<0$, which contradicts the assumption that $A>0$. Therefore, $t=\bar{\pi}^*_{m+1,n+1}=0$ holds.

\vspace{0.4\baselineskip}\textit{Step 2: prove that ${\ddot{\pi}}$ is a feasible solution of problem in Eq.~\eqref{eq:partial_kpp_main}.}

We verify the constraints as follows.

(1) Since $\bar{\pi}^*\geq 0$, we have $\ddot{\pi}\geq 0$.

(2) $(\bar{M}\odot\bar{\pi}^*)\mathbbm{1}_{n+1} = \left[
\begin{matrix}
    M\odot\ddot{\pi} & \bm{a}\odot\bar{\pi}^*_{1:m,n+1}\\
    \bm{b}\odot\bar{\pi}^*_{m+1,1:n} & 0
\end{matrix}\right] \left[\begin{matrix}
    \mathbbm{1}_n\\
    1
\end{matrix}\right]
=\left[\begin{matrix}
    \bm{p}\\
    \Vert \bm{q} \Vert_1 -s
\end{matrix}\right]$, then
$(M\odot\ddot{\pi})\mathbbm{1}_{n} + \bm{a}\odot\bar{\pi}^*_{1:m,n+1} = \bm{p}$, and $(M\odot\ddot{\pi})\mathbbm{1}_{n}\leq \bm{p}$.

(3) Similarly, from $\mathbbm{1}_{m+1}^{\top}(\bar{M}\odot\bar{\pi}^{*})= (\bm{q},\Vert \bm{q} \Vert_1 -s)^{\top}$, we have $\mathbbm{1}_{m}^{\top}(M\odot\ddot{\pi})\leq \bm{q}$.

(4) $\mathbbm{1}_{m}^{\top}(M\odot\ddot{\pi})\mathbbm{1}_{n} = s$ holds because $t=0$ as in Step 1.

(5) $\forall i \in \mathcal{I}, (\bar{M}\odot\bar{\pi}^{*})_{i,:}\mathbbm{1}_{n+1} = (M\odot\ddot{\pi})_{i,:}\mathbbm{1}_n + a_i\bar{\pi}^*_{i,n+1} = p_i$. Since $a_i=0$, we have $(M\odot\ddot{\pi})_{1,:}\mathbbm{1}_n = p_i$.

(6) $\forall j \in \mathcal{J}, \mathbbm{1}_{m+1}^{\top}(\bar{M}\odot\bar{\pi}^{*})_{:,j} = \mathbbm{1}_m^{\top}(M\odot\ddot{\pi})_{:,j} + b_j\bar{\pi}^*_{m+1,j} = q_j$. Since $b_j=0$, we have $\mathbbm{1}_m^{\top}(M\odot\ddot{\pi})_{:,j} = q_j$.

Therefore, we have $\ddot{\pi}\in\Pi^s(\bm{p},\bm{q};M)$, and $\ddot{\pi}$ is a feasible solution to the problem in Eq.~\eqref{eq:partial_kpp_main}.

\vspace{0.4\baselineskip}\textit{Step 3: prove that ${M\odot\ddot{\pi}} $ is the optimal transport plan of problem in Eq.~\eqref{eq:partial_kpp_main}. }

Suppose there exist a transport plan $M\odot\gamma$ with $\gamma\in\Pi^s(\bm{p},\bm{q};M)$ such that $$\sum_{i=1}^m\sum_{j=1}^nM_{i,j}\gamma_{i,j}G_{i,j} < \sum_{i=1}^m\sum_{j=1}^nM_{i,j}\ddot{\pi}_{i,j}G_{i,j}.$$ We construct $\bar{\gamma}$ as follows. For $i\leq m, j\leq n$, $\bar{\gamma}_{i,j} = \gamma_{i,j}$. $\bar{\gamma}_{i,n+1} = p_i - \sum_{j=1}^n\gamma_{i,j}, \forall i \leq m$.  $\bar{\gamma}_{m+1,j} = q_j - \sum_{i=1}^n\gamma_{i,j}, \forall j \leq n$. $\bar{\gamma}_{m+1,n+1}=0$. Easily, we can verify that $\bar{\gamma}$ is in $\Pi(\bar{\bm{p}},\bar{\bm{q}};\bar{M})$. Meanwhile,
\begin{equation}
\begin{split}
        \langle \bar{M}\odot\bar{\gamma}, \bar{G} \rangle_F =& \sum_{i=1}^m\sum_{j=1}^nM_{i,j}\gamma_{i,j}C_{i,j} + \xi (\Vert \bm{p}\Vert_1+ \Vert \bm{q}\Vert_1 -2s)\\
        < &\sum_{i=1}^m\sum_{j=1}^nM_{i,j}\ddot{\pi}_{i,j}G_{i,j}+ \xi (\Vert \bm{p}\Vert_1+ \Vert \bm{q}\Vert_1 -2s)\\
        = &\langle \bar{M}\odot\bar{\pi}^{*}, \bar{G} \rangle_F.
\end{split}
\end{equation}
This contradicts the fact that $\bar{M}\odot\bar{\pi}^{*}$ is the optimal transport plan of problem $\mbox{min}_{\bar{\pi} \in \Pi(\bar{\bm{p}},\bar{\bm{q}};\bar{M})} \langle \bar{M}\odot\bar{\pi}, \bar{G} \rangle_F$. Therefore, $M\odot\ddot{\pi} $ is the optimal transport plan of problem in Eq.~\eqref{eq:partial_kpp_main}.

\hfill$\blacksquare$

\section{Proof of Theorem~\ref{thm:dual}}
We rewrite the $\chi^2$-regularized model as
\begin{equation}
\begin{split}
    &\min_{\pi}\sum_{i,j}M_{i,j}\pi_{i,j}G_{i,j} + \epsilon \sum_{i,j}\frac{(M_{i,j}\pi_{i,j})^2}{p_iq_j} \\
    &\mbox{s.t. }\sum_{j}M_{i,j}\pi_{i,j} = p_i,\sum_{i}M_{i,j}\pi_{i,j} = q_j, \pi_{i,j}\geq 0.
\end{split}
\end{equation}
The Lagrange function is 
\begin{equation}
\begin{split}
    L(\pi,\phi,\psi)
    =&\sum_{i,j}M_{i,j}(G_{i,j}-\phi(x_i)-\psi(y_j))\pi_{i,j} + \epsilon \sum_{i,j}\frac{M_{i,j}\pi_{i,j}^2}{p_iq_j}\\
    +& \sum_i\phi(x_i)p_i + \sum_{j}\psi(y_j)q_j,
\end{split}
\end{equation}where we utilize $M_{i,j}^2 = M_{i,j}$.
Minimizing $L(\pi,\phi,\psi)$ \textit{w.r.t.} $\pi$ ($\pi_{i,j}\geq 0$) is equivalent to 
\begin{equation}
    \sum_{i,j} \min_{\pi_{i,j}\geq 0}\left\{M_{i,j}(G_{i,j}-\phi(x_i)-\psi(y_j))\pi_{i,j} + \epsilon\frac{M_{i,j}\pi_{i,j}^2}{p_iq_j}\right\}.
\end{equation}
If $M_{i,j} = 0$, the minimizer $\pi_{i,j}$ could be arbitrary non-negative value. If $M_{i,j} = 1$, 
the minimizer is $\pi_{i,j} =\frac{1}{2\epsilon}(-G_{i,j}+\phi(x_i)+\psi(y_j))_+p_iq_j$, where $a_+=\max\{a,0\}$. Hence, $M_{i,j}\pi_{i,j} = \frac{1}{2\epsilon}M_{i,j}(-G_{i,j}+\phi(x_i)+\psi(y_j))_+p_iq_j.$
The dual problem is 
\begin{equation}
    \begin{split}
    &\max_{\phi,\psi}\min_{\pi}L(\pi,\phi,\psi)\\
    =&\max_{\phi,\psi}  \sum_i\phi(x_i)p_i + \sum_{j}\psi(y_j)q_j -\frac{1}{2\epsilon}\sum_{i,j}M_{i,j}[(-G_{i,j}+\phi(x_i)+\psi(y_j))_+]^2p_iq_j\\
    &+\frac{1}{4\epsilon}\sum_{i,j}M_{i,j}(-G_{i,j}+\phi(x_i)+\psi(y_j))_+^2p_iq_j\\
    =&\max_{\phi,\psi}  \sum_i\phi(x_i)p_i + \sum_{j}\psi(y_j)q_j -\frac{1}{4\epsilon}\sum_{i,j}M_{i,j}(-G_{i,j}+\phi(x_i)+\psi(y_j))_+^2p_iq_j.
    \end{split}
\end{equation}
By the strong duality, the optimal solution $\phi^*, \psi^*$ satisfies
\begin{equation}
    (M\odot\pi^*)_{i,j} =\frac{1}{2\epsilon}M_{i,j}(-G_{i,j}+\phi^*(x_i)+\psi^*(y_j))_+p_iq_j.
\end{equation}

\hfill$\blacksquare$

\section{Implementation Details for I2I Translation Experiment}
The experiment consists of two steps. In the first step, we learn the potentials $\phi$ and $\psi$ based on Eq.~\eqref{eq:dual} and compute the transport plan as Eq.~\eqref{eq:prim_dual_solution}. 
In the second step, we learn the manifold barycentric projection $T_{MBP}$ by Eq.~\eqref{eq:gan_based} or the manifold sampling map $T_{MSP}$ by Eq.~\eqref{eq:msp}. {During training, at each step, we randomly sample a small set of unpaired images and merge them with all paired images to construct a mini-batch. We assign uniform mass to all samples within the mini-batch.} 
\subsection{Details for the First Step}
\textit{Architectures of the potentials $\bm{\phi}$ and $\bm{\psi}$ in Eq.~\eqref{eq:dual}.} 
Both ${\phi}$ and ${\psi}$ consist of a feature extractor, followed by an output head. The output head is constructed as follows:
FC(1024,1024) $\rightarrow$ ReLU $\rightarrow$ FC(1024,1024) $\rightarrow$ ReLU $\rightarrow$ FC(1024,1024) $\rightarrow$ ReLU $\rightarrow$ FC(1024,1024) $\rightarrow$ ReLU $\rightarrow$ FC(1024,1),
where ``FC($a,b$)'' is the Fully-Connected layer with input/output channel of $a$/$b$. The feature extractor is pretrained and fixed. For digits, we train autoencoders for source and target images respectively, and take the encoder part of the autoencoder as the feature extractor. The architecture of the encoder part is as follows: Conv(1,32,3,1) $\rightarrow$ GN(4) $\rightarrow$ ReLU $\rightarrow$ Conv(32,64,3,2) $\rightarrow$ GN(32) $\rightarrow$ ReLU $\rightarrow$ Conv(64,128,3,2) $\rightarrow$ GN(32) $\rightarrow$ ReLU $\rightarrow$ Conv(128,256,3,2) $\rightarrow$ GN(32) $\rightarrow$ L2-Norm. The architecture of the decoder part of the autoencoder is as follows: Tconv(256,128,3,2) $\rightarrow$ GN(32) $\rightarrow$ ReLU $\rightarrow$ Tconv(128,64,3,2) $\rightarrow$ GN(32) $\rightarrow$ ReLU $\rightarrow$ Tconv(64,32,3,2) $\rightarrow$ GN(32) $\rightarrow$ ReLU $\rightarrow$ Tconv(32,1,3,1) $\rightarrow$ Sigmoid. ``Conv($a,b,k,s$)'' and ``Tconv(a,b,k,s)'' are respectively the Convolutional layer and the Transposed-Convolutional layer, where $a$ and $b$ are the input and output channel respectively, the kernel size is $k\times k$, and the stride is $s$. ``GN($m$)'' is the Group Normalization layer with $m$ groups. ``L2-Norm'' is the $L_2$-normalization. For the natural animal images, the feature extractor is taken as the image encoder (``ViT-B/32'') of CLIP~\citep{radford2021learning}. 

\vspace{0.4\baselineskip}\textit{Training details for learning the potentials $\bm{\phi}$ and $\bm{\psi}$ in Eq.~\eqref{eq:dual}.}
When training with Eq.~\eqref{eq:dual}, the optimization algorithm is Adam, the learning rate is 1e-5, and the batch size is 64. $\epsilon$ in Eq.~\eqref{eq:l2_reg_kpg} is set to 0.005.

\subsection{Details for the Second Step}
\textit{Architectures of $\bm{T'}$ in Eq.~\eqref{eq:gan_based} and $\bm{D}$ in Eq.~\eqref{eq:loss_gan}.} For digits, the architectures of $T'$ is as follows: Conv(1,64,4,2) $\rightarrow$ BN $\rightarrow$ ReLU $\rightarrow$ Conv(64,128,4,2) $\rightarrow$ BN $\rightarrow$ ReLU $\rightarrow$ Conv(128, 128,3,1)  $\rightarrow$ BN  $\rightarrow$ ReLU $\rightarrow$ Conv(128,128,3,1) $\rightarrow$ BN $\rightarrow$ ReLU $\rightarrow$ Tconv(128,64,4,2) $\rightarrow$ BN $\rightarrow$ ReLU $\rightarrow$ Tconv(64,1,4,2)  $\rightarrow$ Sigmoid, where ``BN'' is the Batch Normalization layer.
The architectures of $D$ is as follows: Conv(1,64,4,2) $\rightarrow$ ReLU $\rightarrow$ Conv(64,128,4,2) $\rightarrow$ BN $\rightarrow$ ReLU $\rightarrow$ Conv(128,256,4,2) $\rightarrow$ BN $\rightarrow$ ReLU $\rightarrow$ Conv(256,1,3,0).

For natural animal images, to reduce the training burden, we map the images to the lantent feature space using the encoder of Stable Diffusion and then perform our MBP and MSP in the latent space. In the latent space, we construct $T'$ and $D$ using linear layers. Specifically, the architecture of $T'$ is as follows: Linear(4096,2048) $\rightarrow$ ReLU $\rightarrow$ Linear(2048, 512) $\rightarrow$ ReLU $\rightarrow$ Linear(512, 128) $\rightarrow$ ReLU $\rightarrow$ Linear(128,512) $\rightarrow$ ReLU $\rightarrow$ Linear(512,2048) $\rightarrow$ ReLU $\rightarrow$ Linear(2048,4096). For MSP, the noise $z$ is 128-dimensional which is added to the features after layer Linear(512, 128). The architecture of $D$ for MBP is as follows:  Linear(4096,2048) $\rightarrow$ ReLU $\rightarrow$ Linear(2048, 512) $\rightarrow$ ReLU $\rightarrow$ Linear(512, 1). For MSP, we concatenate source features and generated target features as the input of $D$, for which the input dimension of $D$ is changed to 8192.

\vspace{0.4\baselineskip}\textit{Training details for learning $\bm{T'}$.} 
When training $T'$ with Eq.~\eqref{eq:gan_based}, the optimization algorithm is Adam, the learning rate is 1e-4, and the batch size is 64.  $\bar{d}$ is set to cosine distance in feature space.

\section{Runtime and Peak Memory Scaling Comparison}\label{app:time_memory}
{

\begin{table}[t]
    \centering
    \setlength{\tabcolsep}{10pt}
    \begin{tabular}{llcccccc}
        \toprule
        {Sample size} && {5000} & {10000} & {20000} & {30000} & {50000} & {80000} \\
        \midrule
        \multirow{2}{*}{{Time (s)}} 
        & Sinkhorn & 0.8 & 1.8 & 12.0 & 28.4 & OOM & OOM \\
        & Neural & 21.5 & 41.6 & 87.6 & 132.8 & 217.3 & 346.9 \\
        \midrule
        \multirow{2}{*}{{Memory (GB)}} 
        & Sinkhorn & 0.68 & 2.25 & 8.94 & 20.12 & OOM & OOM \\
        & Neural & 0.02 & 0.02 & 0.02 & 0.02 & 0.02 & 0.02 \\
        \bottomrule
    \end{tabular}
        \caption{Runtime (in seconds) and peak memory (in GB) scaling comparison between Sinkhorn's algorithm and neural transport method across different sample sizes ($m$). ``OOM'' indicates Out-of-Memory.}
    \label{tab:scaling_comparison}
    % } 
\end{table}
% Here, we compare the scalability of our neural-dual approach against the primal formulation (solved via Sinkhorn) using the same data as the ``Toy experiments for evaluating KPG-RL-KP and KPG-RL-GW'' in Section 6.1. We consider different sample size $m$ ($m=n$ in this experiment),  ranging from $5,000$ to $80,000$. We set $\epsilon=0.001$ for both primal and neural-dual formulations. For the neural-dual formulation, we train the networks for 100 epoch. The experiment is conducted on a Tesla-V100 GPU (32GB). The runtime and peak memory are reported in Table~\ref{tab:scaling_comparison}, we can see that as the sample size increases, both runtime and peak memory of primal formulation increases rapidly (nearly exponential). When the sample size is larger than 50000, the primal formulation out-of-memmory. Conversely, the neural-dual formulation maintains a constant memory usage ($\approx 0.02$ GB) in training regardless of $m$, as it relies on mini-batch optimization rather than storing full coupling matrices. Meanwhile, the runtime for neural-dual formulation seems to increase linearly as sample size increases. The results indicate that for larger sample sizes, the neural-dual formulation is a feasible way to solve the optimal transport model. While for smaller sample sizes, the primal formulation takes shorter computational time.

We evaluate the scalability of our neural transport (based on dual formulation) approach against the primal formulation (solved via Sinkhorn), utilizing the same data setup as the ``Toy experiments for evaluating KPG-RL-KP and KPG-RL-GW'' in Section 6.1. We vary the sample size $m$ (setting $m=n$) from $5,000$ to $80,000$, with $\epsilon$ fixed at $0.001$ for both methods. The neural transport model is trained for 100 epochs. All experiments are conducted on a Tesla-V100 GPU (32GB). Table~\ref{tab:scaling_comparison} reports the runtime and peak memory usage. We observe that as the sample size increases, Sinkhorn's algorithm seems to exhibit nearly exponential growth in both runtime and memory, encountering out-of-memory (OOM) errors when $m \geq 50,000$. In contrast, the neural transport approach maintains constant peak memory usage ($\approx 0.02$ GB) regardless of sample size, as it relies on mini-batch optimization rather than storing full coupling matrices. Furthermore, its runtime seems to scale linearly with sample size. These results indicate that while Sinkhorn's algorithm is computationally efficient for smaller sample sizes, the neural transport approach offers superior scalability for large-scale transport problems.
}

% \noindent
% {\bf Theorem} {\it Let $u,v,w$ be discrete variables such that $v, w$ do
% not co-occur with $u$ (i.e., $u\neq0\;\Rightarrow \;v=w=0$ in a given
% dataset $\dataset$). Let $N_{v0},N_{w0}$ be the number of data points for
% which $v=0, w=0$ respectively, and let $I_{uv},I_{uw}$ be the
% respective empirical mutual information values based on the sample
% $\dataset$. Then
% \[
% 	N_{v0} \;>\; N_{w0}\;\;\Rightarrow\;\;I_{uv} \;\leq\;I_{uw}
% \]
% with equality only if $u$ is identically 0.} \hfill\BlackBox

% \noindent
% {\bf Proof}. We use the notation:
% \[
% P_v(i) \;=\;\frac{N_v^i}{N},\;\;\;i \neq 0;\;\;\;
% P_{v0}\;\equiv\;P_v(0)\; = \;1 - \sum_{i\neq 0}P_v(i).
% \]
% These values represent the (empirical) probabilities of $v$
% taking value $i\neq 0$ and 0 respectively.  Entropies will be denoted
% by $H$. We aim to show that $\fracpartial{I_{uv}}{P_{v0}} < 0$....\\

% {\noindent \em Remainder omitted in this sample. See http://www.jmlr.org/papers/ for full paper.}

\vskip 0.2in
\bibliography{refe_correct}

\end{document}